%% file: SA_Tweedie_main_unmarked.tex
\documentclass[12pt, article
]{imsart}

\RequirePackage[OT1]{fontenc}
\RequirePackage{amsthm,amsmath} %,natbib
\usepackage{natbib}
\RequirePackage[colorlinks,citecolor=blue,urlcolor=blue]{hyperref}
\hypersetup{
    linktocpage=true,
    colorlinks=true,
    bookmarks=true,
    citecolor=blue,
    urlcolor=blue,
    linkcolor=blue,
    citebordercolor={1 0 0},
    urlbordercolor={1 0 0},
    linkbordercolor={.7 .8 .8},
    breaklinks=true,
    pdfpagelabels=true,
    }
\RequirePackage{hypernat}
\usepackage{amssymb}
\usepackage{multirow}
\usepackage{capt-of}
\usepackage{amsmath}
\usepackage{algorithm}
\usepackage[noend]{algpseudocode}
\usepackage{graphicx}

\usepackage{lineno}
\graphicspath{ {./figures/} }

\usepackage{booktabs,caption,fixltx2e}
\usepackage[flushleft]{threeparttable}
\usepackage{subcaption}
\usepackage{relsize}

\usepackage[compact]{titlesec}
\titlespacing{\section}{1pt}{*0.5}{*0.1}
\titlespacing{\subsection}{0.7pt}{*0.3}{*0.1}
\titlespacing{\subsubsection}{0pt}{*0}{*0}

\RequirePackage{xcolor, colortbl} % To provide color for soul (for english editing), for adding cell color of table
\RequirePackage{todonotes}
\definecolor{LightCyan}{rgb}{0.88,1,1}
\makeatletter
\newcommand{\hlynob}[1]{%
    \setbox\@tempboxa\hbox{#1}%
    \ifdim\wd\@tempboxa>\linewidth
    \noindent
    \colorbox{white}{%
        \parbox{\dimexpr\linewidth-2\fboxsep}{#1}%
    }%
    \else
    \colorbox{white}{#1}%
    \fi}%Highlighter.
\makeatother

\makeatletter
\newcommand{\hlboth}[1]{%
    \setbox\@tempboxa\hbox{#1}%
    \ifdim\wd\@tempboxa>\linewidth
    \noindent
    \colorbox{white}{%
        \parbox{\dimexpr\linewidth-2\fboxsep}{\textcolor{black}{#1}}%
    }%
    \else
    \colorbox{white}{\textcolor{black}{#1}}%
    \fi}%Highlighter.
\makeatother
\usepackage{framed}
\usepackage{enumitem}

\newcommand{\modified }{\color{black}  }

\makeatletter
\def\BState{\State\hskip-\ALG@thistlm}
\makeatother

\input{mac}
\begin{document}
	
\thispagestyle{empty}		
	\begin{frontmatter}
		\title{Global dense vector representations for words or items using shared parameter alternating Tweedie model % in cancer classification
}
		%Title on one or two lines
		%\newline without capitals, except after a colon
		%}} }
		
		\vglue 5mm

		\begin{aug}
			%\author{\fnms{Xukun} \snm{Li}\ead[label=e1]{ xukun@ksu.edu}}
			%,
			\author{\fnms{Taejoon } \snm{Kim$^1$}\ead[label=e1]{kimtj6895@ksu.edu}}
and
\author{\fnms{Haiyan} \snm{Wang$^{2a}$}\corref{} \ead[label=e2]{hwang@ksu.edu}}
%and
%\author{\fnms{Santosh} \snm{Ghimire$^3$} \ead[label=e3]{santoshghimire@ioe.edu.np} }
			\affiliation{{\small Department of Statistics, Kansas State
					University, 101 Dickens Hall, Manhattan, KS 66506}
				%\printead{e2}
			}

			\address{
               $^1$  Department of Statistics and Biostatistics, California State University East Bay\\
               $^2$ Department of Statistics, Kansas State University\\
              % $^3$ Department of Applied Sciences and Chemical Engineering, Pulchowk Campus, Tribhuvan University
				% }
			}

			\runauthor{Kim \& Wang
}
		\end{aug}
		\footnotetext{\textit{$^*$ Running head: SA-Tweedie for Matrix factorization and prediction}}
		
		\footnotetext{\textit{$^{a}$ Author to whom correspondence may be
				addressed. Email: hwang@ksu.edu}}
		%\medskip
		%\footnotetext{\textit{$^a$ Affiliation   }}
		
		\runtitle{Matrix factorization and prediction using SA-Tweedie}

\setcounter{page}{1}

\vspace{0.2in}
\begin{abstract}
In this article, we present a model for analyzing the cooccurrence count data derived from practical fields such as user-item or item-item data from online shopping platform, cooccurring word-word pairs in sequences of texts. Such data contain important information for developing recommender systems or studying relevance of items or words from non-numerical sources. Different from traditional regression models, there are no observations for covariates. Additionally, the cooccurrence matrix is typically of so high dimension that it does not fit into a computer’s memory for modeling.
We extract numerical data by defining windows of cooccurrence using weighted count on the continuous scale. Positive probability mass is allowed for zero observations. We present Shared parameter Alternating Tweedie (SA-Tweedie) model and an algorithm to estimate the parameters. We introduce a learning rate adjustment used along with the Fisher scoring method in the inner loop to help the algorithm stay on track of optimizing direction. Gradient descent with Adam update was also considered as an alternative method for the estimation. Simulation studies and an application showed that our algorithm with Fisher scoring and learning rate adjustment outperforms the other two methods. Pseudo-likelihood approach with alternating parameter update was also studied. Numerical studies showed that the pseudo- likelihood approach is not suitable in our shared parameter alternating regression models with unobserved covariates.
\end{abstract}

	\end{frontmatter}

\input{chapter4_unmarked.tex}

\bibliographystyle{apa} %apacite
\bibliography{references}
% renumbering the equations as A1 etc.
\renewcommand{\theequation}{A.\arabic{equation}}
  % redefine the command that creates the equation no.
 \setcounter{equation}{0}  % reset counter
 % \section*{APPENDIX}  % use *-form to suppress numbering

 \setcounter{section}{0}\renewcommand\thesection {A}%!%  use this line to renumber the theorems or lemmas as A.1 A.2 etc.
%\setcounter{theorem}{0}
%\renewcommand\theproposition{A.\arabic{theorem}}
%\renewcommand\thelemma{A.\arabic{theorem}}
%\renewcommand\thetheorem{A.\arabic{theorem}}

% \setcounter{lemma}{0}
% \renewcommand\thelemma{A.\arabic{lemma}}

%\appendix
%\input{appendixA}

\end{document}

%% file: mac.tex
\makeatletter
\@addtoreset{equation}{section}
\renewcommand{\theequation}{\thesection.\arabic{equation}}
\makeatother

\relax

\newcommand{\vb}{{\bf b}}

\newcommand{\vg}{{\bf g}}

\newcommand{\vw}{{\bf w}}
\newcommand{\vx}{{\bf x}}

\newcommand{\wtb}{{\widetilde b}}
\newcommand{\wtvb}{\mbox{\boldmath $\wtb$}}

\newcommand{\wtw}{{\widetilde w}}
\newcommand{\wtvw}{\mbox{\boldmath $\wtw$}}

\newcommand{\wtbeta}{{\widetilde \beta}}
\newcommand{\wtvbeta}{\mbox{\boldmath $\wtbeta$}}

\newcommand{\vbeta}{\mbox{\boldmath $\beta$}}

\newcommand{\bay}{\begin{array}}
\newcommand{\eay}{\end{array}}

\newcommand{\Var}{\mbox{Var}}

\newcommand{\iidsim}{\stackrel{i.i.d.}{\sim}}

\newcommand{\bqa}{\begin{eqnarray*}}
\newcommand{\eqa}{\end{eqnarray*}}
\newcommand{\bqan}{\begin{eqnarray}}
\newcommand{\eqan}{\end{eqnarray}}
\newcommand{\bqt}{\begin{quote}}
\newcommand{\eqt}{\end{quote}}
\newcommand{\bt}{\begin{tabbing}}
\newcommand{\et}{\end{tabbing}}
\newcommand{\bit}{\begin{itemize}}
\newcommand{\eit}{\end{itemize}}
\newcommand{\ben}{\begin{enumerate}}
\newcommand{\een}{\end{enumerate}}
\newcommand{\beq}{\begin{equation}}
\newcommand{\eeq}{\end{equation}}
\newcommand{\bdefi}{\begin{definition}}
\newcommand{\edefi}{\end{definition}}
\newcommand{\bpro}{\begin{proposition}}
\newcommand{\epro}{\end{proposition}}
\newcommand{\blem}{\begin{lemma}}
\newcommand{\elem}{\end{lemma}}
\newcommand{\bth}{\begin{theorem}}
\newcommand{\bco}{\begin{corollary}}
\newcommand{\eco}{\end{corollary}}
\newcommand{\bdes}{\begin{description}}
\newcommand{\edes}{\end{description}}

\newtheorem{definition}{Definition}[section]
\newtheorem{theorem}[definition]{Theorem}

%% file: chapter4_unmarked.tex
% \documentclass{article}
% \usepackage[utf8]{inputenc}
% \usepackage{multirow}
% \usepackage{graphicx}
% \usepackage{amssymb}
% \usepackage{url}
% \usepackage{amsmath}
% \graphicspath{ {./figures/} }

% \title{Research 2}
% \author{kimtj6895}
% \date{June 2022}

% \input{mac}

% \begin{document}

% \maketitle
%\LARGE
%\Large
% \cleardoublepage
% \chapter{Alternating Tweedie regression model with shared parameters}
\section{Introduction}
Data tables often summarize relationships between response and explanatory variables. In regression analysis, rows represent observations, while columns list covariates explaining variations in the response. This article explores modeling with data matrices that aggregate information on pairs of members, differing from traditional regression matrices.

For instance, in product recommendation systems, a matrix may represent user-product interactions, such as Amazon ratings. Rows correspond to users, columns to products, and entries capture ratings or browsing times. Here, users and products aren't explanatory variables but entities requiring analysis to infer patterns for recommendations.

Another example is the word-word co-occurrence matrix in NLP. Each entry captures the frequency of a target word co-occurring with a context word within a defined window. These sparse matrices are typically large with many zeros. They can be used to uncover relationships between words. Learned dense vectors, or word embeddings, enable clustering similar words and support downstream tasks like sentiment analysis and named entity recognition.

In both examples, co-occurrence counts summarize data effectively but are not the only approach. For instance, Word2Vec models each word in a context window as a binary variable, treating the vocabulary as multiclass data. It uses multinomial logistic regression to estimate the probability of a word being a context or center word, using the dot product of their dense representations as the input for a softmax function. This approach faces challenges with large vocabularies, necessitating techniques like hierarchical softmax and negative sampling to reduce computational demands. In contrast, GloVe aggregates co-occurrence counts and uses weighted least squares regression to model them. However, GloVe does not consider the fact that the variation of such counts also changes with the mean.

\hlboth{
To handle unseen words and rare cases, Facebook's FAIR lab developed FastText, which extends Word2Vec by representing words as bags of character n-grams, effectively addressing challenges in morphologically rich languages.}

\hlboth{ Recent advancements in natural language processing (NLP) have focused on contextualized word embeddings that capture word meanings based on their surrounding context within entire sentences or passages. For instance, models like ELMo \citep{Peters_et_al2018} use deep bidirectional LSTMs, while BERT \citep{devlin-etal-2018-bert} employs transformer-based architectures pre-trained on large corpora to achieve deep contextual embeddings through tasks like masked language modeling and next sentence prediction. These innovations have significantly advanced various NLP applications by offering richer, context-aware representations of language.}

\hlboth{ Simultaneously, the scaling up of large language models (LLMs) based on transformer architectures with self-attention mechanisms has enabled more efficient sequence processing and enhanced language understanding. Notable examples like GPT-3 \citep{GPT3} have demonstrated that larger models, when trained on massive datasets, deliver superior performance on a wide range of tasks without the need for fine-tuning. Similarly, T5 \citep{T5} has shown that framing all NLP tasks as text-to-text problems allows for versatile embeddings, further broadening the applicability of LLMs across diverse domains.}

\hlboth{
Despite the impressive performance of LLMs, there are critical limitations that pose challenges to their widespread adoption. \cite{Freestone2024WordER} compared state-of-the-art LLMs, such as LLaMA2-7B \citep{touvron2023llama2openfoundation}, ADA-002, and PaLM2, with classical models like Universal Sentence Encoder and BERT. While LLMs generally outperformed classical methods in analogy tasks, their heavy reliance on enormous computational resources and large training datasets remains a significant bottleneck. The sheer size of modern LLMs, such as GPT-4, with billions of parameters, demands extensive hardware capabilities (e.g., high-performance pods of GPUs or TPUs), making them costly to train, deploy, and maintain, not to mention their environmental impact.}

\hlboth{With the rise of LLMs, there has been increasing interest in applying these models to improve the accuracy of recommender systems. While LLMs have the potential to enhance recommender systems when large-scale text data is available, they are less effective in handling sparse user-item interaction data that is characteristic of many recommender systems \citep{raza2024comprehensivereviewrecommendersystems}. Their high computational cost makes them impractical for real-time recommendation systems, especially on platforms with a high number of users and items. In fact, when applied to collaborative filtering-based systems, LLMs often fail to capture the latent factors that underlie user-item interactions in a way that other specialized models, like matrix factorization or hybrid models, can.}

\hlboth{
The increasing resource demands of large-scale models highlight the need for innovative solutions that reduce their size and computational requirements without sacrificing performance. Transformer-based models, which rely on global word embeddings to understand language in context, could potentially be optimized to improve both efficiency and accessibility. Enhancing these embeddings offers an opportunity to create lightweight models capable of delivering competitive performance while mitigating resource consumption.}

\hlboth{One promising approach is Matryoshka Representation Learning (MRL) introduced by \cite{Kusupati_et_al_2024}, which focuses on creating flexible embedding models that can shrink in size for specific tasks, preserving high performance while reducing computational demands. Building on this idea, the Matryoshka-Adaptor framework has been successfully applied to LLMs, resulting in significant reductions in dimensionality and improved computational efficiency. These advancements underscore the need for more efficient word representation models that strike a balance between performance and resource consumption, particularly in scenarios where deploying large models is impractical.}

\hlboth{
Global word representations have a wide range of applications that can benefit from these advances. For instance, in fields such as deepfake detection (\citealt{deepfake1}, \citealt{deepfake2}), medical informatics (\citealt{IoT3}, \citealt{IoT8}), and climate research \citep{climate11}, efficient word embeddings are essential for processing large volumes of textual data, extracting actionable insights, and making decisions in real-time. Additionally, emerging applications in Internet of Things (IoT) systems (\citealt{IoD6}, \citealt{IoD7}) and blockchain-based technologies \citep{IoT10} also leverage word embeddings for tasks like pattern recognition, anomaly detection, and performance optimization. The versatility of these embeddings across domains highlights their fundamental role in natural language understanding and text-driven decision-making.}

\hlboth{Given these pressing challenges and opportunities, this paper proposes a foundational approach for modeling global word representations from a probabilistic perspective. The rest of the paper is organized as follows: Section \ref{modeling} introduces the proposed SA-Tweedie model, detailing its formulation and probability foundation. Section \ref{small_numerical} presents a simulation study to explore the model's characteristics and validate its performance. Section \ref{large_scale_app} discusses scalability issues and demonstrates the model’s ability to train on large datasets, such as the entire Wikipedia corpus, with an application to Named Entity Recognition. Finally, the paper concludes with a summary of the findings and implications for future research.
}

\section{The probability distribution for the proposed SA-Tweedie model}
\label{modeling}
\subsection{MLE for alternating Tweedie regression}

In NLP tasks, the probability of one word appearing in the context of another is typically very small, and the large volume of text results in sparse occurrences. While the count of such co-occurrences is a binomial random variable, it can often be approximated by a Poisson distribution, though the data frequently contains excessive zeros.

The Compound Poisson Gamma distribution is well-suited for this scenario. This distribution models the sum of independent Gamma random variables with identical shape and scale parameters, where the number of Gamma variables follows a Poisson distribution. Belonging to the Tweedie family, this distribution allows for shared shape and scale parameters across identical item-item pairs from different sources, while different pairs may have distinct parameters. These parameters are ultimately estimated within a regression framework.

In this section, we introduce the alternating Tweedie regression model with shared parameters and detail its parameter estimation. Before presenting the regression model, we explore the Tweedie distribution in greater depth and explain the relationship between the parameters of the Gamma and Tweedie distributions.

The Tweedie distribution family contains several commonly used distributions which are determined by the power parameter $p$.
\bit
\item The Tweedie distribution belongs to exponential family. All exponential dispersion models that are closed to scale transformation belongs to the Tweedie distribution family.
    %(see Theorem 4.1 of \citealt{Jorgensen:1997}).
    In particular, if $X$ has Tweedie distribution with mean $\mu$ and dispersion parameter $\phi$, then $cX$ has Tweedie distribution with mean $c\mu$ and dispersion parameter $c^{2-p}\phi$. It has mean and variance relationship as follows
$$E(Y)=\mu \quad \operatorname{Var}(Y)=\phi \mu^p,$$
where $\phi$ is the dispersion parameter having value greater than 0.
\item The Compound Poisson Gamma distribution corresponds to the case that the power parameter satisfies $p \in (1,2)$. In this case, the distribution has non-negative support and can have a discrete mass at zero, making it useful to model responses that are a mixture of zeros and positive values.
\item The distributions that correspond to other power values of $p$:
\bit
\item When $p<0$, the distribution is the Extreme stable distribution.
\item When $p=0$, the distribution is the Normal distribution.
\item When $p=1$, the distribution coincides with the Poisson distribution.
\item When $p=2$, the distribution is the Gamma distribution.
\item When $2<p<3$, the distribution is the Positive stable distribution.
\item When $p=3$, the distribution is the Inverse Gaussian distribution.
\item When $p>3$, the distribution is the Positive stable distribution.
\item When $p=\infty$, the distribution is the Extreme stable distribution.
\eit
\item The mean domain is $R$ if $p$ is 0 or $\infty$, and is $R_+$ for all other $p \neq 0$. Therefore, except when the distribution is Normal or Extreme stable with $p=\infty$, the log link is frequently used as the link function for the Tweedie regression.
\item The positive zero mass is only allowed when $1 \leq p<2$. Since our cooccurrence count has many zeros, we focus our attention to the Compound Poisson Gamma distribution which is the case when $1<p<2$.
\eit

The Gamma and Poisson distributions together have parameters $\alpha$, $\beta$, and $\lambda$. The Tweedie distribution uses parameters $\mu$, $\phi$, $p$. Next, we will see how they are related in the case of Compound Poisson Gamma.

Let $N \sim \operatorname{Poisson}(\lambda), \quad X_i \iidsim \operatorname{Gamma}(\alpha, \beta).$ Assume $N$ and  $X_i$'s are independent for all $i$.
For $Y=0$, the probability comes from the Poisson distribution
$$
\operatorname{Pr}[Y=0]=\frac{\lambda^y e^{-\lambda}}{y !}=e^{-\lambda}.
$$

For $Y>0$, note that $Y \mid N=n \sim \operatorname{Gamma}(n \alpha, \beta)$ by additive property of the Gamma distribution. Then the distribution of $Y$ when $Y>0$ can be obtained with the law of total probability as the product of conditional distribution of $Y\mid N$ and the marginal distribution of $N$, %( $\because$ if $x_i \sim \operatorname{Gamma}\left(\alpha_i, \beta\right)$ and $x_i$ 's are indep., then $\sum_{i=1}^n x_i \sim \operatorname{Gamma}\left(\sum_{i=1}^n \alpha_i, \beta\right)$. easy to proof by using moment gorenating foc.) To obtain manginal dist. of $Y$, use the following fact.
\bqa
f_Y(y) = \sum_{n=0}^{\infty} f_{Y, N}(y, n) =\sum_{n=0}^{\infty} f_{Y I N}(y \mid n) \cdot f_N(n).
\eqa

Recall, $N \sim \operatorname{Poisson}(\lambda)$ and $Y \mid N=n \sim \operatorname{Gamma}(n \alpha, \beta)$. Therefore, for $y>0$,
\bqan
\label{formual:Tweedie-pdf}
\nonumber
f_Y(y) &=&\sum_{n=0}^{\infty} \frac{\beta^{n \alpha}}{\Gamma(n \alpha)} y^{n \alpha-1} e^{-\beta y} \cdot \frac{\lambda^n e^{-\lambda}}{n !} \\
&=&e^{-\beta y} \cdot e^{-\lambda} \cdot \sum_{n=0}^{\infty} \frac{\beta^{n x}}{\Gamma(n \alpha)}
\nonumber
y^{n \alpha-1} \cdot \frac{\lambda^n}{n !} \\
&=&\exp \left\{-\beta y-\lambda +\log \left(\sum_{n=0}^{\infty} \frac{\beta^{n \alpha}}{\Gamma(n \alpha)} \cdot y^{n \alpha-1} \cdot \frac{\lambda^n}{n !}\right)\right\}.
\eqan
The above form was derived by using the Poisson distribution and conditional distribution of Gamma. Meanwhile, the Tweedie distribution has its own canonical form in exponential family format when $Y>0$:
\bqan \label{canonical}
f\left(y\right) =\exp \left\{\frac{y \theta - \kappa\left(\theta \right)}{\phi}+c\left(y, \phi, p\right)\right\},
\eqan
where $\theta = \frac{\mu^{1-p}}{1-p}$ and $\kappa(\theta) = \frac{\mu^{2-p}}{2-p}$. This implies that
\bqa
\lambda = \frac{\mu^{2-p}}{\phi(2-p)}, \quad \alpha =\frac{2-p}{p-1}, \quad \frac{1}{\beta}=\phi (p-1) \mu^{p-1},
\eqa
and the $c(y, \phi, p)$ corresponds to the log sum in (\ref{formual:Tweedie-pdf}).

%%%%%%%%%%%%%%%%%%%%%%%%%%%%%%%%%%%%%%%%%%%%%%%%%%%%%%%%%%%%%%%%%%%%

\begin{figure}[h!]
\includegraphics[width = \textwidth]{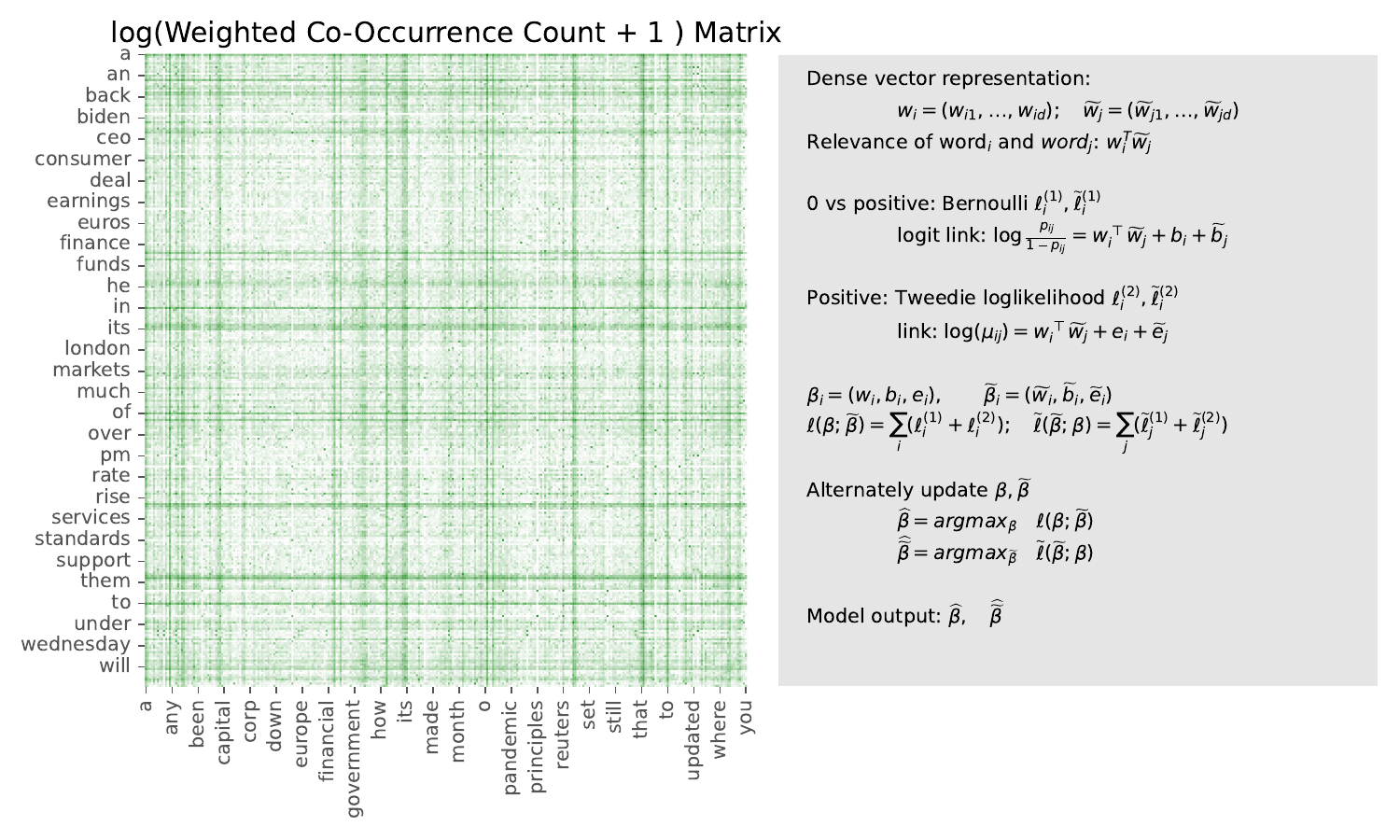}
\caption{{\modified Illustration of model input and desired output. Left panel: Model input - the natural log of (weighted occurrence count +1) matrix for top 300 words from Reuter Business news data. Right panel: Shared parameter Tweedie modeling process and output}}
\label{fig:data_illus}
\end{figure}

Once we observe the cooccurence count from the data and assuming the count is from the Compound Poisson Gamma, we could express probability density function in canonical form for the cooccurrence counts {\modified using a generalized linear model format.
 Figure \ref{fig:data_illus} illustrates the model input and desired output. The input data is the cooccurrence matrix as illustrated in the heatmap for the most frequently observed 300 words extracted from 2000 Reuter business news. The zero entries are represented with white color in the heatmap. The outputs of the model are the dense vector representations estimated for each word from the model.}

The model is as follows:
% $Y_{ij} \sim $ Compound Poisson Gamma (i.e. $1<p<2$) and its pdf could be expressed in canonical form as below.

For $y_{i j}=0$ case,
$$
f_0(0) =\operatorname{Pr}\left[Y_{i j}=0\right]=\exp \left(-\lambda_{i j}\right), $$
where $$\lambda_{i j} =\frac{\mu_{i j}^{2-p_{ij}}}{\phi_{ij} (2-p_{ij})} \quad \text {(see page 5 of \citealt{Bonat-Kokonendji:2017})} $$
For $y_{i j}>0$ case,
$$ f\left(y_{i j}\right) =\exp \left\{\frac{y_{i j} \theta_{i j}-\kappa\left(\theta_{i j}\right)}{\phi_{ij}}+c\left(y_{i j}, \phi_{ij}, p_{ij}\right)\right\}
$$
where
\bqan
\nonumber
&& \theta_{i j}=\frac{\mu_{i j}^{1-p_{ij}}}{1-p_{ij}}=\frac{\exp \left\{(1-p_{ij})\left(\vw_{i}^{\top} \wtvw_{j}+b_{i}+\wtb_{j}\right)\right\}}{(1-p_{ij})}, \\
% &a(\phi)=\phi \\
\nonumber
&& \kappa\left(\theta_{i j}\right)=\frac{\mu_{i j}^{2-p_{ij}}}{2-p_{ij}}=\frac{\exp \left\{(2-p_{ij})\left(\vw_{i}^{\top} \wtvw_{j}+b_{i}+\wtb_{j}\right)\right\}}{(2-p_{ij})}, \\
\label{function:c_fct_1}
&& c\left(y_{i j}, \phi_{ij}, p_{ij}\right)=\log \left(\sum_{k=1}^{\infty} \frac{\gamma_{ij}^{k a_{ij}}}{\Gamma(k a_{ij})} \cdot y_{i j}^{k a_{ij}-1} \cdot \frac{\lambda_{ij}^{k}}{k !}\right), \quad \text{ if } 1 < p_{ij} < 2, \\
&&
\label{function:c_fct_2}
c(y_{i j}, \phi_{i j}, p_{i j})\!=\!\frac{1}{\pi y_{i j}} \!\sum_{n=1}^{\infty}\! \frac{\Gamma(1\!-\!a_{ij} n) \phi_{ij}^{n(\!-\!a_{ij}\!-\!1)}(p_{ij}\!-\!1)^{\!-\!a_{ij} n}}{(\!-\!1)^{n}\Gamma(1+n)(p_{ij}\!-\!2)^{n} y_{i j}^{\!-\!a_{ij} n}} \sin (n \pi a_{ij}), \text{ if } p_{ij} \!>\! 2, \qquad \\
&&
\label{function:c_fct_3}
c(y_{i j}, \phi_{i j}, p_{i j}) = \frac{1}{\pi y_{ij}} \sum_{k=1}^{\infty}
\frac{\Gamma\left(1+\frac{k}{a_{ij}}\right)(-y_{ij})^{k}}{k ! \lambda_{ij}^{\frac{k}{a_{ij}}} \kappa_{p_{ij}}^{\frac{k}{a_{ij}}}(1)} \sin \left(\frac{-k \pi}{a_{ij}}\right), \,  \text{ if } p_{ij} \!<\! 0,
\eqan
with
$$
\lambda_{ij}\!=\!\frac{\mu_{ij}^{2-p_{ij}}}{\phi_{ij}(2\!-\!p_{ij})}, \quad a_{ij}\!=\!\frac{2\!-\!p_{ij}}{p_{ij}\!-\!1}, \quad \frac{1}{\gamma_{ij}}\!=\phi_{ij}(p_{ij}\!-\!1) \mu_{ij}^{p_{ij}-1}, \quad \kappa_{p_{ij}}(1) \!=\! \frac{(p_{ij}\!-\!1)^{-a_{ij}}}{p_{ij}\!-\!2}(\!-\!1)^{\frac{1}{p_{ij}-1}}.
$$

We use commonly used $\log$ link function, which accounts for the mean domain well as mentioned earlier.
$$
\log \mu_{i j}=\vw_{i}^{T} \wtvw_{j}+b_{i}+\wtb_{j}, \quad i, j=1, \cdots, n \quad \text { and } \quad \vw_{i}, \wtvw_{j} \in \mathbb{R}^{d},  b_i, \wtb_{j} \in \mathbb{R}.
$$
Thus, the $\log$-likelihood function for $i^{th}$ row of data limited to the element $y_{ij}>0$ case while $\wtvw$ and $\wtvb$ are held fixed is
\bqan
\label{formula:Tweedie-lli}
\ell_i^{(1)}(\vw_i, b_i;\wtvw, \wtvb) = \sum_{j=1}^n \left[\frac{y_{i j} \theta_{i j}-\kappa\left(\theta_{i j}\right)}{\phi_{i j}}+c\left(y_{i j}, \phi_{i j}, p_{i j}\right)\right]I(y_{ij}>0),
\eqan
and the $\log$-likelihood function for $j^{th}$ column of data limited to the element $y_{ij}>0$ case while $\vw$ and $\vb$ are held fixed is
\bqan
\label{formula:Tweedie-lltildej}
\widetilde{\ell}^{(1)}_j(\wtvw_j, \wtb_j;\vw, \vb) = \sum_{i=1}^n \left[\frac{y_{i j} \theta_{i j}-\kappa\left(\theta_{i j}\right)}{\phi_{i j}}+c\left(y_{i j}, \phi_{i j}, p_{i j}\right)\right]I(y_{ij}>0),
\eqan
where
$$p_{ij} = p_i + \tilde{p}_{j}, \quad \phi_{ij} = \phi_i \cdot \tilde{\phi}_{j}.$$
Here, the power $p_{i j}$ is decomposed into two terms additively, one depends on the row and the other depends on the column index. The dispersion parameter is a scaling parameter so we wrote it as the product of two dispersion parameters, again one depends on the row and the other depends on the column. This decomposition is inspired by the fact that the item-item cooccurrence count data matrix is symmetric. If the data matrix is user-item matrix, this additive effect may not be sufficient. Unfortunately, if each $y_{i j}$ has its own $p_{i j}$, $\phi_{i j}$ and $\mu_{i j}$, then we do not have enough degrees of freedom to estimate those parameters.

Holding $\wtvw$ and $\wtvb$ fixed, the total $\log$ likelihood $\ell$ for all rows  is
\bqa
% \ell &=&\log \left[\prod_{i, j=1}^{n} f\left(y_{i j}\right)\right] \\
\ell(\vw, \vb; \wtvw, \wtvb) &=& \sum_{i=1}^{n} \sum_{j=1}^{n} \left\{\log f\left(y_{i j}\right) \cdot I\left(y_{i j}>0\right)+\log f_{0}\left(y_{i j}\right) \cdot I\left(y_{i j}=0\right)\right\} \\
&=& \sum_{i=1}^{n} \ell_{i}^{(1)}(\vw_i, b_i; \wtvw, \wtvb) + \sum_{i=1}^{n} \sum_{j=1}^{n} \ell_{i j}^{(0)}(\vw_i, b_i; \wtvw_j, \wtb_j),
\eqa
where
\bqa
\ell_{ij}^{(0)}(\vw_i, b_i; \wtvw_j, \wtb_j) = -\lambda_{i j}I\left(y_{i j}=0\right) \text{ and } \lambda_{i j}=\frac{\mu_{i j}^{2-p_{i j}}}{\phi_{i j}\left(2-p_{i j}\right)}.
\eqa
This is because $\operatorname{Pr}\left[Y_{ij} = 0\right]=\exp \left(-\lambda_{i j}\right)$.
For $y_{i j} =0 $ case, we also have the log link. Therefore,
$\log \mu_{i j}=\eta_{i j}$ and $\mu_{i j}=\exp (\eta_{i j})$.

Similarly, holding $\vw$ and $\vb$ fixed, the total $\log$ likelihood $\widetilde{\ell}$ for all columns is
\bqa
\widetilde{\ell}(\wtvw, \wtvb; \vw, \vb) % &=& \sum_{i=1}^{n} \sum_{j=1}^{n} \left\{\log f\left(y_{i j}\right) \cdot I\left(y_{i j}>0\right)+\log f_{0}\left(y_{i j}\right) \cdot I\left(y_{i j}=0\right)\right\} \\
&=& \sum_{j=1}^{n} \widetilde{\ell}_{j}^{(1)}(\wtvw_j, \wtb_j; \vw, \vb) + \sum_{i=1}^{n} \sum_{j=1}^{n} \widetilde{\ell}_{i j}^{(0)}(\wtvw_j, \wtb_j; \vw_i, b_i),
\eqa
where
\bqa
\widetilde{\ell}_{i j}^{(0)}(\wtvw_j, \wtb_j; \vw_i, b_i) = -\lambda_{i j}I\left(y_{i j}=0\right) \text{ and } \lambda_{i j}=\frac{\mu_{i j}^{2-p_{i j}}}{\phi_{i j}\left(2-p_{i j}\right)}.
\eqa

To derive partial derivatives for the case $y_{ij}>0$, note that
\bqa
&& \mu_{i j}=\kappa^{\prime}\left(\theta_{i j}\right) =\exp \left(\eta_{i j}\right) \quad \Rightarrow \frac{\partial \mu_{i j}}{\partial \eta_{i j}}=\exp \left(\eta_{i j}\right), \, \Var(Y_{ij})=\phi_{ij} \mu_{ij}^{p_{ij}}, \\
&& \eta_{i j}=\log \left(\mu_{i j}\right), \quad \frac{\partial \theta_{i j}}{\partial \mu_{i j}}=\frac{1}{\partial \mu_{i j} / \partial \theta_{i j}}=\frac{1}{\kappa^{\prime \prime}\left(\theta_{j}\right)}=\frac{1}{V_{i j}}=\frac{1}{\mu_{i j}^{p_{i j}}}.
\eqa
The first order derivatives of $i^{th}$ row log-likelihood and $j^{th}$ column log-likelihood which are components of the Score functions in the case $y_{ij}>0$ are
\bqa
\frac{\partial \ell_{i}^{(1)}}{\partial w_{i_1 k}}\!\!\!\!&&= \sum_{j=1}^n \frac{\partial \ell_{i j}^{(1)}}{\partial \theta_{i j}} \cdot \frac{\partial \theta_{i j}}{\partial \mu_{i j}} \cdot \frac{\partial \mu_{i j}}{\partial \eta_{i j}} \cdot \frac{\partial \eta_{i j}}{\partial w_{i_1 k}} = \sum_{j=1}^n \frac{y_{i j}-\mu_{i j}}{\phi_{i j}} \cdot \frac{1}{V_{i j}} \cdot e^{\eta_{i j}} \cdot \wtw_{j k} \cdot I(i=i_1) I(y_{i j}>0) \\ \!\!\!\!&&= \sum_{j=1}^n \frac{y_{i_1 j}-\mu_{i_1 j}}{\phi_{i_1 j}} \cdot \frac{1}{V_{i_1 j}} \cdot \mu_{i_1 j} \cdot \wtw_{j k} \cdot I(i=i_1) I(y_{i j}>0), \\
\frac{\partial \ell_{i}^{(1)}}{\partial b_{i_1}} \!\!\!\!&&= \sum_{j=1}^n \frac{\partial \ell_{i j}^{(1)}}{\partial \theta_{i j}} \cdot \frac{\partial \theta_{i j}}{\partial \mu_{i j}} \cdot \frac{\partial \mu_{i j}}{\partial \eta_{i j}} \cdot \frac{\partial \eta_{j}}{\partial b_{i_1}} = \sum_{j=1}^n  \frac{y_{i j}-\mu_{i j}}{\phi_{i j}} \cdot \frac{1}{V_{i j}} \cdot \mu_{ij} \cdot I(i=i_1) I(y_{i j}>0), \\
\frac{\partial \widetilde{\ell}_{j}^{(1)}}{\partial \wtw_{j_1 k}} \!\!\!\!&&= \sum_{i=1}^n  \frac{\partial \widetilde{\ell}_{i j}^{(1)}}{\partial \theta_{i j}} \cdot \frac{\partial \theta_{i j}}{\partial \mu_{i j}} \cdot \frac{\partial \mu_{i j}}{\partial \eta_{i j}} \cdot \frac{\partial \eta_{i j}}{\partial \wtw_{j_1 k}} = \sum_{i=1}^n \frac{y_{i j}-\mu_{i j}}{\phi_{i j}} \cdot \frac{1}{V_{i j}} \cdot \mu_{i j} \cdot w_{i k} \cdot I(j=j_1) I(y_{i j}>0), \\
\frac{\partial \widetilde{\ell}_{j}^{(1)}}{\partial \wtb_{j_1}} \!\!\!\!&&= \sum_{i=1}^n \frac{\partial \widetilde{\ell}_{i j}^{(1)}}{\partial \theta_{i j}} \cdot \frac{\partial \theta_{i j}}{\partial \mu_{i j}} \cdot \frac{\partial \mu_{i j}}{\partial \eta_{i j}} \cdot \frac{\partial \eta_{j}}{\partial \wtb_{j_1}} = \sum_{i=1}^n \frac{y_{i j}-\mu_{i j}}{\phi_{i j}} \cdot \frac{1}{V_{i j}} \cdot \mu_{ij} \cdot I(j=j_1) I(y_{i j}>0).
\eqa

The first order derivatives for $\ell_{i j}^{(0)}$ and $\widetilde{\ell}_{i j}^{(0)}$ are
\bqa
\frac{\partial \ell_{i j}^{(0)}}{\partial w_{i_1 k}} &=& -\frac{\partial \lambda_{i j}}{\partial \mu_{i j}}  \frac{\partial \mu_{i j}}{\partial \eta_{i j}}  \frac{\partial \eta_{i j}}{\partial w_{i_1 k}} = %-\frac{\mu_{i j}^{1-p_{i j}}}{\phi_{i j}}  \mu_{i j}  \widetilde{w}_{j k}  I(i=i_1) I(y_{i j}=0) =
-\frac{\mu_{i j}^{2-p_{i j}}}{\phi_{i j}}  \widetilde{w}_{j k}  I(i=i_1) I(y_{i j}=0), \\
\frac{\partial \ell_{i j}^{(0)}}{\partial b_{i_1}} &=& -\frac{\partial \lambda_{i j}}{\partial \mu_{i j}}  \frac{\partial \mu_{i j}}{\partial \eta_{i j}}  \frac{\partial \eta_{i j}}{\partial b_{i_1}} = %-\frac{\mu_{i j}^{1-p_{i j}}}{\phi_{i j}}  \mu_{i j}  I(i=i_1) I(y_{i j}=0) =
-\frac{\mu_{i j}^{2-p_{i j}}}{\phi_{i j}}  I(i=i_1) I(y_{i j}=0), \\
\frac{\partial \widetilde{\ell}_{i j}^{(0)}}{\partial \widetilde{w}_{j_1 k}} &=& -\frac{\partial \lambda_{i j}}{\partial \mu_{i j}}  \frac{\partial \mu_{i j}}{\partial \eta_{i j}}  \frac{\partial \eta_{i j}}{\partial \widetilde{w}_{j_1 k}} = %-\frac{\mu_{i j}^{1-p_{i j}}}{\phi_{i j}}  \mu_{i j}  w_{i k}  I(j=j_1) I(y_{i j}=0) =
-\frac{\mu_{i j}^{2-p_{i j}}}{\phi_{i j}}  w_{i k}  I(j=j_1) I(y_{i j}=0), \\
\frac{\partial \widetilde{\ell}_{i j}^{(0)}}{\partial \wtb_{j_1}} &=& -\frac{\partial \lambda_{i j}}{\partial \mu_{i j}}  \frac{\partial \mu_{i j}}{\partial \eta_{i j}}  \frac{\partial \eta_{i j}}{\partial \wtb_{j_1}} = %-\frac{\mu_{i j}^{1-p_{i j}}}{\phi_{i j}}  \mu_{i j}  I(j=j_1) I(y_{i j}=0) =
-\frac{\mu_{i j}^{2-p_{i j}}}{\phi_{i j}}  I(j=j_1) I(y_{i j}=0).
\eqa
Hence, the $k^{th}$ component, $k=1, \cdots, d$, of the score functions corresponding to $i^{th}$ row and $j^{th}$ column of the log likelihood are
\bqa
\left[U_{w_{i}}\right]_{k} &=& \frac{\partial \ell}{\partial w_{i k}} %= \sum_{i=1}^{n} \sum_{j=1}^{n} \frac{\partial \ell_{i j}}{\partial w_{i_1 k}},
= \sum_{j=1}^{n} \left\{\frac{y_{i j}-\mu_{i j}}{\phi_{i j}} \cdot \frac{1}{V_{i j}} \cdot \mu_{i j} \cdot \wtw_{j k} \cdot I\left(y_{i j}>0\right)+\frac{-\mu_{i j}^{2-p_{i j}}}{\phi_{i j}} \cdot \wtw_{j k} \cdot I\left(y_{i j}=0\right)\right\},
\\
\left[U_{\wtw_{j}}\right]_{k} &=& \frac{\partial \widetilde{\ell}}{\partial \wtw_{j k}} % = \sum_{i=1}^{n} \sum_{j=1}^{n} \frac{\partial \ell_{i j}}{\partial \widetilde{w}_{j_1 k}},
= \sum_{i=1}^{n} \left\{\frac{y_{i j}-\mu_{i j}}{\phi_{i j}} \cdot \frac{1}{V_{i j}} \cdot \mu_{i j} \cdot w_{i k} \cdot I\left(y_{i j}>0\right)+\frac{-\mu_{i j}^{2-p_{i j}}}{\phi_{i j}} \cdot w_{i k}\cdot I\left(y_{i j}=0\right)\right\}, \\
\text{and } \qquad && \\
U_{b_{i}} &=& \frac{\partial \ell}{\partial b_{i}} % = \sum_{i=1}^{n} \sum_{j=1}^{n} \frac{\partial \ell_{i j}}{\partial b_{i_1}}
= \sum_{j=1}^{n} \left\{\frac{y_{i j}-\mu_{i j}}{\phi_{i j}} \cdot \frac{1}{V_{i j}} \cdot \mu_{i j} \cdot I\left(y_{i j}>0\right)+\frac{-\mu_{i j}^{2-p_{i j}}}{\phi_{i j}} \cdot I\left(y_{i j}=0\right)\right\},
\\
U_{\wtb_{j}} &=& \frac{\partial \widetilde{\ell}}{\partial \wtb_{j}} % = \sum_{i=1}^{n} \sum_{j=1}^{n} \frac{\partial \ell_{i j}}{\partial \wtb_{j_1}} \\
= \sum_{i=1}^{n} \left\{\frac{y_{i j}-\mu_{i j}}{\phi_{i j}} \cdot \frac{1}{V_{i j}} \cdot \mu_{i j} \cdot I\left(y_{i j}>0\right)+\frac{-\mu_{i j}^{2-p_{i j}}}{\phi_{i j}} \cdot I\left(y_{i j}=0\right)\right\}.
\eqa
% Let $\vbeta_{i}=\left(\begin{array}{c} \vw_{i} \\ b_{i}\end{array}\right), \quad \wtvbeta_{j}=\left(\begin{array}{c}\wtvw_{j} \\ b_{j}\end{array}\right)$.
Let $\vbeta_i = (\vw_i^\top, b_i)^\top$ and $\wtvbeta_j = (\wtvw_j^\top, \wtb_j)^\top$.
From the partial derivatives of the log likelihoods, we can see that the second order partial derivatives with respect to $\vbeta_i$ and $\vbeta_{i_1}$ is zero when $i$ is different from $i_1$. That is, the cross terms $\frac{\partial^2 \ell}{\partial \vw_{i} \partial \vw_{i_1}} = 0$ and $\frac{\partial^2 \widetilde{\ell}}{\partial \wtvw_{i} \partial \wtvw_{i_1}} = 0$ and $\frac{\partial^2 \ell}{\partial \vw_{i} \partial b_{i_1}} = 0$ and $\frac{\partial^2 \widetilde{\ell}}{\partial \wtvw_{i} \partial \wtb_{i_1}} = 0$ when $i \neq i_1$. Therefore, the estimation of $\vbeta_i$ and $\vbeta_{i1}$ can be conducted separately.
Denote
\bqa
I_{\vbeta_i}=\left[\begin{array}{cc}
I_{\vw_i \vw_i} & I_{\vw_i b_i} \\
I_{b_i \vw_i} & I_{b_i b_i}
\end{array}\right], \quad I_{\wtvbeta_j}=\left[\begin{array}{cc}
I_{\wtvw_j \wtvw_j} & I_{\wtvw_j \wtb_j} \\
I_{\wtb_j \wtvw_j} & I_{\wtb_j \wtb_j}
\end{array}\right],
\eqa
where the individual components in the matrix $I_{\vbeta_i}$ and $I_{\wtvbeta_j}$ are negative expectations of the second order partial derivatives. We list them below and detailed derivation is omitted.
%can be found in the Appendix.
% \ref{Appendix:A}.

The $(k,r)^{th}$ entry, $k,r=1,\ldots,d$, of the Information matrix for $\vw_i$, $\wtvw_j$, $b_i$, and $\wtb_j$ respectively are
\bqa
\left[I_{\vw_{i_1} \vw_{i_2}}\right]_{k, r} &=& -E\left[\frac{\partial^{2} \ell}{\partial w_{i_1 k} \partial w_{i_2 r}}\right] = -\sum_{i=1}^{n} \sum_{j=1}^{n} E\left[\frac{\partial^{2}\left( \ell_{i j}^{(0)} + \ell_{i j}^{(1)}\right)}{\partial w_{i_1 k} \partial w_{i_2 r}}\right] \\
&=& \sum_{j=1}^{n} \frac{\mu_{i_1 j}^2}{\phi_{i_1 j} V_{i_1 j}} \cdot \wtw_{j k} \cdot \wtw_{j r} \cdot I(i_1=i_2) \cdot I\left(y_{i_1 j}>0\right) + \\
&& \sum_{j=1}^{n} \frac{\left(2-p_{i_1 j}\right)}{\phi_{i_1 j}} \cdot \mu_{i_1 j}^{2-p_{i_1 j}} \cdot \wtw_{j k} \cdot \wtw_{j r} \cdot I(i_1=i_2) \cdot I\left(y_{i_1 j}=0\right),
\\
\left[I_{\wtvw_{j_1} \wtvw_{j_2}}\right]_{k, r} &=& -E\left[\frac{\partial^{2} \widetilde{\ell}}{\partial \wtw_{j_1 k} \partial \wtw_{j_2 r}}\right] = -\sum_{i=1}^{n} \sum_{j=1}^{n} E\left[\frac{\partial^{2} \left(\widetilde{\ell}_{i j}^{(0)} + \widetilde{\ell}_{i j}^{(1)}\right)}{\partial \wtw_{j_1 k} \partial \wtw_{j_2 r}}\right] \\
&=& \sum_{i=1}^{n} \frac{\mu_{i j_1}^2}{\phi_{i j_1} V_{i j_1}} \cdot w_{i k} \cdot w_{i r} \cdot I(j_1=j_2) \cdot I\left(y_{i j_1}>0\right) + \\
&& \sum_{i=1}^{n} \frac{\left(2-p_{i j_1}\right)}{\phi_{i j_1}} \cdot \mu_{i j_1}^{2-p_{i j_1}} \cdot w_{i k} \cdot w_{i r} \cdot I(j_1=j_2) \cdot I\left(y_{i j_1}=0\right),
\eqa
\bqa
I_{b_{i_1} b_{i_2}} &=& -E\left[\frac{\partial^{2} \ell}{\partial b_{i_1} \partial b_{i_2}}\right] = -\sum_{i=1}^{n} \sum_{j=1}^{n} E\left[\frac{\partial^{2} \left( \ell_{i j}^{(0)} + \ell_{i j}^{(1)}\right)}{\partial b_{i_1} \partial b_{i_2}}\right] \\
&=& \sum_{j=1}^{n} \frac{\mu_{i_1 j}^2}{\phi_{i_1 j} V_{i_1 j}} \cdot I(i_1=i_2) \cdot I\left(y_{i_1 j}>0\right) + \sum_{j=1}^{n} \frac{\left(2-p_{i_1 j}\right)}{\phi_{i_1 j}} \cdot \mu_{i_1 j}^{2-p_{i_1 j}} \cdot I(i_1=i_2) \cdot I\left(y_{i_1 j}=0\right), \\
I_{\wtb_{j_1} \wtb_{j_2}} &=& -E\left[\frac{\partial^{2} \widetilde{\ell}}{\partial b_{j_1} \partial b_{j_2}}\right] = -\sum_{i=1}^{n} \sum_{j=1}^{n} E\left[\frac{\partial^{2}\left( \widetilde{\ell}_{i j}^{(0)} + \widetilde{\ell}_{i j}^{(1)}\right)}{\partial b_{j_1} \partial b_{j_2} }\right] \\
&=& \sum_{i=1}^{n} \frac{\mu_{i j_1}^2}{\phi_{i j_1} V_{i j_1}} \cdot I(j_1=j_2) \cdot I\left(y_{i j_1}>0\right) + \sum_{i=1}^{n} \frac{\left(2-p_{i j_1}\right)}{\phi_{i j_1}} \cdot \mu_{i j_1}^{2-p_{i j_1}}  \cdot I(j_1=j_2) \cdot I\left(y_{i j_1}=0\right).
\eqa
The off diagonal components of $I_{\vbeta_i}$, $I_{\wtvbeta_j}$ are
\bqa
&& \left[I_{\vw_{i_1} b_{i_2}}\right]_k = -E\left[\frac{\partial^{2} \ell}{\partial w_{i_1 k} \partial b_{i_2}}\right] = -\sum_{i=1}^{n} \sum_{j=1}^{n} E\left[\frac{\partial^2 \left( \ell_{i j}^{(0)} + \ell_{i j}^{(1)}\right)}{\partial w_{i_1 k} \partial b_{i_2}}\right] \\
&=& \left\{\sum_{j=1}^{n} \frac{\mu_{i_1 j}^2}{\phi_{i_1 j}} \cdot \frac{\widetilde{w}_{j k}}{V_{i_1 j}} \cdot I\left(y_{i_1 j}>0\right) + \sum_{j=1}^{n} \frac{\left(2-p_{i_1 j}\right)}{\phi_{i_1 j}} \cdot \mu_{i_1 j}^{2-p_{i_1 j}} \wtw_{j k} \cdot I\left(y_{i_1 j}=0\right)\right\} \cdot I(i_1=i_2),
\eqa
and
\bqa
&& \left[I_{\wtvw_{j_1} \wtb_{j_2}}\right]_k = -E\left[\frac{\partial^{2} \widetilde{\ell}}{\partial \wtw_{j_1 k} \partial \wtb_{j_2}}\right] = -\sum_{i=1}^{n} \sum_{j=1}^{n} E\left[\frac{\partial^2 \left( \widetilde{\ell}_{i j}^{(0)} + \widetilde{\ell}_{i j}^{(1)}\right)}{\partial \wtw_{j_1 k} \partial \wtb_{j_2}}\right] \\
&=& \left\{\sum_{i=1}^{n} \frac{\mu_{i j_1}^2}{\phi_{i j_1}} \cdot \frac{w_{i k}}{V_{i j_1}} \cdot I\left(y_{i j_1}>0\right) + \sum_{i=1}^{n} \frac{\left(2-p_{i j_1}\right)}{\phi_{i j_1}} \cdot \mu_{i j_1}^{2-p_{i j_1}} w_{i k} \cdot I\left(y_{i j_1}=0\right) \right\}\cdot I(j_1=j_2).
\eqa
Note that the components $-E\left[\frac{\partial^2 \ell}{\partial w_{i k} \partial b_{i}}\right]$ are non-zero, so we can not separately estimate $\vw_i$ and $b_i$. They have to be estimated together. Similarly, $\wtvw_j$ and $\wtb_j$ have to be estimated together.

Based on the Fisher scoring algorithm, the updating formulae for $\vbeta_i = (\vw_i^\top, b_i)^\top$ and $\wtvbeta_j = (\wtvw_j^\top, \wtb_j)^\top$ are respectively,
\bqan
\label{MLE:exactupdate}
%\begin{aligned}
\vbeta_{i}^{(t+1)} = \vbeta_{i}^{(t)}+I_{\vbeta_{i}^{(t)}}^{-1} \cdot U_{\vbeta_{i}^{(t)}}, \qquad
\wtvbeta_{j}^{(t+1)} = \wtvbeta_{j}^{(t)}+I_{\wtvbeta_{j}^{(t)}}^{-1} \cdot U_{\wtvbeta_{j}^{(t)}}.
%\end{aligned}
\eqan
Note that $\vbeta_i$ and $\wtvbeta_j$ can not be estimated simultaneously because estimation of $\vbeta_i$ requires to use the entire set of $\{\wtvbeta_j, j=1,\ldots,n\}$ as fixed values. Reversely, $\{\vbeta_i, i=1,\ldots,n\}$ serve as data when estimating $\wtvbeta_j$. Since the estimate of $\vbeta_i$ relies on the current value of $\wtvbeta$ which is not necessarily equal to the true value of the parameter, the $\vbeta_i$'s estimate is not final even if the first updating equation in (\ref{MLE:exactupdate}) converged while the same value of $\wtvbeta$ is used. When $\wtvbeta$ changes in further iterations, the estimate of $\vbeta_i$ will change too. Running either of equations in (\ref{MLE:exactupdate}) is generally referred as the iteratively reweighted least squares (IRLS) algorithm. In our case, we have to alternatingly rerun IRLS update by switching what is held fixed. In addition, since all the $\wtvbeta$ are needed when estimating $\vbeta_i$ but the IRLS update only works with one component at a time, $\wtvbeta_j$, the process internally depends on each other when we loop through $i$ or $j$ over different iterations. Below we will describe the algorithm for the estimating process.

The algorithm starts with setting the following initial values:
\bit
\item Specify values of $\phi$ and $p$. \Comment{(see Section \ref{subsection:impact-of-p-phi})}
\item Initialize elements of $\vbeta_i^{(0)}$ and $\wtvbeta_i^{(0)}$. For example, with i.i.d.  Uniform(-0.5, 0.5) values.
\item It is natural to use the negative log likelihood as the loss for fitting each row and column. However, the log likelihood functions $\ell$ and $\widetilde{\ell}$ both contain the $c(y_{i j}, \phi_{i j}, p_{i j})$. This term does not involve regression parameters $\vbeta$ and $\wtvbeta$. Therefore, it is redundant to compute the term $c(y_{i j}, \phi_{i j}, p_{i j})$ when estimating $\vbeta$ and $\wtvbeta$. This is specially the case for Tweedie regression because $c(y_{i j}, \phi_{i j}, p_{i j})$ has log sum of infinite many terms. In Python package sklearn.linear\_model.TweedieRegressor, the loss is defined as Half Tweedie deviance $max(y_i, 0)^{2-p} / (1-p) / (2-p) - y_i  \mu_i^{1-p} / (1-p) + \mu_i^{2-p} / (2-p)$ when the observed response is $y_i$ (see the Python source code at \url{https://github.com/scikit-learn/scikit-learn/blob/36958fb240fbe435673a9e3c52e769f01f36bec0/sklearn/_loss/loss.py}). The first term in it does not depend on the regression parameter $\beta$. The other two terms are the negative value of $(y\theta - \kappa(\theta))$ in the log likelihood (\ref{formula:Tweedie-lli}). Therefore, using the Half Tweedie deviance as the loss is equivalent to using the -$(y\theta - \kappa(\theta))$. We define our loss as follows:
\bqa \label{our_tweedie_loss}
Loss(\vbeta, \wtvbeta) = -\sum_{i=1}^n \sum_{j=1}^n (y_{i j}\theta_{i j} - \kappa(\theta_{i j})).
\eqa

% scikit-learn
% class HalfTweedieLoss(BaseLoss):
%     """Half Tweedie deviance loss with log-link, for regression.
% url = https://github.com/scikit-learn/scikit-learn/blob/36958fb240fbe435673a9e3c52e769f01f36bec0/sklearn/_loss/loss.py
%  loss(x_i) = max(y_true_i, 0)**(2-p) / (1-p) / (2-p)
%                     - y_true_i * exp(raw_prediction_i)**(1-p) / (1-p)
%                     + exp(raw_prediction_i)**(2-p) / (2-p)

% \bqa
% loss_i = -\ell_{i}, \quad \widetilde{loss}_j = -\widetilde{\ell}_{j},
% \eqa
% where $\ell_{i}$ and $\widetilde{\ell}_{j}$ are given in (\ref{formula:Tweedie-lli}) and (\ref{formula:Tweedie-lltildej}).
% The overall loss is the average of total loss from all rows and columns:
% \bqa
% Loss = (\sum_i loss_i +  \sum_j \widetilde{loss}_j) / 2.
% \eqa
Note that the loss is a function of $\vbeta$ and $\wtvbeta$, which will change as the algorithm proceeds.
\item Set convergence threshold such as $\epsilon = 10^{-4}$ and maximum number of iterations ($maxit$). We use relative convergence criteria on the loss function by checking if the relative change in loss defined below  % magnitude of change in the loss function compared to the absolute value of current loss function
is less than the threshold. This is consistent with $glm2$ package in R which uses the relative convergence criteria on the log likelihood.
\bqan
\label{formula:relative-change-loss}
\text{Relative change in Loss } = \frac{|Loss(\vbeta^{(t+1)}, \wtvbeta^{(t+1)}) - Loss(\vbeta^{(t)}, \wtvbeta^{(t)})|}{|Loss(\vbeta^{(t+1)}, \wtvbeta^{(t+1)})| + 0.1}.
\eqan
\eit

\begin{algorithm}
\caption{Alternating Tweedie regression}
\label{algorithm:AlternatingTweedie}
\begin{algorithmic}
\Require $I_{\vbeta_{i}^{(t)}}^{-1}$ and $I_{\wtvbeta_{j}^{(t)}}^{-1}$ exist.
\State

\State $t \gets 0$, $Loss^{(t)} \gets 10^{20}$
\State $converged = False$
\While{$t \leq maxit$}
%\While{$\sum_{i=1}^{n} || \vbeta_{i}^{(t+1)} - \vbeta_{i}^{(t)} || \ge \epsilon$ or $\sum_{i=1}^{n} || \wtvbeta_{i}^{(t+1)} - \wtvbeta_{i}^{(t)} || \ge \epsilon$}
%
\While{$converged = False$}
\State $i \gets 1$
\While{$i \leq n$}
\State retrieve $i$th row of data from SQLite database.
\For{\texttt{epoch $\in$ 1,\ldots, n\_epoch}}
\State compute $U_{\vbeta_i^{(t)}}$ and $I_{\vbeta_i^{(t)}}$.
\State update $\vbeta_i^{(t+1)}$ using current value of \{ $\wtvbeta_{j}^{(t)}, j\!=\!1, \ldots, n$ \}, and $\vbeta_i^{(t)}$
\State based on formulae in (\ref{MLE:exactupdate}) or (\ref{MLE:lrupdate}) or Adam update.
\EndFor
\State
% \State $i \gets i+1$
% \EndWhile
% \State $j \gets 1$
% \While{$j \leq n$}
\For{\texttt{epoch $\in$ 1,\ldots, n\_epoch}}
\State compute $U_{\wtvbeta_i^{(t)}}$ and $I_{\wtvbeta_i^{(t)}}$.
\State update $\wtvbeta_i^{(t+1)}$ using current value of \{ $\vbeta_{j}^{(t)}, j\!=\!1, \ldots, n$ \}, and $\wtvbeta_i^{(t)}$
\State based on formulae in (\ref{MLE:exactupdate}) or (\ref{MLE:lrupdate}) or Adam update.
\EndFor
\State $i \gets i+1$
\EndWhile
\State
\For{$k=1, \ldots, n$}
\State retrieve $k$th row of data from SQLite database.
\State recompute $U_{\vbeta_k^{(t+1)}}$,  $U_{\wtvbeta_k^{(t+1)}}$ and their $L_2$ norms using $\vbeta^{(t+1)}$ and $\wtvbeta^{(t+1)}$ values.
\State recompute $loss_k$ and $\widetilde{loss}_k$ using $\vbeta^{(t+1)}$ and $\wtvbeta^{(t+1)}$ values.
\EndFor
\State compute the $Loss(\vbeta^{(t+1)}, \wtvbeta^{(t+1)})$.
\State check if the Relative change in Loss (\ref{formula:relative-change-loss}) is less than $\epsilon$.
% \State check if the condition $converged = \mid Loss^{(t+1)} - Loss^{(t)} \mid / (|Loss^{(t+1)}| + 0.1) < \epsilon$
% \State is True or False.
\EndWhile
\State $t \gets t+1$
\EndWhile
% \State Stabilized $\vbeta$ and $\wtvbeta$ lead to globally or locally optimized value of the loss function of the Alternating Tweedie regression.
\end{algorithmic}
\end{algorithm}

When the algorithm stops, either the maximum number of iterations is reached or the algorithm converged. If $t$ is less than $maxit$, the algorithm has converged. With our loss definition, we are making an assumption that $\phi$ and $p$ are not involved in likelihood comparison. This makes sense if the likelihood function is evaluated at the same set of $\phi$ and $p$ values. If the likelihood has to be compared with different $\phi$ and $p$ pairs, the likelihood corresponds to different pairs can not be compared because the infinite summation in (\ref{function:c_fct_1}), (\ref{function:c_fct_2}), (\ref{function:c_fct_3}) are different. %The convergence criteria we used seems to be reasonable because our algorithm uses the Fisher scoring algorithm which does not involve derivative with respect to $\phi$ and $p$.

% Here is an outline of the algorithm.
% \bit
% \item[1.] For given values of $\phi$ and $p$, define the log likelihood function of alternating Tweedie regression as the objective function.
% \item[2.] For each $i$, update $\vbeta_i^{(t+1)}$ using current value of $\wtvbeta_{j}^{(t)}, \quad j=1, \ldots, n$, and $\vbeta_i^{(t)}$.
% \item[3.] Repeat step 2 for all $i=1, \ldots, n$.
% \item[4.] For each $i$, update $\wtvbeta_i^{(t+1)}$ using current value of $\vbeta_{j}^{(t)}, \quad j=1, \ldots, n$, and $\wtvbeta_i^{(t)}$.
% \item[5.] Repeat step 4 for all $i=1, \ldots, n$.
% \item[6.] Check whether the algorithm converged or not by comparing the convergence threshold with the $L_2$ norm of the change in $\vbeta$ and $\wtvbeta$ from successive iterations. That is, whether $\sum_{i=1}^{n} || \vbeta_{i}^{(t+1)} - \vbeta_{i}^{(t)} || < \epsilon$ and $\sum_{i=1}^{n} || \wtvbeta_{i}^{(t+1)} - \wtvbeta_{i}^{(t)} || < \epsilon$ where $\epsilon = 10^{-4}$ is the convergence threshold.

Also, note that the updating formula (\ref{MLE:exactupdate}) does not have a learning rate. If the model is completely correct, this updating formula will lead to good estimate of $\vbeta$ and $\wtvbeta$. However, if there is any disturbance or model misspecification, it is likely that $I^{-1} U$ might be updating the estimated $\vbeta$ or $\wtvbeta$ in a wrong direction. This is particularly prone to happen in the case that the dimension of the parameters is large. Therefore, we would like to consider adaptive update such that the pace of update is reducing as the number of iterations increases. Specifically, we introduce a learning rate and adjust it as the iteration number grows in the following manner:
\bqan
\label{MLE:lrupdate}
\begin{aligned}
\vbeta_{i}^{(t+1)} &= \vbeta_{i}^{(t)}+ \frac{lr}{t^{1/4}} \cdot I_{\vbeta_{i}^{(t)}}^{-1} \cdot U_{\vbeta_{i}^{(t)}}, \qquad t = 1, 2, \ldots , \\
\wtvbeta_{j}^{(t+1)} &= \wtvbeta_{j}^{(t)}+ \frac{lr}{t^{1/4}} \cdot I_{\wtvbeta_{j}^{(t)}}^{-1} \cdot U_{\wtvbeta_{j}^{(t)}}, \qquad t = 1, 2, \ldots.
\end{aligned}
\eqan
where $lr$ is a starting learning rate that could be set to $lr=0.5$. As the iterations increase, the effective learning rate becomes $\frac{lr}{t^{1/4}}$, which allows the algorithm to slowly approach the target in later iterations.

% https://ruder.io/optimizing-gradient-descent/index.html#hogwild
Our learning rate adjustment is just a step size change in the Fisher scoring algorithm. The $glm2$ package in R halves the step size when estimated model components become infinity or out of their range or the algorithm diverges. As the Fisher scoring algorithm belongs to a bigger category of gradient descent/ascent algorithm, we can look more into the bigger category.  There are many variants of the gradient descent adjustments %. See for example \cite{Ruder:2016} for review of different gradient descent algorithms
such as Momentum, Adagrad, Adadelta, RMSprop, Adam etc. The Adam is the most popular update and becomes default parameter update method for many machine learning algorithms. With Adam, the direction of update is set to be the ratio of two terms: the exponentially decaying moving average of past gradients in the numerator, and the square root of exponentially decaying moving average of past squared gradients in the denominator. Adam combines the ideas of Momentum and and Adadelta. The moving averages of the gradient and squared gradient reflect the mean and variance of the gradient. The update is defined as follows:
$$
\begin{gathered}
g_{\mathrm{t}, \mathrm{j}}=\nabla_\theta \mathrm{\ell}\left(\theta_{\mathrm{t}-1}\right)_{\mathrm{j}} ; \quad \mathrm{m}_{\mathrm{t}, \mathrm{j}}=\mathrm{B}_1 \mathrm{~m}_{\mathrm{t}-1, \mathrm{j}}+\left(1-\mathrm{B}_1\right) g_{\mathrm{t}, \mathrm{j}}, \quad v_{\mathrm{t}, \mathrm{j}}=\mathrm{B}_2 v_{\mathrm{t}-1, \mathrm{j}}+\left(1-\mathrm{B}_2\right) g_{\mathrm{t}, \mathrm{j}}^2 \\
\mathrm{m}_{0 j}=v_{0 j}=0 ; \quad \hat{\mathrm{m}}_{\mathrm{t}, \mathrm{j}}=\frac{\mathrm{m}_{\mathfrak{t}, \mathrm{j}}}{1-\mathrm{B}_1^{\mathfrak{t}}} ; \quad \hat{v}_{\mathrm{t}, \mathrm{j}}=\frac{v_{\mathrm{t}, \mathrm{j}}}{1-\mathrm{B}_2^t} ; \quad \theta_{\mathrm{t}, \mathrm{j}}=\theta_{\mathrm{t}-1, \mathrm{j}}-\frac{\eta}{\sqrt{\hat{v}_{\mathrm{t}, \mathrm{j}}+\epsilon}} \hat{\mathrm{m}}_{\mathrm{t}, \mathrm{j}}
\end{gathered}
$$
The authors of this method recommended to set $B_1=0.9, B_2=0.999, \epsilon=10^{-8}$. The second row of the equations makes corrections for total coefficients because the coefficient behind $m_{0, j}$ is $B_1^t$ and $m_{0, j}=0$, which lead to the sum of coefficients of non-zero terms being $1-B_1^t$. The same justification holds for $v_{t, j}$.
This update has a step size that takes the steepness into account, as in Adadelta, but also tends to move in the same direction, as in Momentum.

\begin{figure}
    \centering
    \begin{tabular}{cc}
        \includegraphics[width=0.45\linewidth]{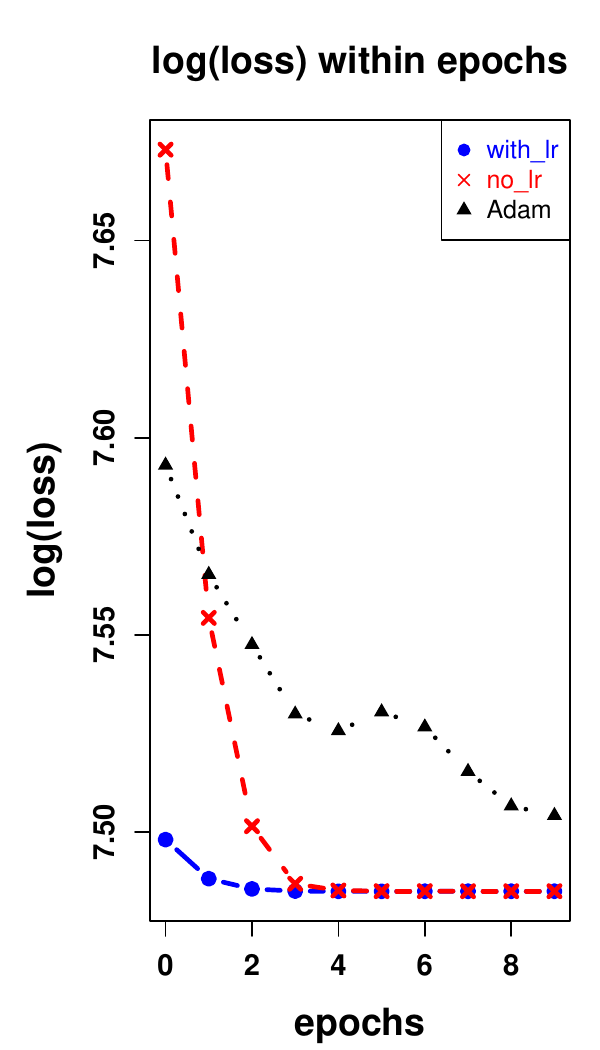} &
        \includegraphics[width=0.45\linewidth]{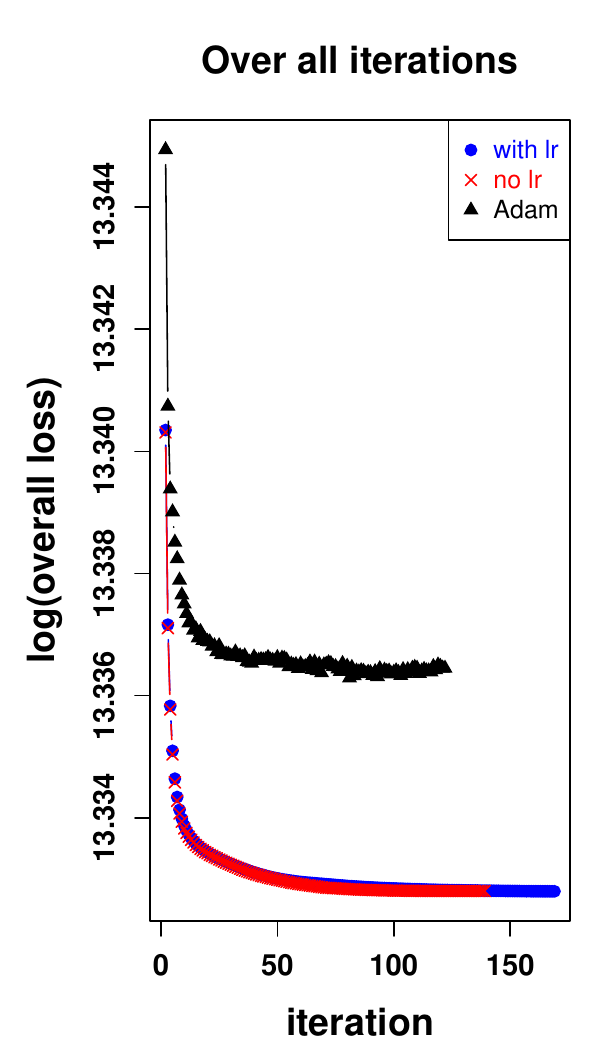}
    \end{tabular}
    \caption{\small Computed log(loss) and log(overall loss) from simulated dataset using the Fisher scoring with or without learning rate adjustment, and gradient descent algorithm with Adam method for parameter update. The left panel depicts how the loss changes over 10 epochs for one row of the parameter update. As epoch number grows, the loss has a general decreasing trend but the Adam's loss has higher values and reduces slower than the other two updates. The right panel is for overall loss versus number of iterations in $\log$ scale. All losses decrease as the iteration number increases but the Adam update has higher values of the overall loss.}
    \label{fig:AdamLossPlot}
\end{figure}

Most of the gradient descent variants do not use the second derivative of the objective function, which is because it is often difficult to compute the second derivative. We believe the use of adaptive learning rate together with the Fisher information matrix in our algorithm provides more benefit than the variants of the gradient descent algorithms. This is because the variants of gradient descent algorithm either ignore the second derivative or uses squared first derivative to approximate the second derivative. On the other hand, the Fisher information provides the variance of the first derivative (i.e., Score function) of the objective function and the inverse of the Fisher information matrix provides the asymptotic variance for the parameters to be estimated according to likelihood inference. % (\citealt{Cox-Hinkley:2017}).
The plots in Figure \ref{fig:AdamLossPlot} shows how the three updating schemes differ when they are applied to a simulated dataset generated with alternating Tweedie regression model (see section \ref{section:TweedieMLEsimulation}). We could see from the left panel which shows the log(loss) for only one row of data with repeating update over 10 epochs, the Adam update is less effective in reducing the loss for each epoch compared to the update with or without learning rate adjustment. When the algorithm continuously updates for over 100 iterations, the log(overall loss) in the right panel also shows that the Adam update is not reducing the loss as fast as the other two updating schemes.
Avoiding the usage of second derivatives of the objective function not only shows no advantage in computation time but also is less effective in reducing the loss. Deriving the second derivatives and implementing it in the Fisher scoring algorithm achieves better result faster.

% The figure above was created from following R file
% /Users/taejoonkim/Research_2/2022_06_15_simultaneous_work_of_CPU_and_GPU/create_table_on_Latex_from_Excel.R
% used simul\_4.sqlite db

Note that in the while loop for $i$ in Algorithm \ref{algorithm:AlternatingTweedie}, we update $\vbeta_i$ and $\wtvbeta_i$ for the same $i$. The $\vbeta_i$'s estimate depends on current value of $\wtvbeta$ which means recently updated value of $\wtvbeta_{j_1}, j_1=1,\ldots,i$ are used in the calculation. Due to such dependence, all parameter updates rely on each other. As a result, the updates of different parameters can only be done sequentially. In addition, it assumes that the data is symmetric so that the $i^{th}$ row and $i^{th}$ column are same. An advantage of this is that the $i^{th}$ row of data is only loaded once. This saves some computation time. If the data matrix is not symmetric, then the algorithm needs to be adjusted by adding another for loop of $j$ such that update for $\vbeta_i, i=1,\ldots,n$ is done consecutively followed by update of $\wtvbeta_j, j=1, \ldots, n$. In this case, we need to read-in data for each row while updating the $\vbeta_i$'s and then we will need to read-in data again for updating $\wtvbeta_j$. This potentially wastes some computational time. However, there is some potential benefit in doing so: successive update for different $i$'s (or different $j$'s) are not interdependent on each other. This makes parallel updating a possible solution. More specifically, when considering $\vbeta_i$'s as the parameter while holding $\{ \wtvbeta_j, \quad j=1, \ldots, n \}$ as the data, the update for different $i$'s is totally free from each other since the calculation of the Information matrix and the Score function only depends on $\{ \wtvbeta_j, \quad j=1, \ldots, n \}$.

It is tempting to use an existing parallel computing algorithm to conduct the update. %\cite{Niu-et-al:2011} introduced an update scheme called
Hogwild update scheme is an example that allows performing parameter update on CPUs in parallel. It works with multiple CPU processors by allowing them to access shared memory without locking the parameters. For this to work, the optimization problems have to be sparse in that each update only modifies a fraction of all parameters. Under this scenario, they showed that the Hogwild update scheme achieves almost an optimal rate of convergence because the possibility of the processors overwrite useful information is small. Our updating equations (\ref{MLE:lrupdate}) only modify part of the model parameters as $i$ or $j$ points to specific row or column. However, that is not a sparse update. Even though we are working with only a block out of all model parameters at a time, there is no guarantee that the parameter update will be sparse. In fact, the parameters in our model are dense representation of words, items, or users, and they are not meant to be sparse. We will explain computational strategies in Section \ref{large_scale_app}.
% Additionally, our computations need to be carried out on GPUs. This point will be further explained in Chapter \ref{chapter:5}.

\subsection{Impact of the parameters $p$ and $\phi$}
\label{subsection:impact-of-p-phi}

The aforementioned formulae assume $p_{i j}$ and $\phi_{i j}$ are known. When they are not known, estimation is needed. The estimation of the dispersion parameters $p_{i j}$ and $\phi_{i j}$ is intractable when we use the likelihood approach.

\bqa
\frac{\partial \ell_i^{(1)}}{\partial \phi_{i j}} &=& \sum_{j=1}^{n}-\frac{1}{\phi_{i j}^{2}}\left(y_{i j} \theta_{i j}-\kappa\left(\theta_{i j}\right)\right)+\frac{\partial c\left(y_{i j}, \phi_{i j}, p_{i j}\right)}{\partial \phi_{i j}}, \\
\frac{\partial \ell_i^{(1)}}{\partial p_{i j}} &=& \sum_{j=1}^{n} \frac{1}{\phi_{i j}}\left[y_{i j} \frac{\partial \theta_{i j}}{\partial p_{i j}}+\frac{\partial \kappa\left(\theta_{i j}\right)}{\partial p_{i j}}\right]+\frac{\partial c\left(y_{i j}, \phi_{i j},  p_{i j}\right)}{\partial p_{i j}}
\eqa
where $c\left(y, \phi, p\right)=\log \left(\sum_{k=1}^{\infty} \frac{\beta^{k \alpha}}{\Gamma(k \alpha)} \cdot y^{k \alpha-1} \cdot \frac{\lambda^{k}}{k !}\right)$. And $\phi$ and $p$ are related in the following manner:
$$
\lambda=\frac{\mu^{2-p}}{\phi(2-p)} \quad, \quad \alpha=\frac{2-p}{p-1} \quad, \quad \frac{1}{\beta}=\phi(p-1) \mu^{p-1}.
$$
Due to excessive calculation involved in the infinite sum, the likelihood estimate of $\phi_{i j}$ and $p_{i j}$ is intractable when they need to be estimated for all $i=1, ..., n$ and $j=1, ..., n$. \cite{Bonat-Kokonendji:2017} suggested to use profile likelihood to estimate the $\phi$ and $p$. In their small scale simulation studies, they find that the profile likelihood approach works well, particularly when true $p$ is 0 or 2 in which case the likelihood approach fails to provide appropriate estimate with good coverage. However, the profile likelihood uses derivative free Nelder-Mead algorithm, %(\citealt{Nelder-Mead:1965}).
which is really slow for large $n$.

For estimating power p from Tweedie model, traditional practice is to first train a Tweedie model with an arbitrary specified p and estimate the $\mu$ of all observations. The inverse weight and residuals from the model fit were then used to obtain an estimate of dispersion parameter (scale). Then the $\mu$ along with inverse weight, residuals, and scale estimate ($\phi$) were used to estimate $p$. See the source code of function estimated\_tweedie\_power at \url{https://www.statsmodels.org/devel/_modules/statsmodels/genmod/generalized_linear_model.html#GLM.estimate_tweedie_power}.
%{\scriptsize \url{ https://www.statsmodels.org/devel/_modules/statsmodels/genmod/generalized_linear_model.html#GLM.estimate_tweedie_power}}.
In our case, we have no covariates to fit the GLM model. The $\wtvbeta$ and $\vbeta$ were alternately updated so we can not use them as the covariates to estimate the power $p$ even when they were held fixed because the estimate of $p$ will change with the estimate of $\vbeta$ and $\wtvbeta$ and the loss functions correspond to different range of $p$ are no longer comparable.

\begin{figure}[H]
    \centering
    \includegraphics[width=0.7\textwidth]{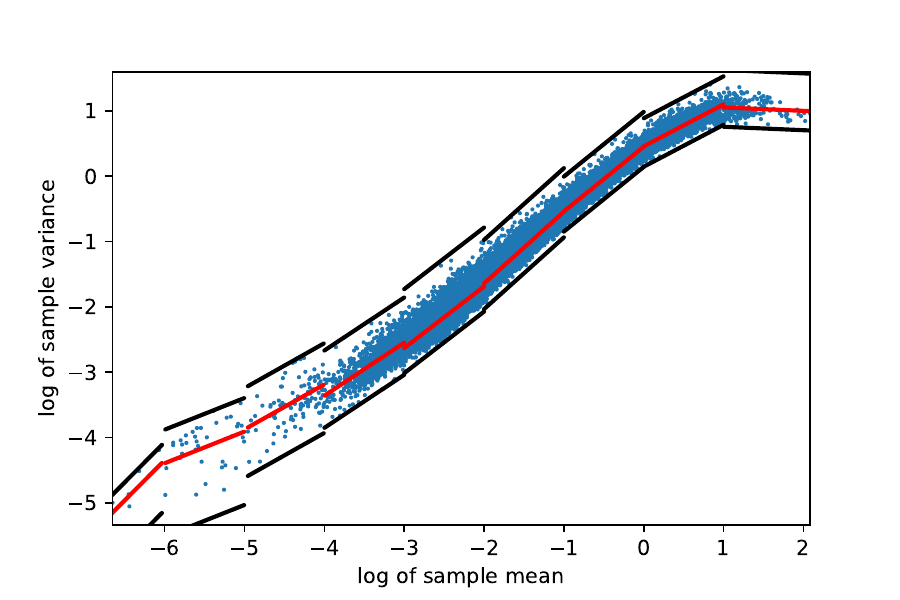}
    \vspace{-0.3in}
    \caption{Relationship between log of sample mean and log of sample variance from Wikipedia data with 50K vocabulary size. The three lines in each interval are the fitted linear regression line and upper and lower bounds with same slope.
    %in each interval are lines having same slope as the fitted line but passing through the data point that is having maximum or minimum signed distance from the fitted line.
    }
    \label{fig.log_mu.log_var}
\end{figure}

Instead, we will make use of the mean variance relationship $\Var(Y) = \phi \mu^p$ for the Tweedie distribution. That is, $\log(\Var(Y)) = \log\phi + p\log\mu$. Then we check the log of mean and log of variance of the observed response variable over different rows. The two appears to have piece-wise linear relationship. See Figure \ref{fig.log_mu.log_var} for an example, which was generated for the January, 2022 Wikipedia dump with vocabulary size 50K. The majority of $\log\mu$ and $\log(\Var(Y))$ pairs shows clear piecewise linear relationship when $\log\mu$ is between -4 and 2. For log mean smaller than -4 or greater than 1, there are limited number of data points. Based on such piecewise linear relationship, we regressed the log of sample variance against the log of sample mean of the response variable on each interval. The resulting slope and intercept estimates presented in Table \ref{table:V50K_delta_p_table} give us estimates of $p$ and $\delta = \log(\phi)$ for each interval shown in Figure \ref{fig.log_mu.log_var}.

\begin{table}[H]
\centering
\begin{tabular}{rrrrrrr}
  \hline
interval index & $\widehat{\delta}$ & $\widehat{\delta}_{high}$ & $\widehat{\delta}_{low}$ & $\log(\mu)$ lower bound & $\log(\mu)$ upper bound & $\widehat{p}$ \\
  \hline
   0 & 3.030 & 3.306 & 2.264 & -7.000 & -6.000 & 1.230 \\
     1 & -1.483 & -0.970 & -2.607 & -6.000 & -5.000 & 0.485 \\
     2 & -0.438 & 0.194 & -1.180 & -5.000 & -4.000 & 0.688 \\
     3 & -0.115 & 0.576 & -0.607 & -4.000 & -3.000 & 0.811 \\
     4 & 0.197 & 1.099 & -0.188 & -3.000 & -2.000 & 0.943 \\
     5 & 0.554 & 1.221 & 0.161 & -2.000 & -1.000 & 1.098 \\
     6 & 0.451 & 0.984 & 0.142 & -1.000 & 0.000 & 0.990 \\
     7 & 0.457 & 0.887 & 0.145 & 0.000 & 1.000 & 0.642 \\
     8 & 1.105 & 1.680 & 0.809 & 1.000 & 2.582 & -0.053 \\
   \hline
\end{tabular}
\caption{The estimated $\delta$ and $p$ from fitting linear regression on log of sample mean and log of sample variance on each interval. The $\widehat{\delta}_{high}$ and $\widehat{\delta}_{low}$ are intercepts from the data points that are having maximum or minimum signed distance from the fitted regression line while having corresponding same slope.}
\label{table:V50K_delta_p_table}
\end{table}
In Table \ref{table:V50K_delta_p_table}, we have to carefully consider the estimated parameter $\hat{p}$. The range of $p$ for Tweedie distribution to be well defined is $p \leq 0$, $1 \le p < 2$, or $p \ge 2$, with $p=0$ represents Gaussian distribution, $p=1$ represents Poisson distribution, and $p=2$ represents Gamma distribution. The interval 0 having $\hat{p}$ between 1 and 2 corresponds to Compound Poisson Gamma distribution. For the interval indices 4, 5, and 6, those could be considered as Poisson distribution having $p=1$. The interval 8 could be viewed as Normal distribution with $p=0$. On the other hand, the intervals 1, 2, 3, and 7 have $\hat{p}$ between 0 and 1 where Tweedie distribution is not well defined. Therefore, we might need to consider other distribution such as zero-inflated Tweedie distribution.
\label{MLE:Tweedie}

%\section{Numerical studies }
%\input{application_reuters}
\section{A small simulation study}
\label{small_numerical}
\label{section:TweedieMLEsimulation}
In this section, we present a small simulation study using generated data from the Tweedie regression model. The goal of this study is to examine how fast the algorithms converge. Specifically, we would like to see how well the Fisher scoring algorithm with and without learning rate adjustment and the gradient descent algorithm with Adam update model parameters. The data generation is described below.
\bit
\item We first retrieved the most frequently used 300 words in 2000 Reuters business news articles. For each word, a 50-dim vector ($\vw$) needs to be generated for usage in the Tweedie regression model. The model components are as follows:
\bqa
\eta_{i j} = \vw_i^{\top} \cdot \wtvw_j, \quad \mu_{i j} = \exp{\eta_{i j}}
\eqa
where $\vw_i = \vg_i/|\vg_i|$ and $\wtvw_j = \vg_j/|\vg_j|$ with $\vg_i$ and $\vg_j$ being the 50-dim word vector representations from pre-trained GloVe model for word$_i$ and word$_j$ respectively (\url{https://nlp.stanford.edu/data/glove.6B.zip}). Note that these $\vw_i$'s can be arbitrarily generated from a distribution. Here, we use the GloVe word vectors so that the simulation is reproducible. Below are details:
\item The Tweedie power parameter $p_{i j}$ was generated as $p_i + p_j$ where $p_i$ and $p_j$ are independently sampled from Uniform(0.5, 1). With this generation mechanism, the power $p_{i j}$ is between 1 and 2 so that the count data follows Compound Poisson Gamma distribution.
\item The dispersion parameter $\phi_{i j}$ was generated as $\phi_i \cdot \phi_j$ where $\phi_i$ and $\phi_j$ were taken from the estimated value of the Reuters news dataset from Table \ref{table:reuters_small_delta_p_table}.
    \begin{table}[ht]
\centering
\begin{tabular}{rrrrrrr}
  \hline
interval index & $\widehat{\delta}$ & $\widehat{\delta}_{high}$ & $\widehat{\delta}_{low}$ & $\log(\mu)$ lower bound & $\log(\mu)$ upper bound & $\widehat{p}$ \\
  \hline
   0 & 2.955 & 2.955 & 2.955 & -1.500 & -1.000 & $2.264^*$ \\
     1 & -1.009 & -0.896 & -1.125 & -1.000 & -0.500 & -1.501 \\
     2 & -0.202 & 0.152 & -0.450 & -0.500 & 0.000 & -0.842 \\
     3 & -0.127 & 0.416 & -0.432 & 0.000 & 0.500 & 0.561 \\
     4 & 0.077 & 0.480 & -0.237 & 0.500 & 1.000 & 0.197 \\
     5 & 0.776 & 1.129 & 0.503 & 1.000 & 1.500 & -0.475 \\
     6 & -1.133 & -1.039 & -1.288 & 1.500 & 2.180 & 0.744 \\
   \hline
\end{tabular}
\caption{The estimated $\delta$ and p from fitting piecewise linear regression on log of sample mean and log of sample variance on each interval from Reuters news small dataset. The $\widehat{\delta}_{high}$ and $\widehat{\delta}_{low}$ are intercepts from the data points that are having maximum or minimum signed distance from the fitted piecewise regression line while having corresponding same slope. Note: * indicates that the estimated value 2.264 is based on regression with only two observations which is highly unreliable and we could ignore the interval.}
\label{table:reuters_small_delta_p_table}
\end{table}

\item The count data $Y_{i j} = Y_{j i}$ were independently generated from Tweedie distribution with mean $\mu_{i j}$, power parameter $p_{i j}$, and the dispersion parameter $\phi_{i j}$.
\eit

We applied the alternating Tweedie regression algorithm to the generated data. For comparison, the gradient descent algorithm with Adam update was applied to the data. For the Adam update, we used the Adam optimizer (torch.optim.Adam) from the PyTorch package. A learning rate scheduler was used to reduce learning rate based on ReduceLROnPlateau from torch.optim.lr\_scheduler. The default setting of this function is used, which specifies the loss does not improve in 10 epochs, the learning rate is reduced by factor of 0.1. With the Adam update, computationally we only need to compute the loss and keep track of gradient of loss with respect to parameters being updated by setting .requires\_grad\_(True) for the parameter. The gradient was computed with the .backward() method of the loss tensor. The parameter update was done with the optimizer's .step() method. The optimizer's gradient was reset to zero within each epoch with the .zero\_grad() method. For fair comparison, the innermost for loops of epoch are set to be 10 for all three update methods: with and without learning rate adjustment and Adam update.
% Compared to our Algorithm \ref{algorithm:AlternatingTweedie}, the difference between Adam update and ours is that the two innermost for loops of epoch are changed to two for loops with 30 epochs of the Adam update.

% (see 16_optimizing_by_GD.ipynb on Colab, in hwangmogo@gmail.com)
% (Tweedie MLE with no learning rate: see 15-4-Simulation_for_Tweedie_MLE_approach_adding_epoch_no_lr.ipynb, in hwangmogonew2@gmail.com)
% (Tweedie MLE with lr: see 15-2-Simulation_for_Tweedie_MLE_approach_adding_epoch.ipynb, in hwangmogonew2@gmail.com)
% all above three cases used simul_4.sqlite DB

\begin{figure}[H]
    \centering
    \includegraphics[width=0.8\linewidth]{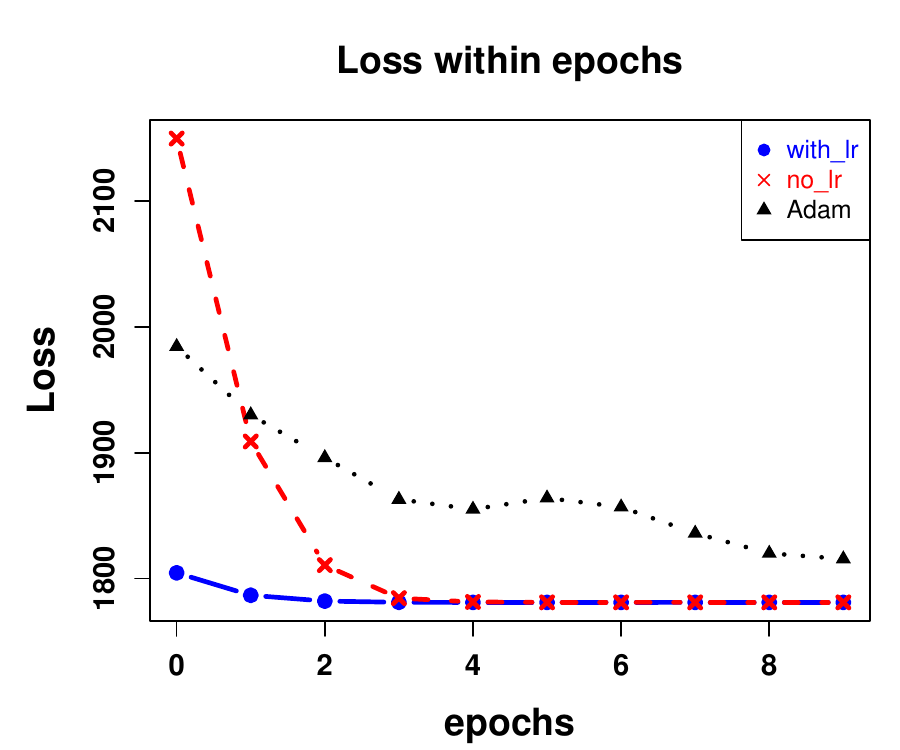}
    \caption{The loss reduction was compared within epochs among three different updates: the alternating Tweedie regression algorithm with and without learning rate adjustment, and Adam update. The results are from the first iteration and first row of data matrix in our Algorithm \ref{algorithm:AlternatingTweedie}.}
    \label{fig:epoch_loss_Tweedie_Adam_comparison}
\end{figure}
\begin{figure}[H]
    \centering
    \includegraphics[width=0.75\linewidth]{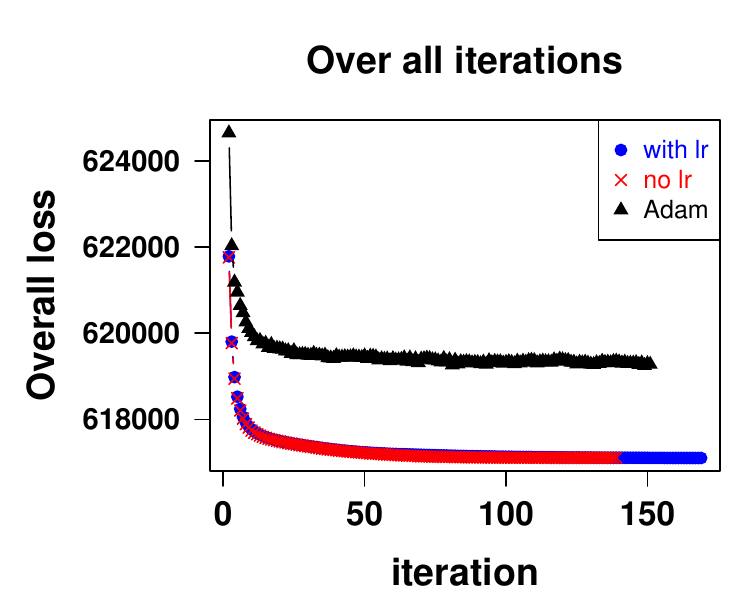}
    \caption{The overall loss over iterations among three different update methods: with or without learning rate adjustment and the Adam update. The Fisher scoring type update with or without learning rate adjustment started with lower overall loss than the Adam update, and reduces the overall loss faster as the iteration number increases.}
    \label{fig:overall_loss_Tweedie_Adam_comparison}
\end{figure}

% see 17-Tweedie_MLE_performance_comparison_within_epoch.ipynb, in account kimtj6895@gmail.com
% used simul_4.sqlite DB
Figure {\ref{fig:epoch_loss_Tweedie_Adam_comparison}} shows how different parameter update method performs within the innermost loop in our alternating Tweedie regression. The three different update methods are Fisher scoring type update with and without learning rate adjustment, and gradient descent Adam update. The loss reductions presented are from the first iteration and first row of data matrix in Algorithm \ref{algorithm:AlternatingTweedie}. All three lead to quick reduction of the loss within 5 epochs. The updates with or without learning rate adjustment reach stable loss value quickly. On the other hand, the Adam update does not stabilize as fast as the other two methods. Further, note that the with or without learning rate adjustment updates' loss values are small compared to that of the Adam update. Taking into account the fact that all three updates used same simulated data and started with identical initial values, the difference in the loss values are purely due to different updating schemes. Deriving the first and second derivatives manually and using the Fisher scoring algorithm makes a difference in finding right direction of the update. This point is even more obvious when we look at the overall loss over many iterations of updates in Figure \ref{fig:overall_loss_Tweedie_Adam_comparison}. This figure contains the loss of all rows and columns of the data matrix from over 120 iterations. The Adam algorithm decreased the overall loss as the iteration increases but it is not as effective as the other two updating methods. Since the Adam update is inefficient in finding the right optimizing direction in the alternating Tweedie regression algorithm, we exclude it from further discussion and only compare the Fisher scoring type update with or without learning rate adjustment in more detail.

% see 17-Tweedie_MLE_performance_comparison_within_epoch.ipynb, in kimtj6895@gmail.com
% used simul_4.sqlite DB

\begin{figure}
    \centering
    \includegraphics[height=6in, width=4in]{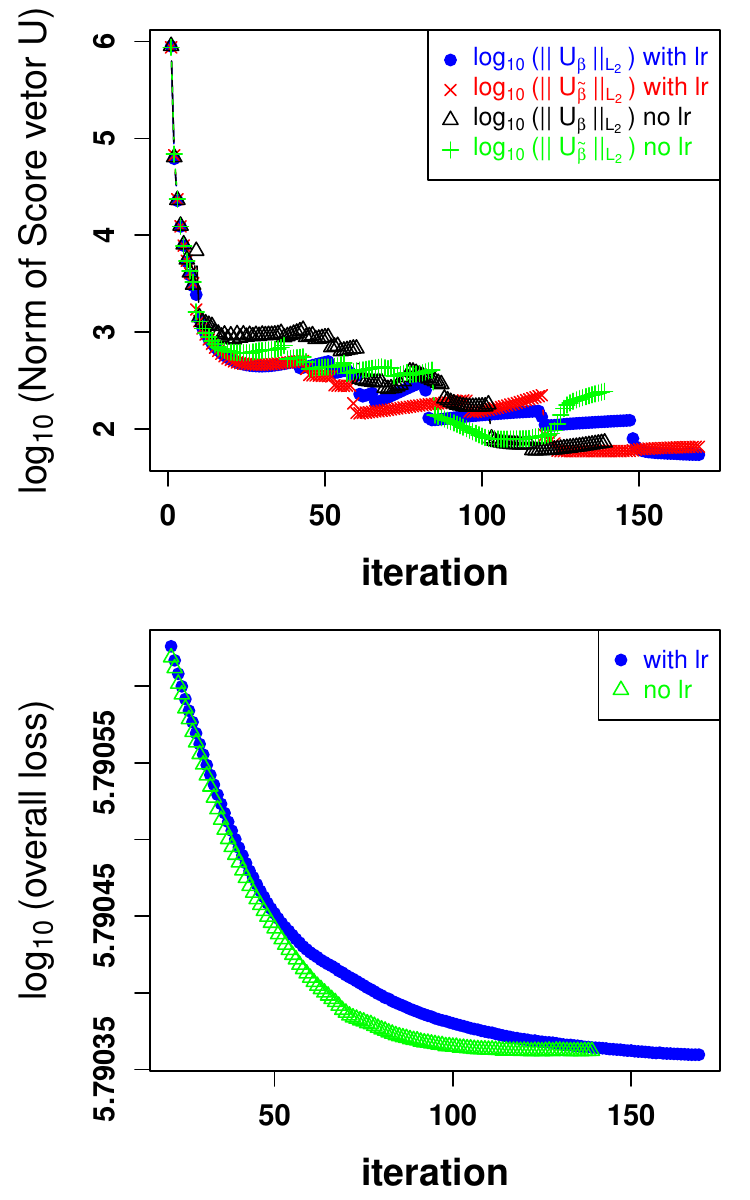}
    \caption{\small The $\log_{10}$ scaled norm of the score vector and the overall loss as iteration proceeds from simulated data. The top panel shows norm of the score vector in $\log_{10}$ scale for two cases: with learning rate, and without learning rate. The bottom panel illustrates $\log_{10}$ overall loss versus iteration for the two cases. Overall, both cases are reducing the overall loss and the norm of score vector. The case with no learning rate is faster to reduce the overall loss in earlier iterations but may not achieve the minimum overall loss in the end. The case with learning rate moves slowly in earlier iterations but shows advantage in the end by finding smaller value in the overall loss.}
    \label{fig:simul_4_loss_Ubetas_norm_combined}
\end{figure}
\begin{figure}
    \centering
    \includegraphics[height=3.2in, width=4in]{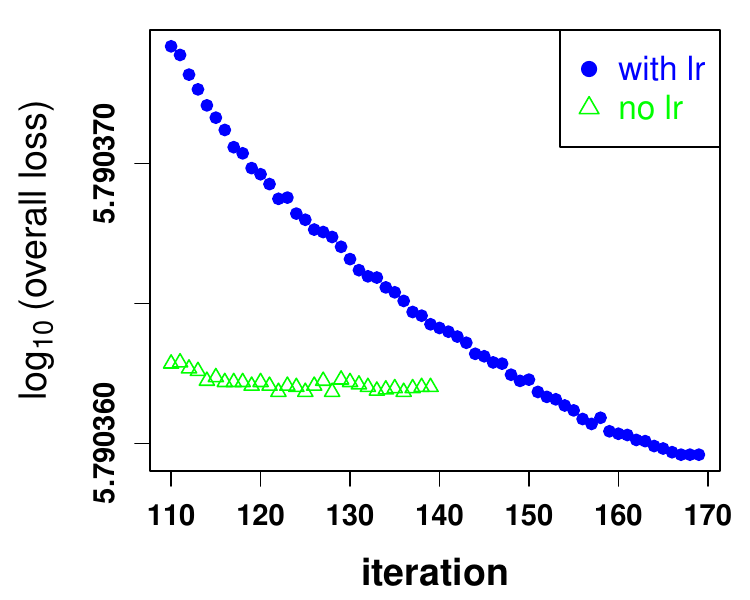}
    \caption{\small The overall loss in $\log_{10}$ scale during iteration between 110 and 170 for the two cases: with learning rate, and without learning rate. The update without learning rate adjustment is stabilized in a certain value before reaching the minimum overall loss. The algorithm with learning rate adjustment continuously reduces the overall loss until satisfying the convergence criterion, even though it was less effective in earlier iterations compared to the one with no learning rate.}
    \label{fig:simul_4_loss_detailed}
\end{figure}

\begin{figure}
\setlength{\tabcolsep}{-2pt}
\renewcommand{\arraystretch}{0.3}
\begin{tabular}{cccc}
    \includegraphics[width=0.25\linewidth]{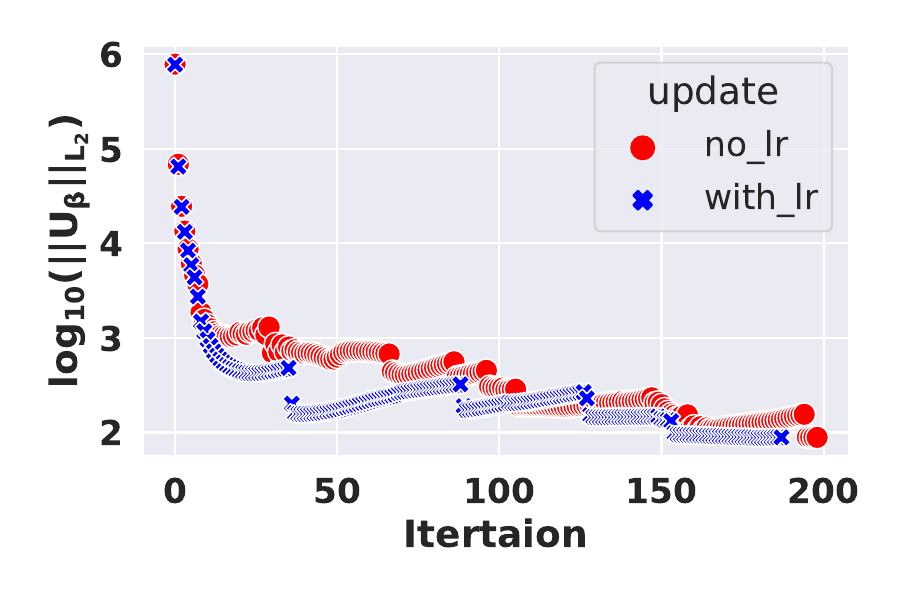} &
    \includegraphics[width=0.25\linewidth]{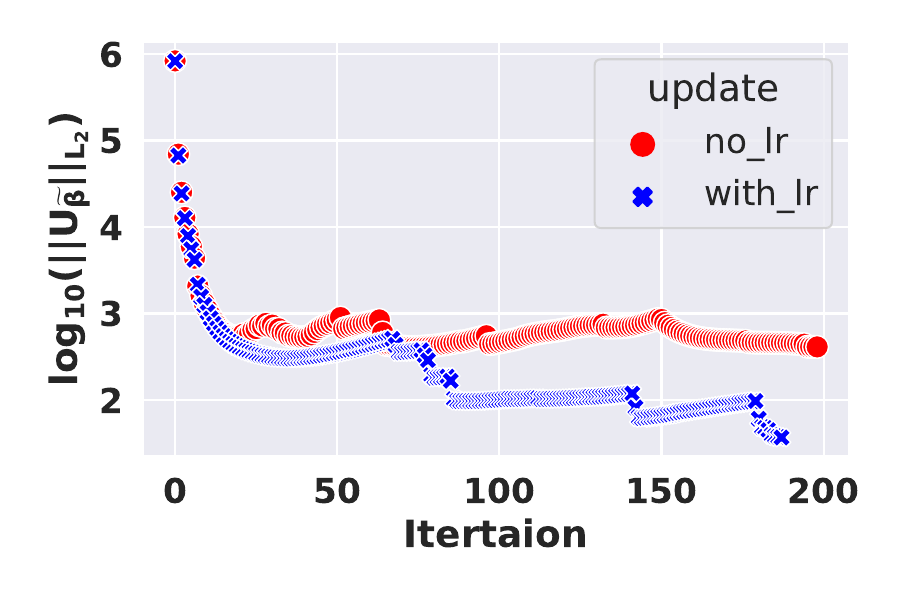} &
    \includegraphics[width=0.25\linewidth]{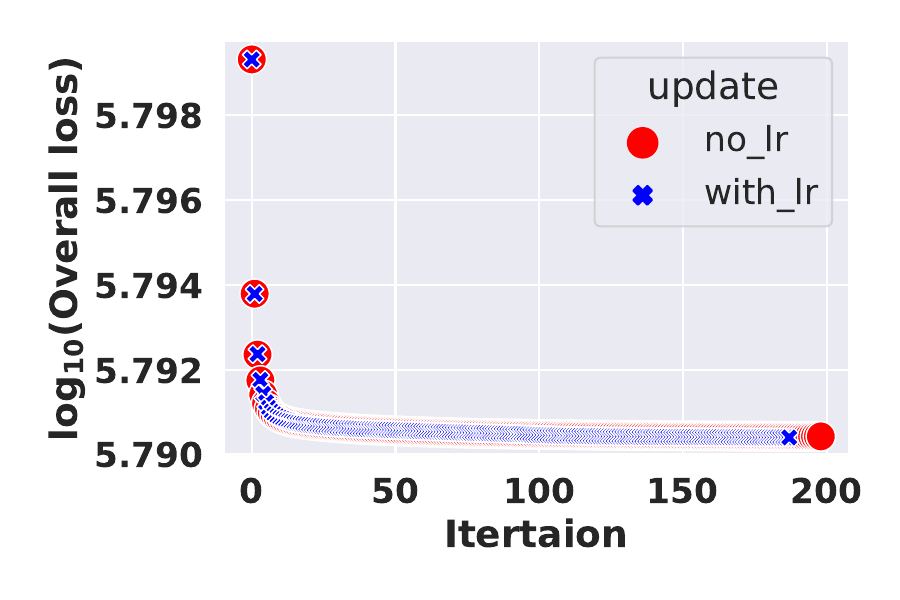} &
    \includegraphics[width=0.25\linewidth]{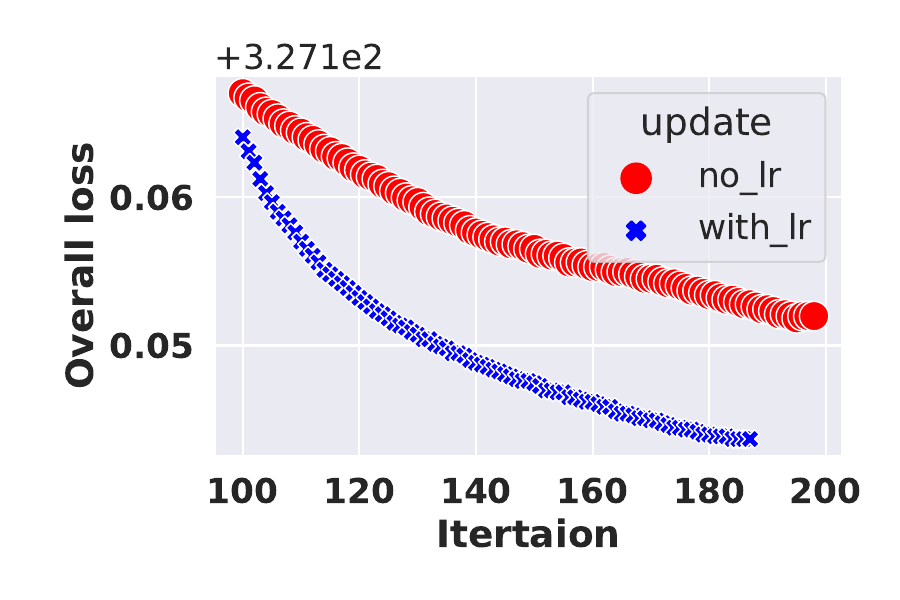} \\
    \includegraphics[width=0.25\linewidth]{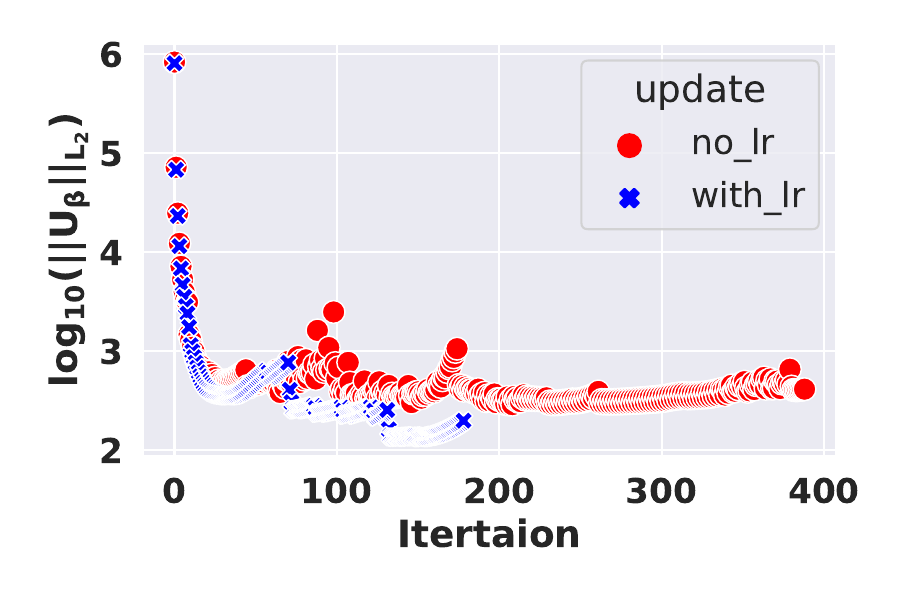} &
    \includegraphics[width=0.25\linewidth]{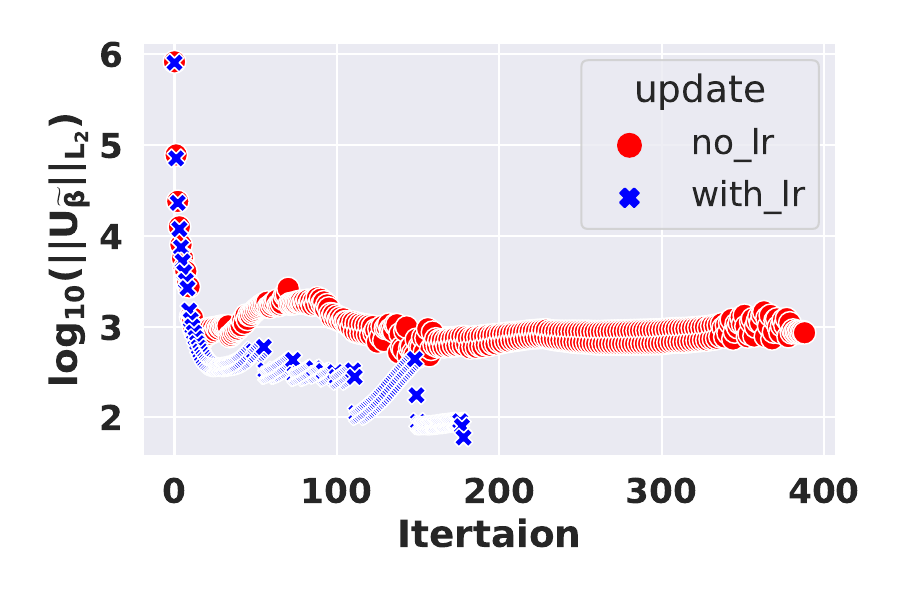} &
    \includegraphics[width=0.25\linewidth]{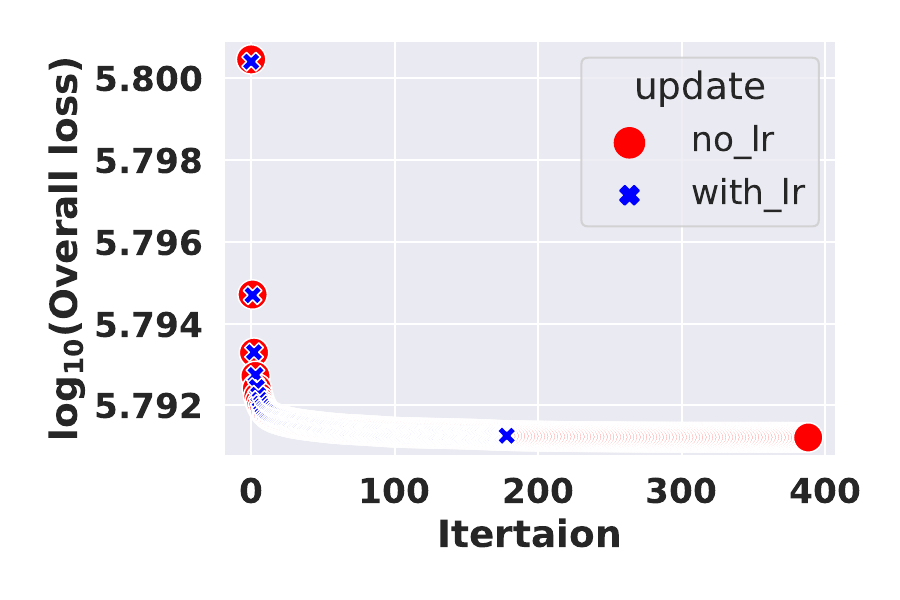} &
    \includegraphics[width=0.25\linewidth]{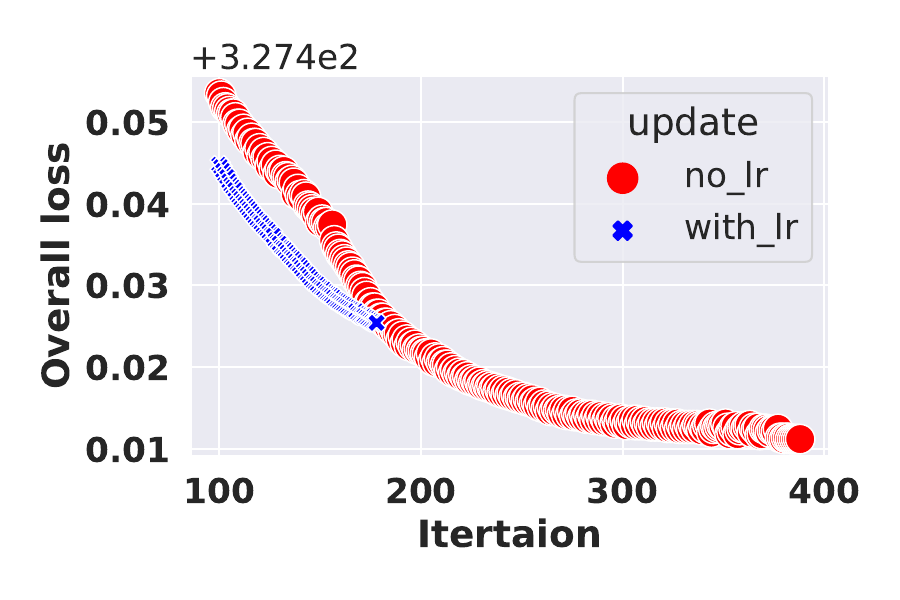} \\
    \includegraphics[width=0.25\linewidth]{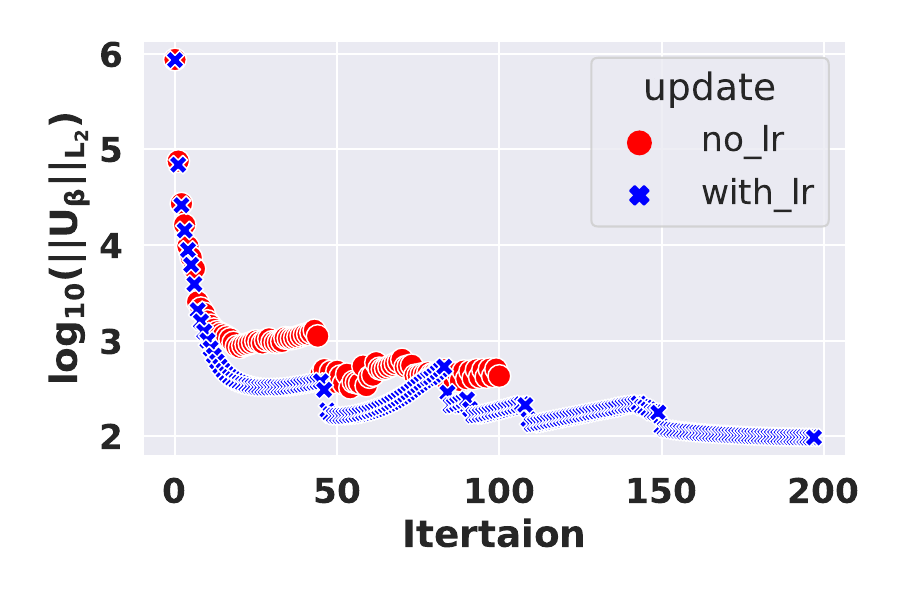} &
    \includegraphics[width=0.25\linewidth]{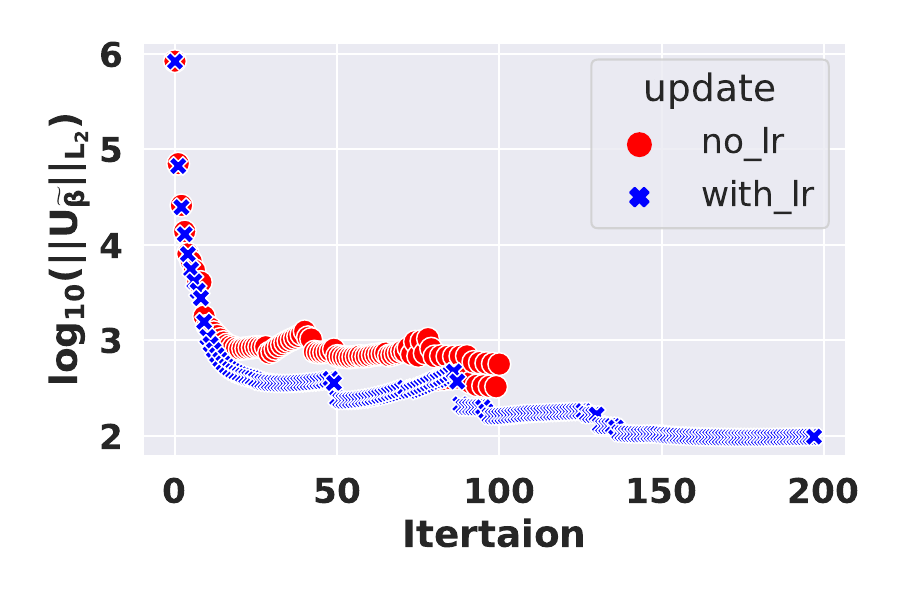} &
    \includegraphics[width=0.25\linewidth]{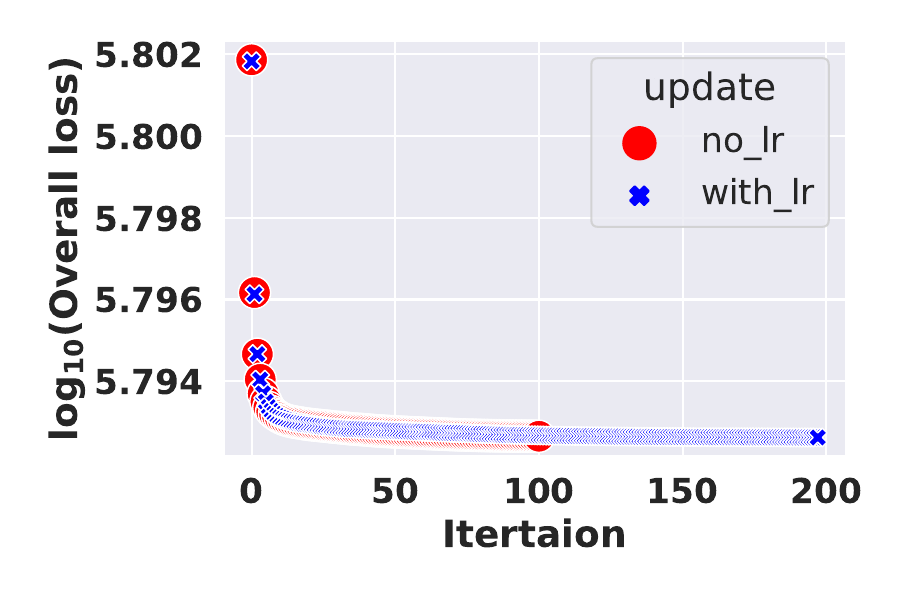} &
    \includegraphics[width=0.25\linewidth]{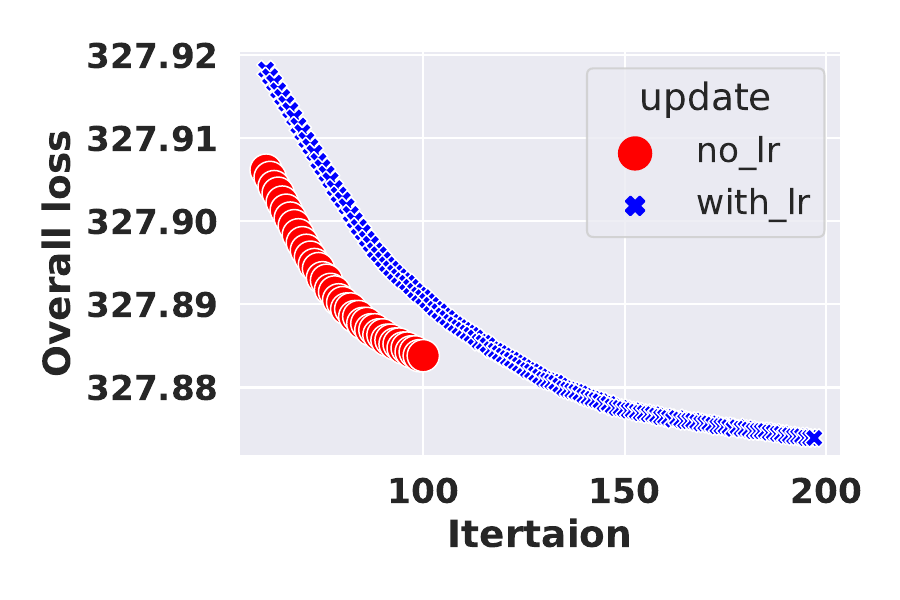} \\
    \includegraphics[width=0.25\linewidth]{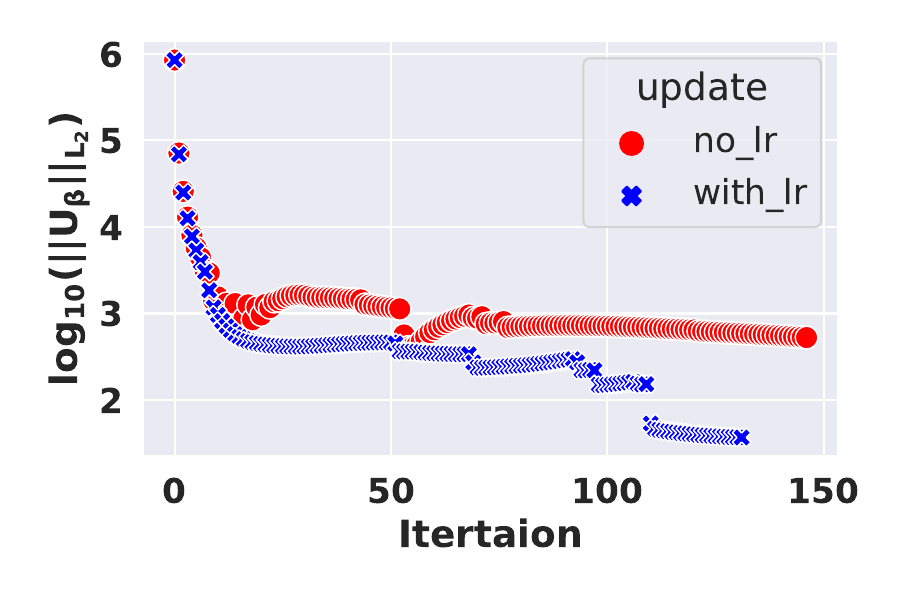} &
    \includegraphics[width=0.25\linewidth]{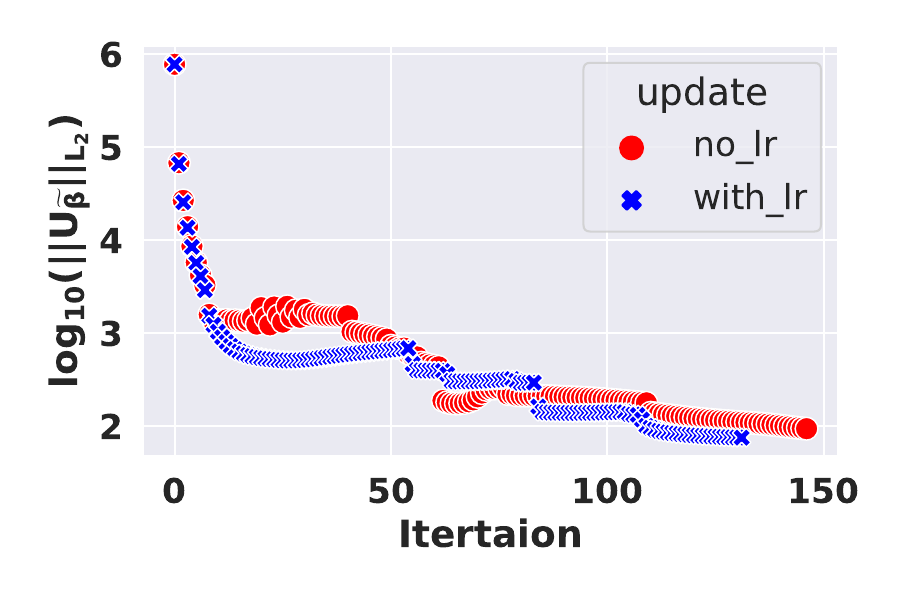} &
    \includegraphics[width=0.25\linewidth]{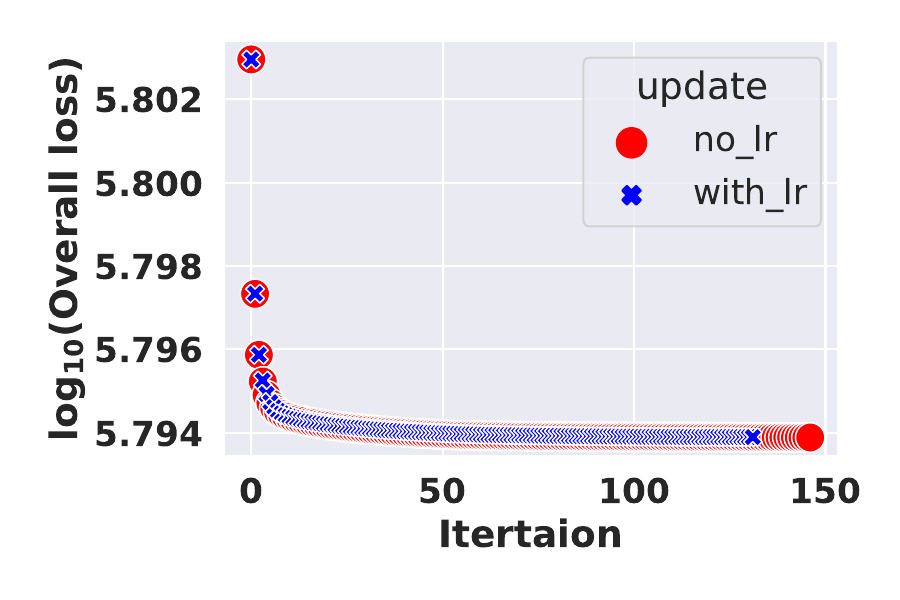} &
    \includegraphics[width=0.25\linewidth]{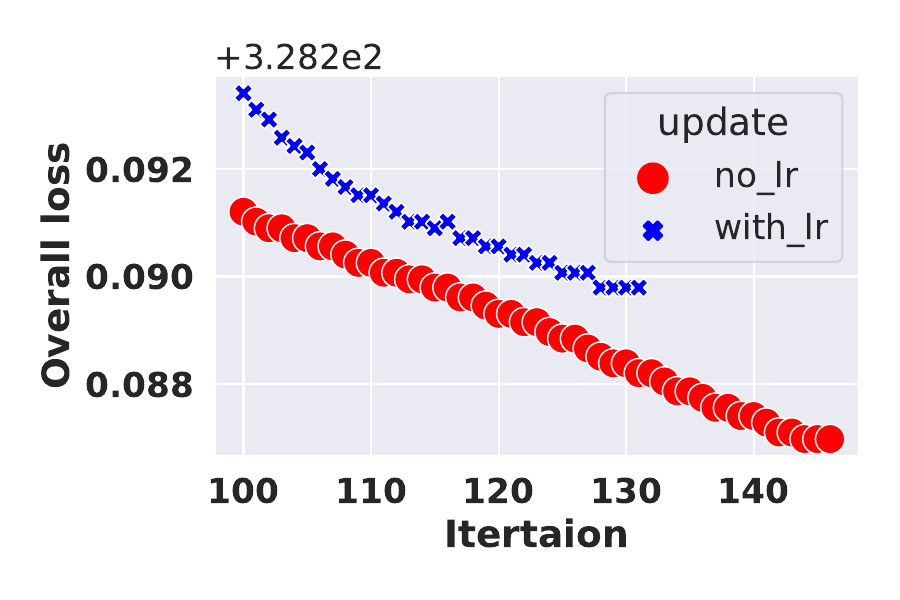} \\
    \includegraphics[width=0.25\linewidth]{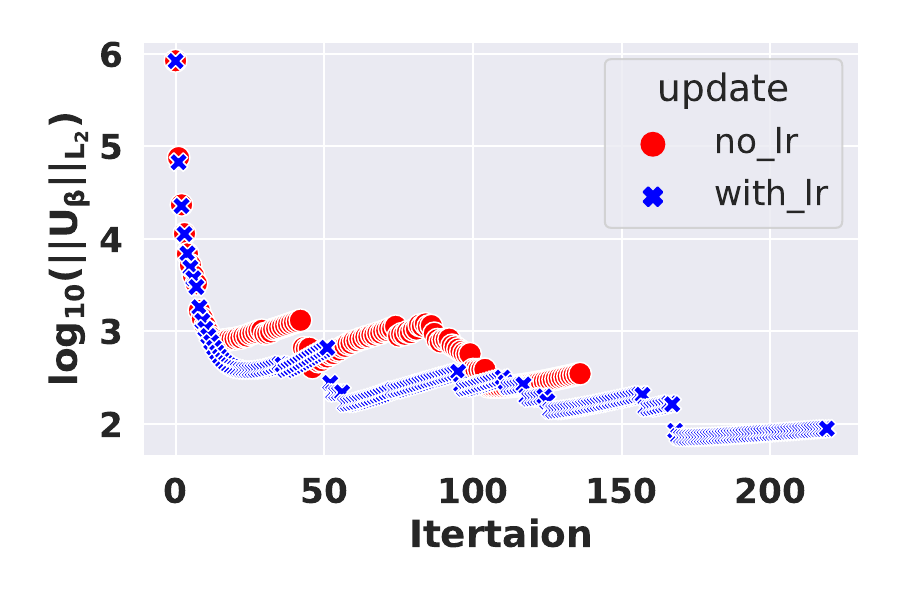} &
    \includegraphics[width=0.25\linewidth]{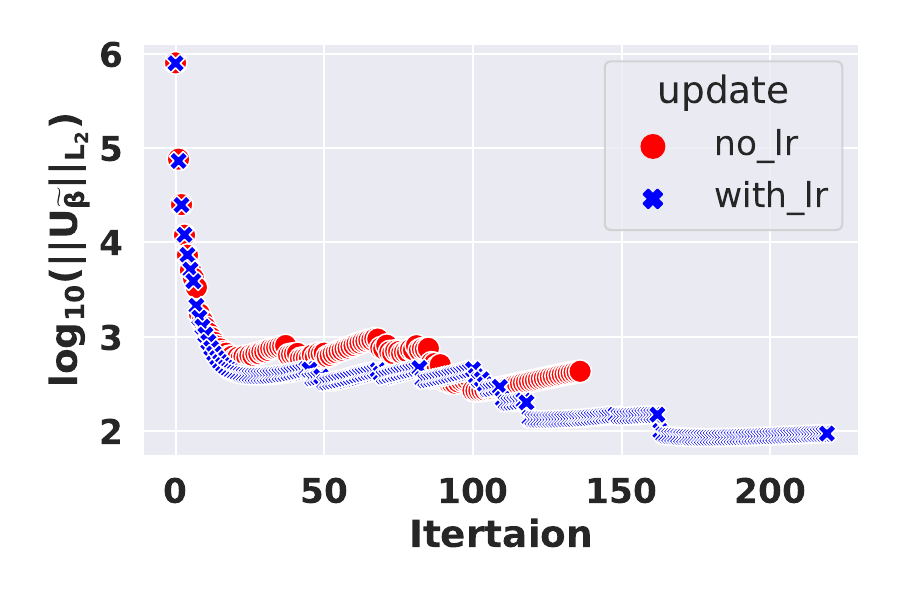} &
    \includegraphics[width=0.25\linewidth]{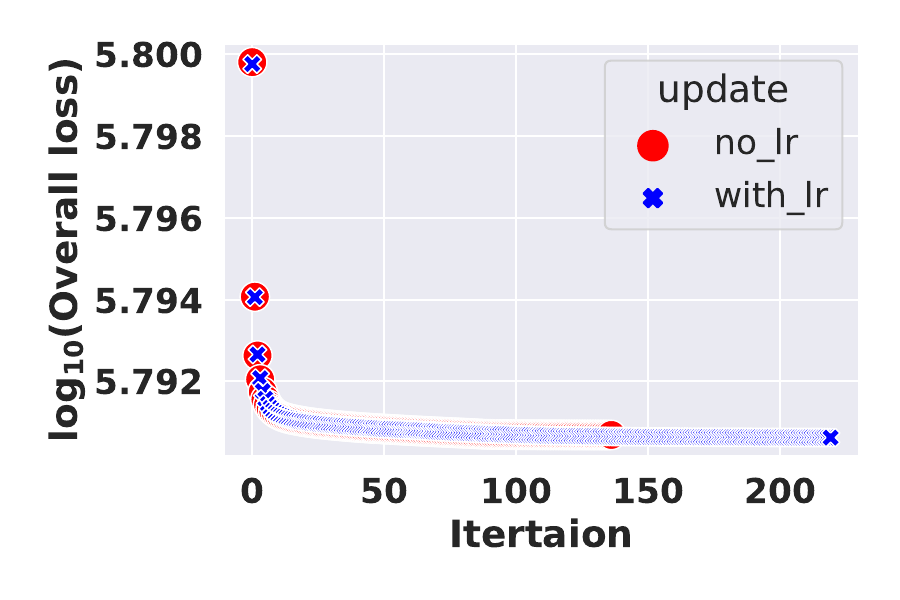} &
    \includegraphics[width=0.25\linewidth]{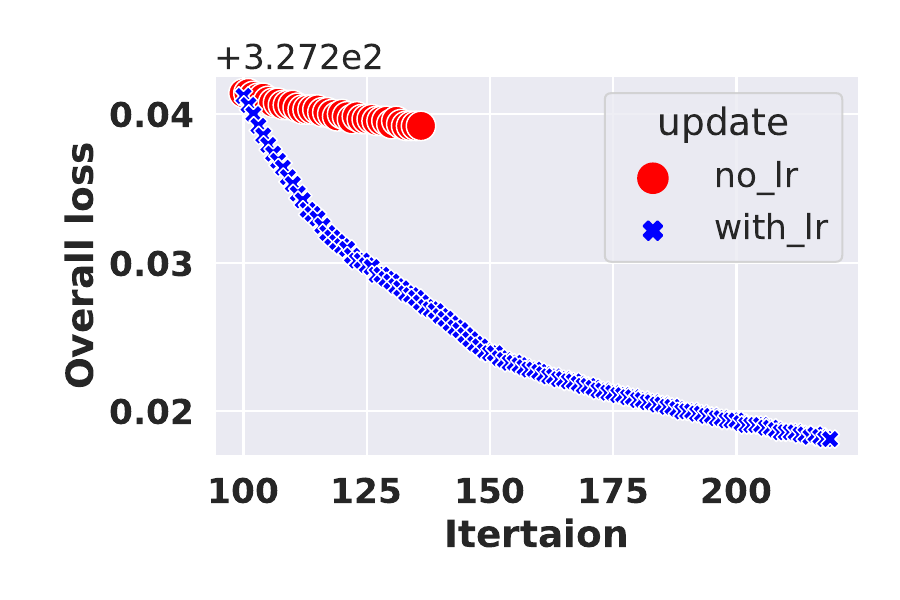} \\
    \includegraphics[width=0.25\linewidth]{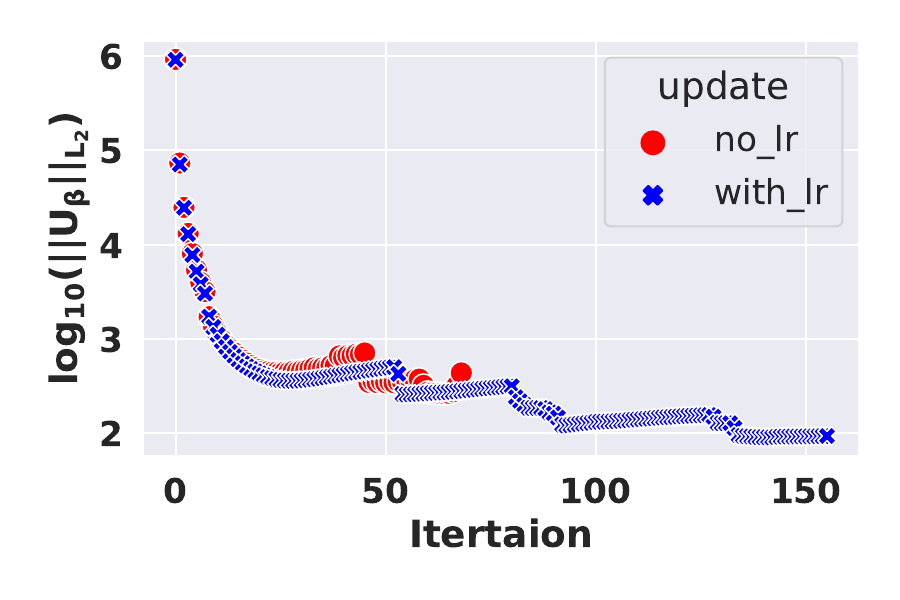} &
    \includegraphics[width=0.25\linewidth]{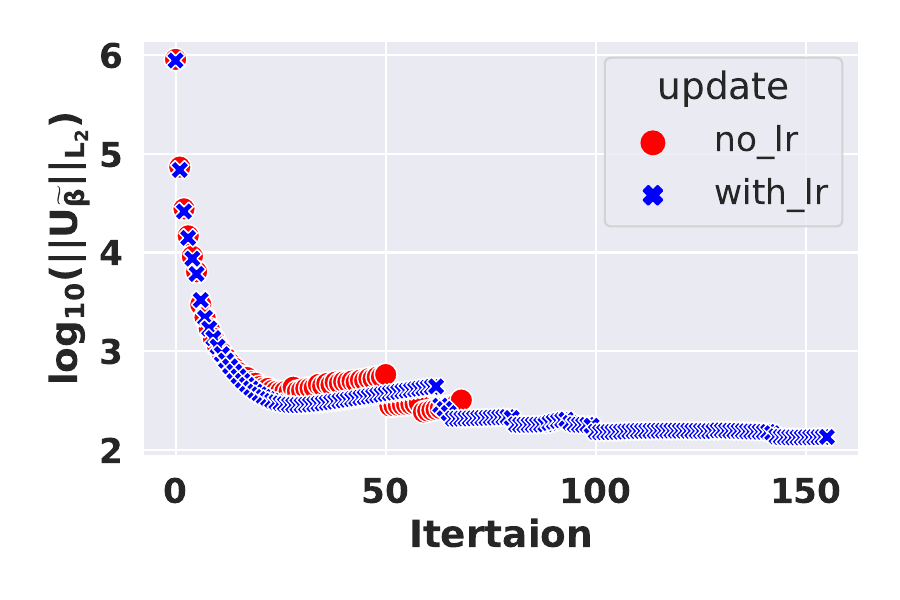} &
    \includegraphics[width=0.25\linewidth]{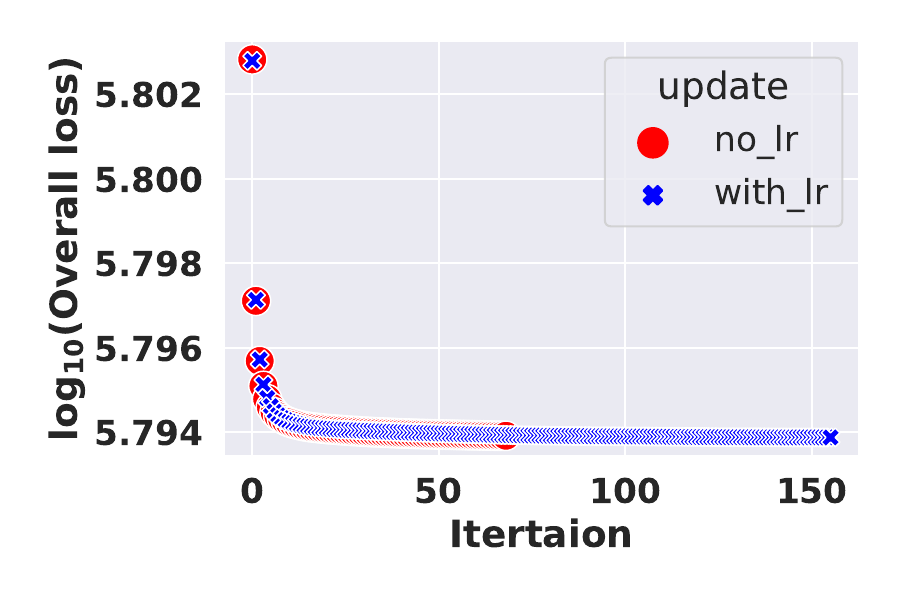} &
    \includegraphics[width=0.25\linewidth]{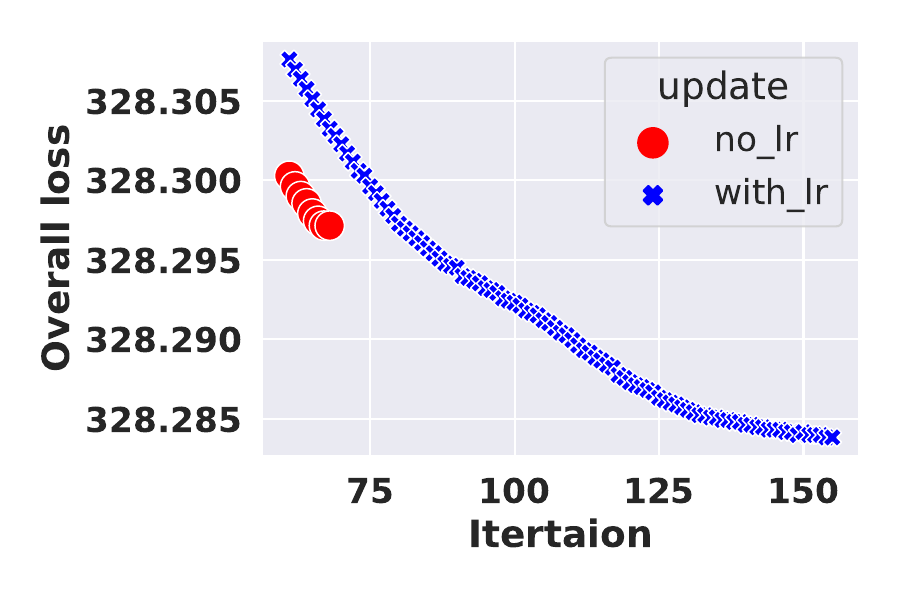} \\
    \includegraphics[width=0.25\linewidth]{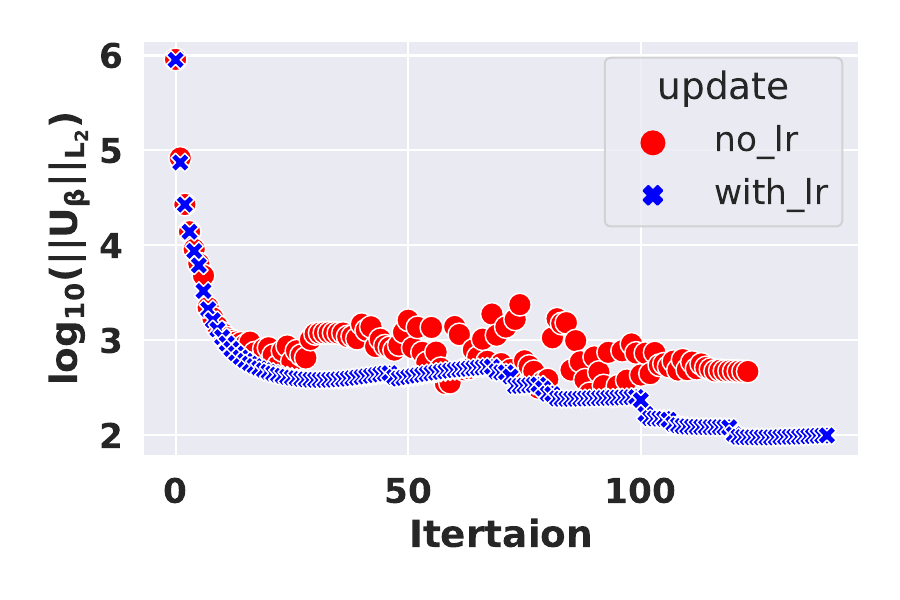} &
    \includegraphics[width=0.25\linewidth]{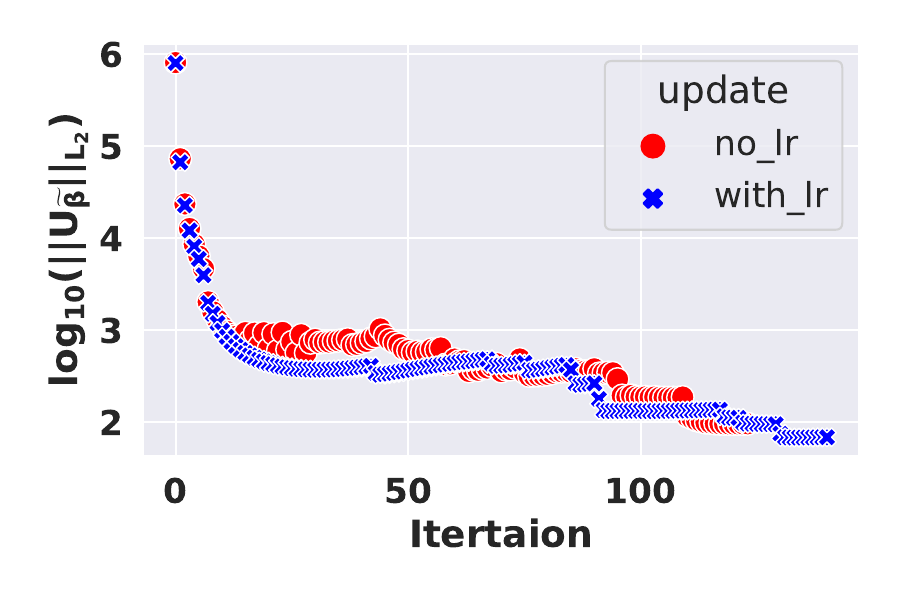} &
    \includegraphics[width=0.25\linewidth]{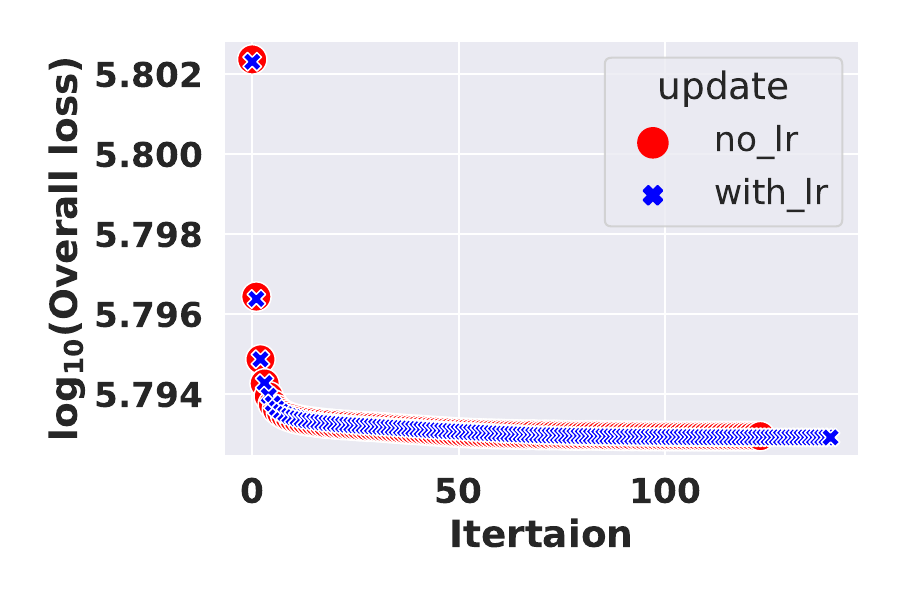} &
    \includegraphics[width=0.25\linewidth]{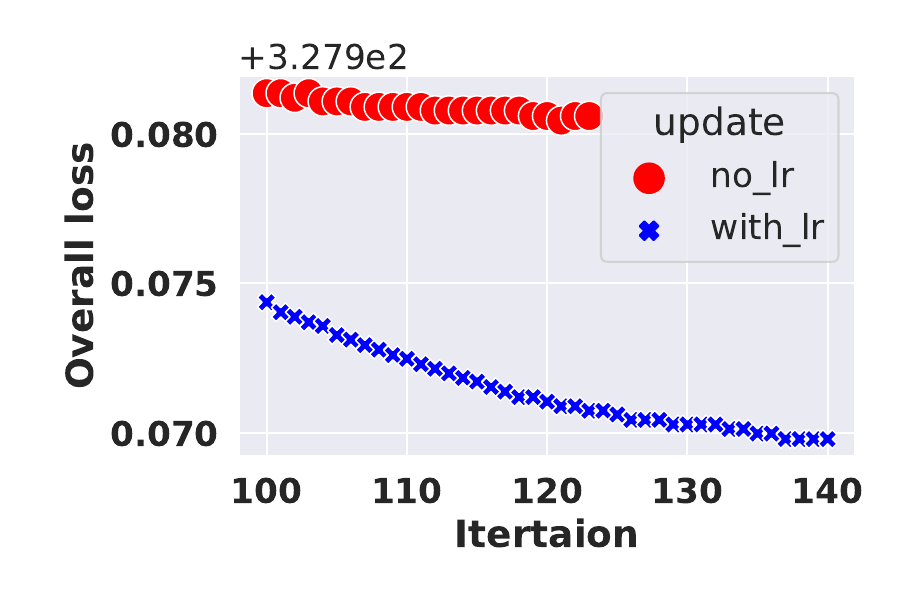} \\
    \includegraphics[width=0.25\linewidth]{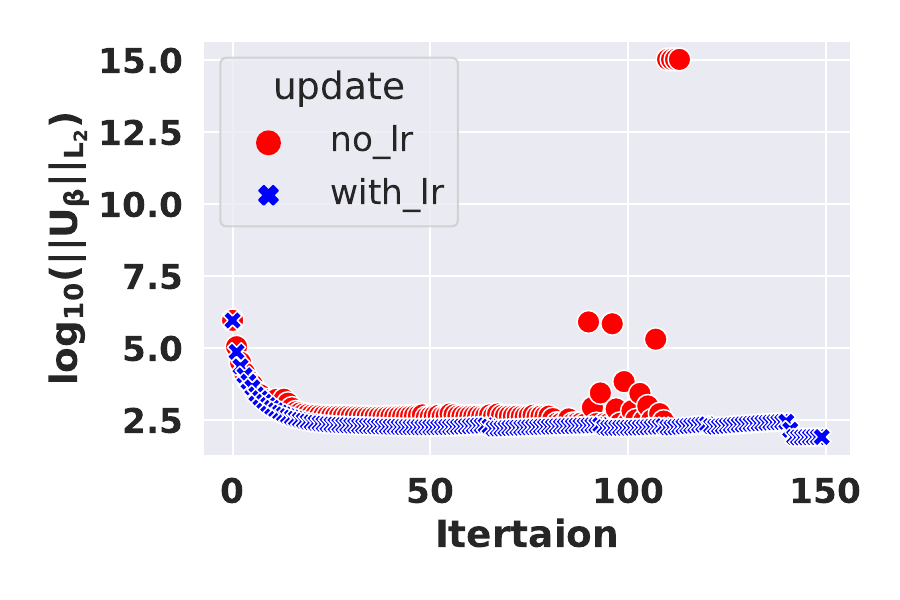} &
    \includegraphics[width=0.25\linewidth]{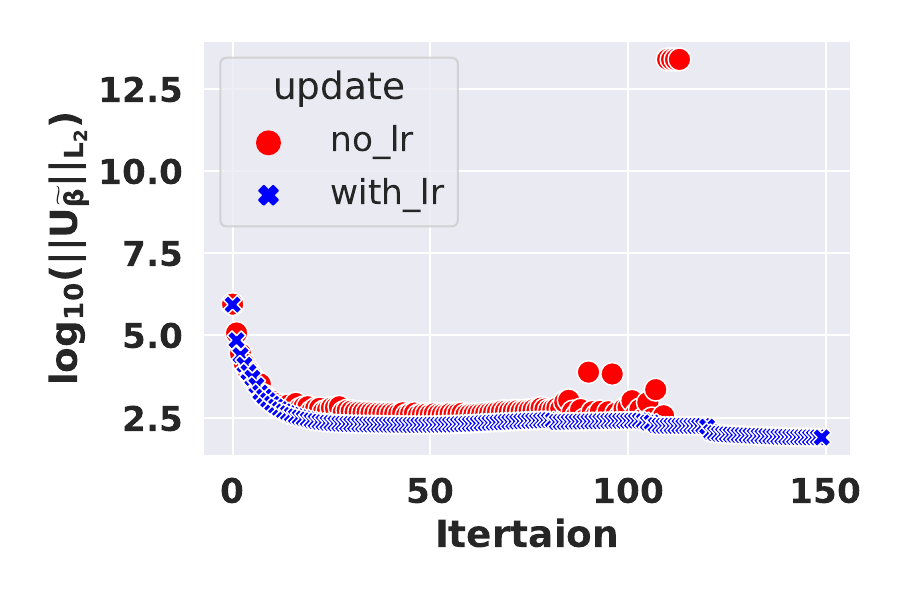} &
    \includegraphics[width=0.25\linewidth]{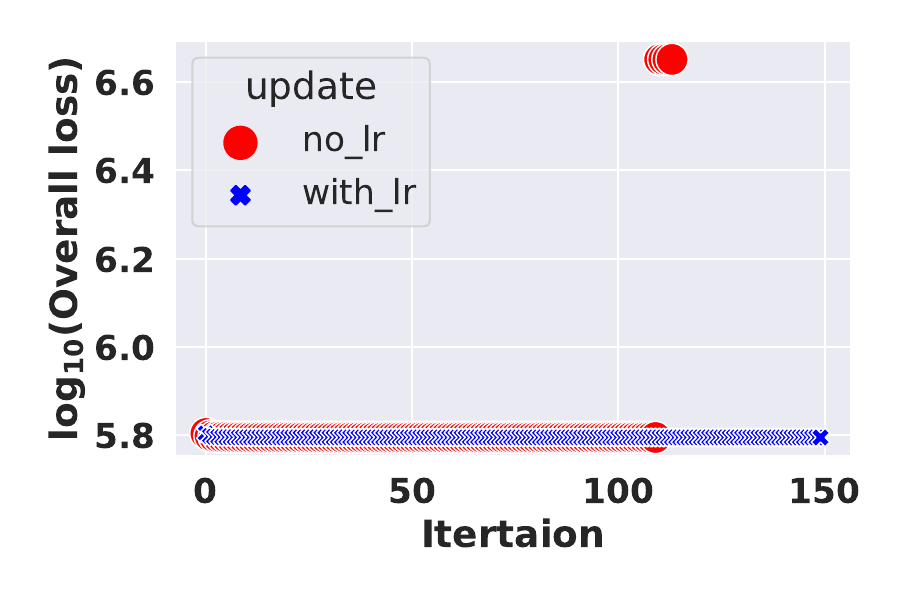} &
    \includegraphics[width=0.25\linewidth]{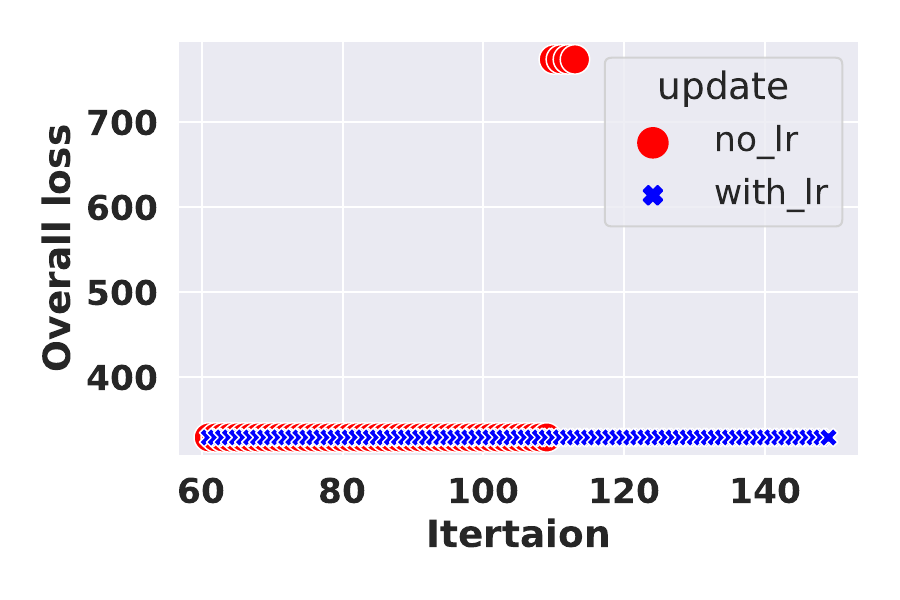}
\end{tabular}
\vspace{-0.2in}
\caption{\small Comparing performance of the alternating Tweedie regression algorithm with or without learning rate over 8 simulated datasets. Each row is for one dataset.}
\label{fig:MLE_Tweedie_overall_simul_plots}
\end{figure}

Figures \ref{fig:simul_4_loss_Ubetas_norm_combined} and \ref{fig:simul_4_loss_detailed} show more detailed examination of the Fisher scoring update with or without learning rate adjustment. We can see from Figure \ref{fig:simul_4_loss_Ubetas_norm_combined}, during the first 20 iterations, the two cases behave similarly in the norm of score vector. However, from 20th to around 130th iteration, the update with no learning rate is more effective in reducing the overall loss than the one with the learning rate. This is not the case for the norm of the score vector. From 20th to around 80th iteration, update with no learning rate is a little less effective in reducing the norm of the score vector than the update with learning rate. However, from 80th to 120th iteration, the case with no learning rate is more aggressive than the one with learning rate. After 120th iteration until convergence, the no learning rate case starts to sacrifice $U_{\wtvbeta}$ severely to improve $U_{\vbeta}$ while maintaining the overall loss in the same neighborhood. On the other hand, small updates with learning rate adjustment help to make the algorithm achieve lower overall loss. Figure \ref{fig:simul_4_loss_detailed} shows how the log loss changes in the last 60 iterations. The one with no learning rate fluctuates a little bit around the same value 617109.2 and was trapped there causing the algorithm to believe the convergence criterion is satisfied. The one with learning rate consistently reduces the overall loss and achieves smaller overall loss 617105.8 when the convergence criterion is satisfied. To verify this pattern is not happened by chance, we generated 8 more datasets and applied the algorithm to them. The results are summarized in Figure \ref{fig:MLE_Tweedie_overall_simul_plots}. Basically, the behavior of the algorithm is same as we illustrated above.

The first column of Figure \ref{fig:MLE_Tweedie_overall_simul_plots} shows the $L_2$ norm of the score vector $U_{\vbeta}$ in $\log_{10}$ scale over all iterations for both with and without learning rate. Both cases perform well during the first 20 iterations by reducing the $L_2$ norm of the score vector $U_{\vbeta}$ quickly. It is a little  surprising that the one with learning rate adjustment was also able to achieve this because with learning rate adjustment $\frac{lr}{it^{1/4}}$, the updated amount in $\hat{\vbeta}$ only gets between a half and a quarter of the amount received from the update without learning rate adjustment. During iterations 20 to 200, the algorithm with learning rate adjustment updates the $\hat{\vbeta}$ by a much smaller fraction (23.6\% to 13.3\%) of the amount updated using the algorithm with no learning rate adjustment. As a result, the algorithm with learning rate adjustment cautiously stays around the estimated value from previous iteration and approaches the target little by little. The result of this is that reducing the learning rate at later iterations optimizes the objective function more efficiently and consistently compared to the algorithm without learning rate adjustment. We see the norm of $U_{\vbeta}$ achieves lower value faster in the case with learning rate adjustment than the other case. Further, there is less fluctuations when learning rate adjustment is used. This means the algorithm without learning rate adjustment often over-corrects its estimate for the model parameters in later iterations. The second column illustrates the $L_2$ norm of score vector $U_{\wtvbeta}$ in $\log_{10}$ scale over all iterations for both with and without learning rate. For the $U_{\wtvbeta}$ part, very similar pattern was observed as the first column.

The third column of Figure \ref{fig:MLE_Tweedie_overall_simul_plots} shows overall loss in $\log_{10}$ over all iterations. Overall trend is both reduces the loss quickly in early iterations and gradually slow down toward the optimal loss. It seems like both with and without learning rate performs similarly well. However, if we look in more detail, we could observe some difference as shown in the last column of the figure. Compare to the reduction in norm of $U_{\vbeta}$ and $U_{\wtvbeta}$, the loss function reduces consistently without much fluctuations. This aspect shows that the loss is not a monotone function with regard to $U_{\vbeta}$'s and $U_{\wtvbeta}$'s norms and there are possibly many different solutions giving similar loss value.

The last column of Figure \ref{fig:MLE_Tweedie_overall_simul_plots} is the zoomed in detailed performance near the end (i.e. near convergence) of the algorithm for 8 simulated datasets. The two updates, with or without learning rate, seemed to have different curves but both achieved similar final loss value in all except for one dataset in the sense that they only differ at the second digit after the decimal point. An exception happened in the last row of Figure  \ref{fig:simul_4_loss_detailed} for the overall loss: the update with no learning rate moved out of the region around the optimal solution leading to a sudden drastic increase in the loss function and couldn't get out of there. Upon convergence, the algorithm stayed there at the much higher final loss value.

{\modified
\section{\hlboth{Scalability \& application to Named Entity Recognition}}
\label{large_scale_app}
\subsection{\hlboth{Scalability to data with large vocabulary size \& training corpus}}
 \hlboth{In this section, we present the application of the proposed method on a large training corpus to learn word embeddings. We downloaded the January 20th, 2022 version of the Wikipedia dump from \url{https://dumps.wikimedia.org/enwiki/20220120/}. This Wikipedia dump has around 4 billion tokens and contains copies of all Wikipedia pages in xml format provided by Wikipedia.} %They are available on the webpage \\ \url{https://en.wikipedia.org/wiki/Wikipedia:Database\_download}.

\hlboth{
  We consider co-occurrence counts of WordPiece tokens.
The WordPiece tokenizer of \citealt{Wu-et-al-Wordpiece:2016} divides words into a limited set of common subword units referred as wordpieces. This tokenizer was used in BERT (\citealt{devlin-etal-2018-bert}) to tokenize words into subword tokens. We adopted the same tokenizer, which gives us
%. The most commonly used BERT base uncased version has
30,522 WordPiece tokens. Removing some international characters leads to 26,531 tokens for us to use. With WordPiece tokenization, the word `cucumber' is broken into the sequence of characters `cu', `\#\#cum', `\#\#ber', where \#\# indicates that the character is a middle part of the word. We tokenized the entire training corpus with this tokenizer and obtained token-token co-occurrence counts ($X_{ij}$) based on our definition in equation (\ref{formula:cooccurrence-count}).}
%\colorlet{shadecolor}{yellow!35}
%\begin{shaded*}
{\modified
\bqan
\label{formula:cooccurrence-count}
X_{ij} =\left\{\begin{array}{cc}
    \sum_{\mbox{s $\in$ all sentences}} 1/d_{ij}(s),  & \mbox{if $d_{ij}(s) \le$ k} \\
    0, & \text{otherwise}
\end{array} \right.
\eqan
where $d_{ij}(s)$ = separation between word i and word j in sentence s.
}
%\end{shaded*}
%{\modified
%\bqa
%X_{ij} =\left\{\begin{array}{cc}
%    \sum_{\mbox{s $\in$ all sentences}} 1/d_{ij}(s),  & \mbox{if $d_{ij}(s) \le$ k} \\
%    0, & otherwise
%\end{array} \right.
%\eqa
%}
%\hly{\textcolor{blue}{
%where $d_{ij}(s)$ = separation between token $i$ and token $j$ in sentence $s$, and $k$ is pre-determined window size 10.

\hlboth{The sparse entries of the co-occurrence counts were stored in a SQLite database to facilitate model training. This sparse format addresses memory limitations and significantly enhances computational efficiency. By storing the data in the database, we can retrieve a single row of the co-occurrence matrix as needed, rather than loading all entries into memory at once. This reduces memory usage and speeds up data retrieval.}

\hlboth{
To further optimize the computational efficiency of data loading, we implemented a memory-efficient technique when fetching individual rows of the co-occurrence matrix. It is important to note that one row of the co-occurrence matrix corresponds to multiple rows in the database. The technique consists of two key strategies:}

%\colorlet{shadecolor}{yellow!35}
%\begin{shaded*}
\begin{enumerate}%[wide = 0pt, before=\vspace*{-\dimexpr\topsep+\partopsep+1.5ex}, after=\vspace*{-\dimexpr\topsep+\partopsep+1.5ex}]
\item
\textbf{Utilizing the `rowid' column in SQLite}: The rowid is automatically generated for most SQLite tables and serves as a fast-access key. SQLite stores data in a B-tree structure, which ensures quick lookups when using the rowid. According to SQLite documentation, searching for a record by its rowid is approximately twice as fast as searching by other indexed values. This optimization significantly reduces search time, especially with large tables, since locating rows in a B-tree requires at most $\log_2(n)$ steps, where $n$ is the total number of rows in the table.

\item \textbf{Restricting the number of entries}: To further enhance efficiency, we restrict the search to only a portion of the rowids rather than scanning the entire table. For instance, when retrieving the first row of a co-occurrence matrix with a vocabulary size of 400K, we specify a rowid range from 0 to 400K using the WHERE command. This is possible because the row indices (distinct from rowids) are already sorted in non-decreasing order. Therefore, when fetching data with a rowid of 0, there is no need to search beyond the first 400K entries. After retrieving each row, we update the starting rowid for the next retrieval based on the entries already processed.
\end{enumerate}
%\end{shaded*}

\hlboth{These two strategies significantly reduced the time required for data retrieval. Without these optimizations, fetching a single row of the co-occurrence matrix (with a vocabulary size of 400K) took approximately 376.42 seconds. With the strategies in place, the retrieval time was reduced to just 0.61 seconds.}

\hlboth{The counts corresponding to a single row of the co-occurrence matrix are used to compute a score vector and an information matrix, which are then utilized to update the embedding vector estimate for that row. While this update is based on token counts in a single row, the mean counts also depend on the embedding vectors of other tokens that appear within the context window. In essence, the update indirectly incorporates the vector representations of all other tokens, refining the embedding vector for each token iteratively. One complete iteration occurs when the algorithm has processed all rows of the matrix.}

\hlboth{
However, due to memory limitations, the co-occurrence counts cannot be retained in memory as the algorithm moves to subsequent rows. As a result, the data must be repeatedly retrieved from the database for each iteration. Since the algorithm requires numerous iterations to converge, each row of the co-occurrence matrix is retrieved multiple times—once per iteration—imposing a substantial computational burden and significantly increasing the time and resources required for convergence.}

\hlboth{
To address this challenge, we designed the algorithm to leverage CPU-GPU parallel processing. While data retrieval is handled by the CPU, parameter updates can be efficiently performed on the GPU. Since the parameter estimation cannot begin until the data is available, the data retrieval and parameter updating are generally performed sequentially. However, we parallelized these processes as follows:}

%\colorlet{shadecolor}{yellow!35}
%\begin{shaded*}
\begin{enumerate}%[wide = 0pt, before=\vspace*{-\dimexpr\topsep+\partopsep+1.5ex}, after=\vspace*{-\dimexpr\topsep+\partopsep+1.5ex}]
\item
\textbf{Data retrieval}: A child process handles data retrieval using Python’s multiprocessing package.
\item
\textbf{Parameter estimation}: The main process, utilizing the GPU through PyTorch, handles the parameter updates.
\item
\textbf{Queue-based data transfer}: Once a row is fetched, it is stored in a shared queue object that connects the child and main processes. The GPU retrieves data from the queue and removes it once used.
\item \textbf{Parallel execution}: While the GPU is processing the current row’s parameter update, the child process retrieves the next row of data. These two operations run in parallel, which reduces time and improves efficiency.

\item{Efficient resource management}: Data retrieval on the CPU only proceeds when the queue is empty, ensuring that memory resources are devoted to the computation and reducing overall memory usage.
\end{enumerate}
%\end{shaded*}

\hlboth{In general, parameter updates on the GPU take longer than data retrieval from the CPU. By running these processes in parallel, we are able to save significant time and improve computational efficiency.}

\hlboth{
In addition to the data retrieval and parallel processing techniques discussed earlier, another key strategy employed to make the computation feasible is data chunking. This approach is controlled by the parameter num\_chunks, which specifies the number of groups into which the co-occurrence counts for a single row of data are partitioned. Each group contains a small subset of entries, enabling batch processing that reduces GPU memory demands while shortening the computation time required to process an entire row of data.
%Although each row contains as many entries as the vocabulary size and some of which may be zeros, the zero entries still contribute to the log-likelihood function. They must be processed to compute their respective contributions to the score vector and the information matrix.
%
For each batch, the computation involves generating as many matrices of size $(d+1)\times(d+1)$ as there are entries in the batch. This process inherently requires balancing memory usage and computational efficiency. The Fisher information matrix calculation is particularly demanding, as it involves outer products of terms $\vw_i$ and $\vw_i^\top$, and between $\wtvw_j$ and $\wtvw_j^\top$. Due to memory constraints, directly computing these outer products without looping is impractical, while using loops is prohibitively slow.
To address this, we utilized the \textbf{einsum} function, which efficiently converts the outer product calculations into a summation over a 3D array, processed in smaller batches. This approach achieves acceptable computational speeds without resorting to explicit loops.The optimal batch size depends on the dimensionality
$d$ of the
 word vectors. Through trial and error, we found that the values 30, 10, and 4 work well for dimension  300, 100, and 50 respectively. However, these values are not rigid. A wide range of batch sizes can yield similar performance. For example, when $d=100$, the computation time for processing one row with $num\_chunks = 10$ is comparable to that with $num\_chunks=4$. Similarly, for $d=50$, the computation time with $num\_chunks=10$  is similar to that with $num\_chunks=4$. }

\hlboth{
With these computational techniques, %CPU-GPU parallel processing and using 26,531 WordPiece tokens in our vocabulary,
 we are able to carry out the model estimation for the SA-Tweedie models trained on the entire Wikipedia data.}

\hlboth{
The SA-Tweedie model was applied to the log(co-occurrence count+1) of token-token pairs in the entire Wikipedia dump data. The reason that we decide to use log count is because the skewness of the raw counts are too big to be modeled well. From the co-occurrence matrix, we explored the raw count data in each row to compute the sample skewness. The histogram of the sample skewnesses from different rows are given in the left panel in Figure \ref{fig:skewness_wordpiece_wiki_all_rows}. We can see that most of the skewness values are much larger than 40. It is very difficult to model the data if the skewness is over six. The log-counts seems to have sample skewness values mostly within 6. Even after log-transformation, the data in most of the rows are still very skewed
%(see Figure \ref{fig:histogram_wordpiece_wiki_cooccur_matrix_random_rows} for histogram of some randomly selected rows)
and have plenty of zero observations. For this reason, we decided to model the log-counts using the alternating Tweedie regression model.
}
\begin{figure}[H]
    \centering
    \begin{tabular}{cc}
    \includegraphics[width=0.45\textwidth]{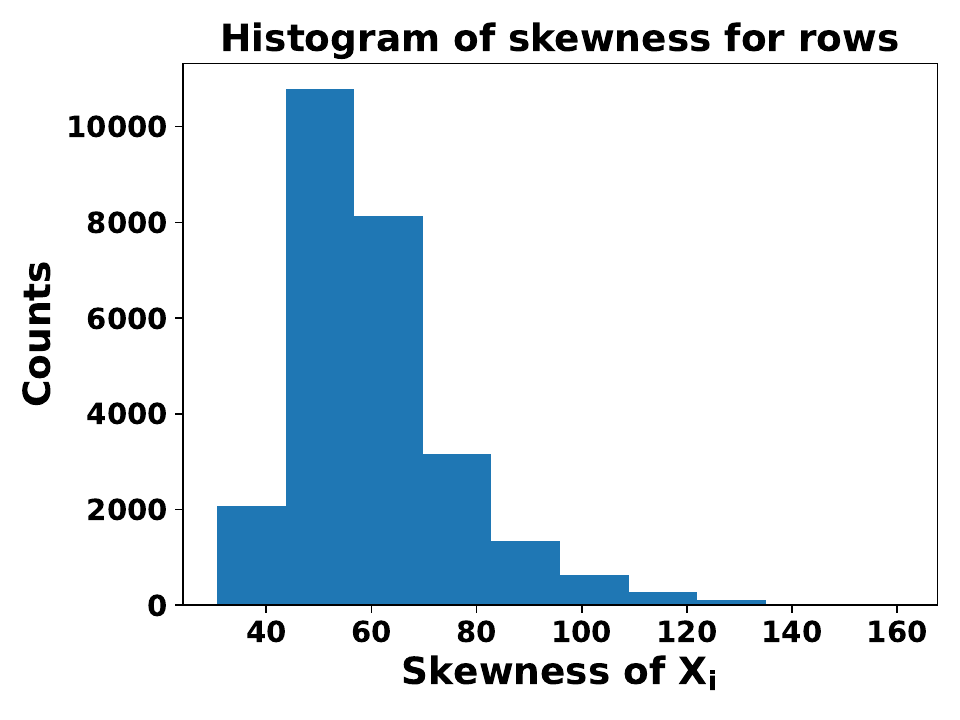} &
    \includegraphics[width=0.45\textwidth]{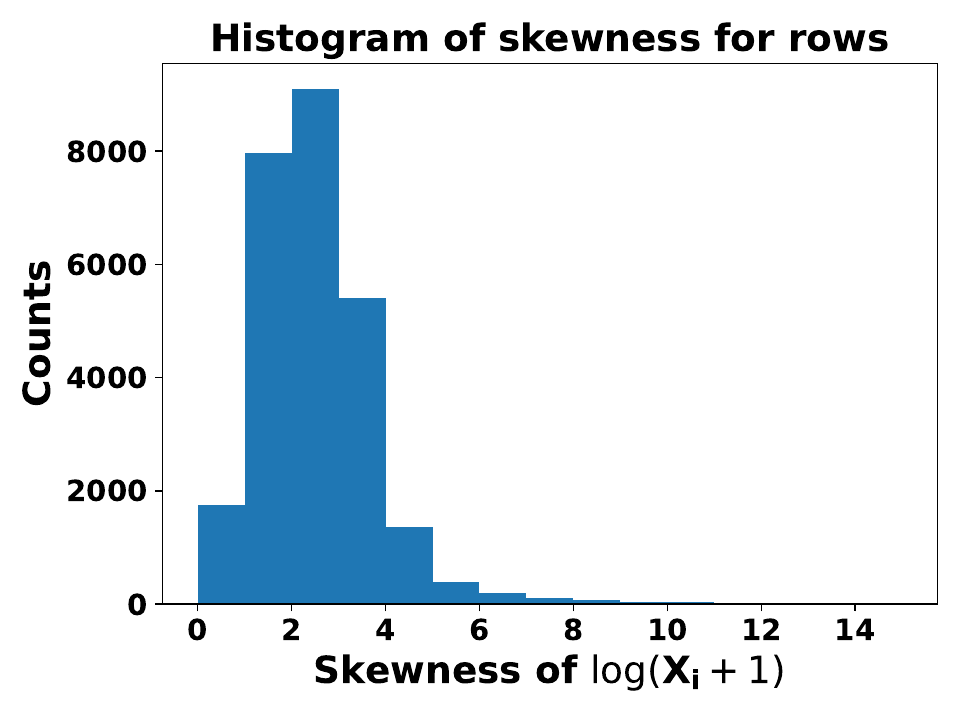}
    \end{tabular}
    \caption{\hlboth{ Histogram of skewness for each row in raw co-occurrence count matrix (left panel) and the log co-occurrence count (right panel) constructed from Wikipedia dump.} }
    \label{fig:skewness_wordpiece_wiki_all_rows}
\end{figure}

\hlboth{
 We use relative convergence criteria on the loss function by checking if the
 %relative change in loss defined below  %
 magnitude of change in the loss function compared to the absolute value of current loss function
is less than a predefined threshold. In computation, we set the convergence threshold as $\epsilon = 10^{-4}$ and the maximum number of iterations ($maxit$) to be 100.
That is, the model is said to be converged if the relative change in Loss below is less than $10^{-4}$.
}
%
%\colorlet{shadecolor}{yellow!35}
%\begin{shaded*}
{\modified
\bqan
\label{formula:relative-change-loss-ch4}
\text{Relative change in Loss } = \frac{|Loss(\vbeta^{(t+1)}, \wtvbeta^{(t+1)}) - Loss(\vbeta^{(t)}, \wtvbeta^{(t)})|}{|Loss(\vbeta^{(t+1)}, \wtvbeta^{(t+1)})| + 0.1}.
\eqan
}
%\end{shaded*}
%}

\hlboth{
Figures \ref{fig:sa_tweedie_19_3_june30_training_loss_plot_vertical} presents the loss curves. They behave nicely in that it drastically goes down in very few iterations and stay with a monotone decreasing trend for all embedding dimensions. The algorithm %for $d=100$ and $d=300$
converged in less than 30 iterations based on the relative convergence criterion. Upon convergence, the values of the loss function are $-8.2186\times 10^8$ and $-8.3284\times 10^8$ for $d=100$ and $d = 300$, respectively. Their relative loss change values are $9.7895\times 10^{-5}$ and $9.4375 \times 10^{-5}$, respectively.
}

\begin{figure}[H]
    \centering
    \includegraphics[width=0.6\textwidth]{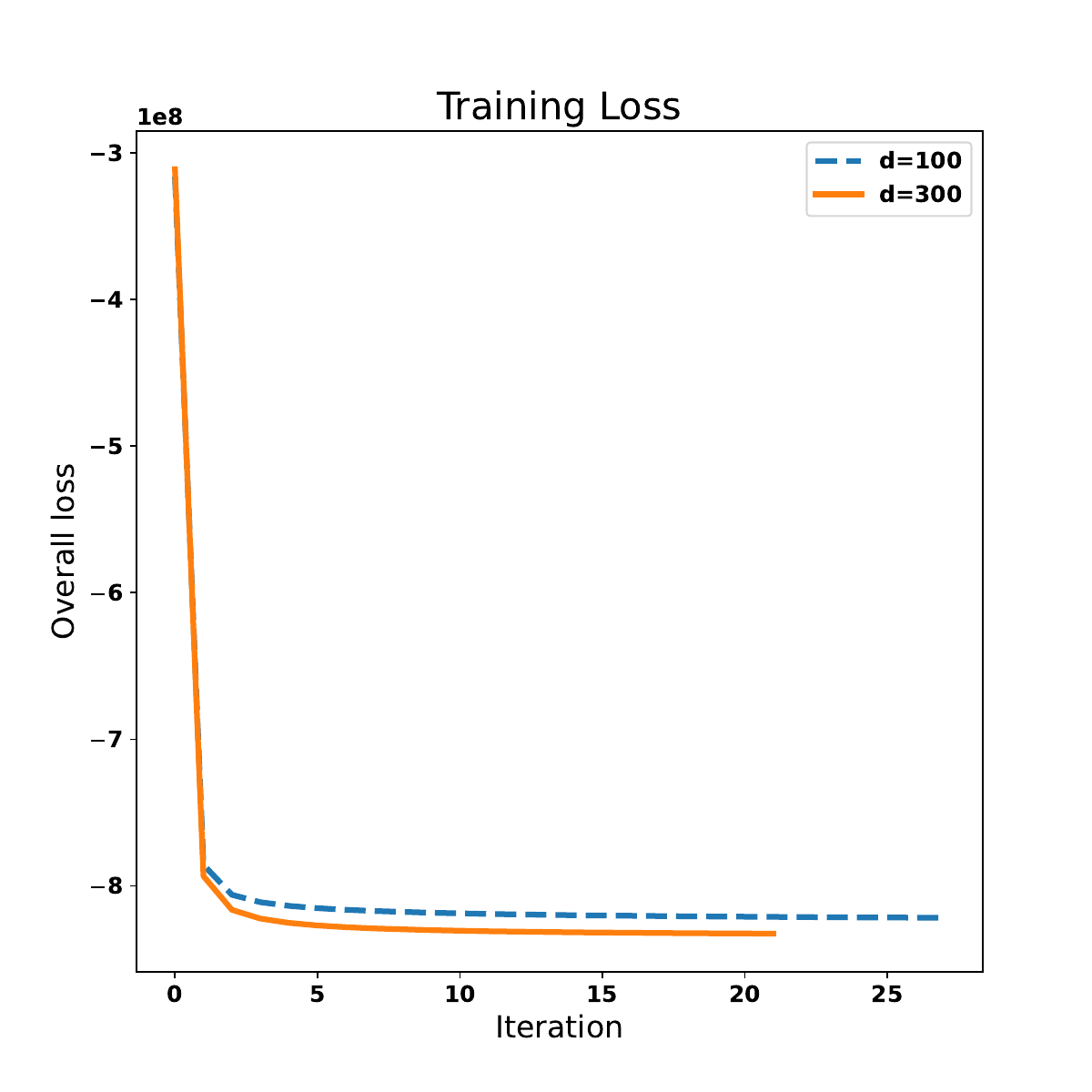}
    \vspace{-0.2in}
    \caption{
    \hlboth{
    Trajectory of the training process of %version 19-3 of
    SA-Tweedie. %Three different embedding dimensions, 100, 300, and 768,
    Two embedding dimensions (100 and 300) are considered.
    %The left panel shows the entire training process. The right panel shows the details after the third iteration.
    The model achieved lower loss with higher embedding dimension.}
    }
    \label{fig:sa_tweedie_19_3_june30_training_loss_plot_vertical}
\end{figure}

\subsection{{\modified Application to NER task on CoNLL-2003 data}}

\hlboth{
In this section, we consider Named Entity Recognition (NER) task using the CoNLL-2003 English benchmark dataset. CoNLL-2003 is a collection of documents from Reuters newswire articles. Each sentence in the dataset is annotated with nine categories representing the beginning or inside of a named entity from four entity types: person, location, organization, and miscellaneous. The nine categories are B-LOC, B-MISC, B-ORG, B-PER, I-LOC, I-MISC, I-ORG, I-PER, O. The `O' category is used for cases that do not belong to any of the four entity types. The %NER task on CoNLL-2003
data has generally very short input text. % compared to the IMDB data.
The output categories are highly unbalanced. The O class is the majority and many other classes have very small number of observations. Such unbalanced multi-class classification is much more challenging than binary classification case such as in Movie sentiment review.}

\hlboth{
\cite{Pennington-etal-2014-glove} presented their result of analysis for this dataset. They used a
comprehensive set of 437,905 discrete features that comes
with the standard distribution of the Stanford NER
model. %(\citealt{finkel-et-al-2005-Stanford-NER}).
In addition, they added 50-dimensional
vectors for each word of a five-word context as continuous features. With these
features as input, they trained a conditional random
field (CRF) with exactly the same setup as the CRFjoin model of \cite{wang-manning-2013-effect}. They reported F1-weighted score $93.2\%$ and $88.3\%$ on the validation and test set performance. In this subsection, we present our result of using a simple BiLSTM model with GloVe, random embedding, or SA-Tweedie embedding with 100 or 300 dimensional word vector representations. The GloVe performance in this simple model (around 88\% on test set) concurs with the numbers reported by \cite{wang-manning-2013-effect} as described above. The simple model with SA-Tweedie embedding achieved over 91\% in F1-weighted score for the test dataset. }
% Table 4 shows results on the NER task with the
% CRF-based model. The L-BFGS training terminates when no improvement has been achieved on
% the dev set for 25 iterations. Otherwise all configurations are identical to those used by Wang and
% Manning (2013).

\hlboth{
 For our analysis, the vocabulary was generated from the training data. It contains all tokens with at least one count. The training and test data were both converted to lower case. Note that lower or upper case letters do contribute to the NER task since many named entities are in upper case letters. By converting to lower case letters, we increased the difficulty level of the NER task.
 }

%We tried with a similar model architecture as used in the previous subsection (BiLSTM + GlobalMaxPool1D + Dense layer with ReLU + Dense layer with softmax activation). Unfortunately, these two datasets have very different nature. The IMDB data has more tokens in each movie review and only two possible outcomes (pos or neg). This translated to more information in the input variable and relatively easier task for the output prediction since it is binary case.

%The NER task on CoNLL-2003 data has generally very short input text. % compared to the IMDB data.
%The output variable has 9 categories highly unbalanced. The O class is the majority and many other classes have very small number of observations. Such unbalanced multi-class classification is much more challenging than binary classification case.

\hlboth{
For this task, we used a simple version of bidirectional LSTM model using our token embedding to perform the classification. The model architecture starts with an embedding layer using the pre-trained token embedding from either SA-Tweedie or GloVe, followed by a bidirectional dynamic RNN layer with variable input size padded to common length and 512 hidden units, a max pooling layer with output dimension 512, a linear layer with output dimension 256 and relu activation, and then a linear activation layer at the end with output dimension 9.  The loss function is the cross entropy.
}
% Train_loader takes a batch of the training data's indices of batch_size. Then apply the collator function to it. In this case, the collator arranges the rows in the batch in decreasing order based on lengths of tokens in the row. Then pad each row with 0s so that all rows have the same length. The returned tokens is a tensor of the indices of tokens in the training vocab. The number of rows of tokens is equal to batch_size, and the number of columns is equal to the maximum length in that batch. The first row has the largest number of nonzero entries. The last row has the least number of nonzero entries. The returned labels stores indices of the labels, also padded and in decreasing of lengths. The returned lengths gives a 1-d tensor of the number of nonzero entries in each row, in the same decreasing order of lengths.

\hlboth{
The model training used dataloader with batchsize = 256 to load the data. The
tokens and labels are padded to the same maximum length for all text data in the same batch.
Adam with lr =0.001 was used to conduct parameter update.  No weight decay was used (i.e. weight decay =0) and amsgrad = False. The maximum $L_2$ norm of gradients w.r.t all parameters were truncated (i.e. clipped) at 10\% of the total norm.}

\hlboth{
15 epochs were trained to fine tune the embedding vectors and other model parameters. Within the 15 epochs, the best model  parameter estimates were chosen to be those that give the smallest validation loss.
GloVe often took 1 to 2 epochs to reach its best validation loss. Random embedding took 7 to 8 epochs to reach its lowest validation loss. SA-Tweedie took 4 to 5 epochs to reach its best validation loss.
The model with the lowest validation loss  was then selected and applied to the test data.
}
% \bit
% \item SA-tweedie 19-3 version vs GloVe
% \item NER CoNLL-2003 prediction result on test data  d=300 BiLSTM512
% \eit
% \begin{table}[ht]
% \centering
% \begin{tabularx}{0.9\textwidth}{lrXXXX}
%   \hline
%  Embedding & Seed & Validation Loss & Test Loss & F1-Weighted Validation & F1-Weighted Test \\
%   \hline
%   Random & 42 & 0.2342 & 0.4484 & 93.61 & 87.78 \\
%   Random & 12 & 0.2312 & 0.3907 & 93.22 & 88.18 \\
%   GloVe & 42 & 0.9102 & 1.5309 & 92.87 & 88.48 \\
%   GloVe & 111 & 0.8706 & 1.4694 & 93.13 & 88.66 \\
%   GloVe & 12 & 0.9321 & 1.5657 & 93.13 & 88.67 \\
%   Random & 111 & 0.2148 & 0.4044 & 94.09 & 88.72 \\
%   SA-Tweedie & 12 & 0.1941 & 0.3353 & 94.85 & 91.00 \\
%   SA-Tweedie & 111 & 0.1896 & 0.3396 & 95.01 & 91.05 \\
%   SA-Tweedie & 42 & 0.1836 & 0.3342 & 95.14 & 91.06 \\
%    \hline
% \end{tabularx}
% \caption{Performance of BiLSTM model with Maxpooling 512 output dimension on three embedding methods}
% \label{tab:NER-f1-weighted}
% \end{table}

\begin{table}[ht]
\centering
\setlength{\tabcolsep}{6pt}
%\colorlet{shadecolor}{yellow!35}
%\begin{shaded*}
{\modified
\begin{tabular}{c|rrr|rrr|rrr}
  \hline
Embedding & \multicolumn{3}{c|}{Random} & \multicolumn{3}{c|}{GloVe} & \multicolumn{3}{c}{SA-Tweedie} \\
  Seed & 12 & 42 & 111 & 12 & 42 & 111 & 12 & 42 & 111 \\
  Epoch & 5 & 8 & 7 & 2 & 2 & 2 & 5 & 5 & 4 \\
  \hline
  Valid loss & 0.231 & 0.234 & 0.214 & 0.932 & 0.910 & 0.870 & 0.194 & 0.183 & 0.189 \\
  Valid F1 B-LOC & 32.51 & 33.41 & 33.08 & 33.01 & 33.09 & 32.14 & 33.69 & 33.40 & 33.01 \\
  Valid F1 B-MISC & 15.64 & 16.15 & 15.75 & 15.10 & 14.54 & 14.28 & 16.20 & 15.71 & 16.05 \\
  Valid F1 B-ORG & 20.37 & 20.16 & 21.19 & 19.62 & 17.55 & 19.57 & 19.47 & 21.2 & 21.33 \\
  Valid F1 B-PER & 24.57 & 23.96 & 24.39 & 22.44 & 21.96 & 22.37 & 24.45 & 24.66 & 24.39 \\
  Valid F1 I-LOC & 3.71 & 4.84 & 4.45 & 4.43 & 4.55 & 4.85 & 5.05 & 5.04 & 4.85 \\
  Valid F1 I-MISC & 4.90 & 4.97 & 4.42 & 3.44 & 3.72 & 3.91 & 4.29 & 4.82 & 4.75 \\
  Valid F1 I-ORG & 7.27 & 7.65 & 8.01 & 5.64 & 5.57 & 5.98 & 6.78 & 8.31 & 7.86 \\
  Valid F1 I-PER & 17.18 & 18.69 & 18.76 & 11.73 & 11.88 & 11.99 & 17.92 & 19.00 & 17.81 \\
  Valid F1 O & 93.36 & 93.27 & 93.76 & 95.47 & 95.40 & 95.50 & 96.38 & 96.23 & 96.41 \\
  Valid F1-weighted & 93.22 & 93.61 & 94.09 & 93.13 & 92.87 & 93.13 & \textbf{94.85} & \textbf{95.14} & \textbf{95.01} \\
  \hline
  Test loss & 0.390 & 0.448 & 0.404 & 1.565 & 1.530 & 1.469 & 0.335 & 0.334 & 0.339 \\
  Test F1 B-LOC & 28.07 & 27.75 & 27.92 & 27.92 & 28.15 & 27.49 & 27.46 & 27.16 & 26.91 \\
  Test F1 B-MISC & 10.58 & 10.90 & 11.03 & 10.15 & 10.05 & 9.43 & 11.28 & 11.08 & 11.10 \\
  Test F1 B-ORG & 22.73 & 23.06 & 24.11 & 20.68 & 18.21 & 20.26 & 22.64 & 24.48 & 24.89 \\
  Test F1 B-PER & 17.73 & 17.28 & 17.48 & 13.81 & 13.64 & 13.90 & 17.00 & 17.30 & 16.40 \\
  Test F1 I-LOC & 2.85 & 4.15 & 3.47 & 3.57 & 3.52 & 3.93 & 3.69 & 3.82 & 3.44 \\
  Test F1 I-MISC & 2.53 & 2.57 & 2.43 & 2.04 & 2.20 & 2.21 & 2.46 & 2.65 & 2.75 \\
  Test F1 I-ORG & 8.70 & 9.04 & 9.13 & 6.75 & 6.65 & 7.19 & 8.70 & 9.58 & 9.58 \\
  Test F1 I-PER & 13.21 & 14.82 & 14.69 & 4.88 & 4.96 & 5.08 & 12.82 & 13.94 & 11.64 \\
  Test F1 O & 89.00 & 88.28 & 89.18 & 93.72 & 93.75 & 93.67 & 95.15 & 94.33 & 95.33 \\
  Test F1-weighted & 88.18 & 87.78 & 88.72 & 88.67 & 88.48 & 88.66 & \textbf{91.00} & \textbf{91.06} & \textbf{91.05} \\
   \hline
\end{tabular}
}
%\end{shaded*}
\caption{\hlboth{ Validation and test metrics using random, GloVe, or SA-Tweedie embedding for NER task. Embedding dimension is 300 for all three methods, and used BiLSTM model with max pooling output dimension 512. Seed: global seed set for random number generation on both CPU and GPU. Epoch: the epoch number that validation loss reached the minimum. Random embedding were generated from the uniform (-0.05, 0.05) as is the default setting in Tensorflow Keras.}}
\label{tab:NER_performance}

\end{table}
% \begin{table}[ht]
% \centering
% \begin{tabular}{llrr|llrr}
%   \hline
%   \multicolumn{4}{c|}{Validation set performance} & \multicolumn{4}{c}{Test set performance}  \\
%  Metric & Category & Glove & SA-Tweedie   & Metric & Category & GloVe & SA-Tweedie \\
% \hline
%  Loss &  & \textcolor{red}{0.7421} & \textbf{0.2572}                   &Loss & & \textcolor{red}{1.2871} & \textbf{0.4478} \\
%  F1 & B-LOC: & 33.41 & 33.40               & F1  & B-LOC: & 28.14 & \textcolor{red}{27.83} \\
%  F1 & B-MISC: & 15.24 & \textbf{16.04}              & F1 & B-MISC: & \textcolor{red}{10.04} & 11.04 \\
%  F1 & B-ORG: & 19.23 & \textbf{20.56}               & F1  & B-ORG: & \textcolor{red}{20.44} & 23.46 \\
%  F1 & B-PER: & 20.61 & \textbf{24.50}               & F1  & B-PER: & \textcolor{red}{11.94} & 17.41 \\
%  F1 & I-LOC: & 4.74 & \textbf{4.81}                 & F1    & I-LOC: & \textcolor{red}{3.57} & 3.78 \\
%  F1 & I-MISC: & 4.28 & 4.64                & F1   & I-MISC: & \textcolor{red}{1.93} & 2.43 \\
%  F1 & I-ORG: & 6.27 & 7.43                 & F1    & I-ORG: & \textcolor{red}{7.45} & 9.08 \\
%  F1 & I-PER: & 11.50 & 18.00               & F1  & I-PER: & \textcolor{red}{4.51} & 12.71 \\
%  F1 & O: & 95.28 & 95.35                   & F1 & O: & 92.93 & \textcolor{red}{92.55} \\
%  F1 & weighted: & \textbf{93.02} & \textbf{94.56}            & F1 & weighted: & \textbf{87.83} & \textbf{90.17} \\
%    \hline
% \end{tabular}
% \caption{ The loss refers to Cross-Entropy loss}
% \end{table}

\begin{table}[ht]
    \centering
 %   \colorlet{shadecolor}{yellow!35}
%\begin{shaded*}
%\begin{enumerate}[wide = 0pt, before=\vspace*{-\dimexpr\topsep+\partopsep+1.5ex}, after=\vspace*{-\dimexpr\topsep+\partopsep+1.5ex}]
{\modified
    \begin{tabular}{lrr} \hline
        & Dev F1 & Test F1 \\
        \hline
       Fine-tuning approach & & \\
       \quad BERT large & 96.6 & 92.8 \\
       \quad BERT base & 96.4 & 92.4 \\
       \hline
       Feature-based approach (BERT base) & & \\
       \quad Embeddings  & 91.0 & - \\
       \quad Second-to-Last Hidden & 95.6 & - \\
       \quad Last Hidden & 94.9 & - \\
       \quad Weighted Sum Last Four Hidden & 95.9 & - \\
       \quad Concat Last Four Hidden & 96.1 & - \\
       \quad Weighted Sum All 12 Layers & 95.5 & - \\
       \hline
    \end{tabular}
    }
%\end{shaded*}

    \caption{\hlboth{ F1 score reported in Table 7 of \cite{devlin-etal-2018-bert} from various BERT model on NER task of CoNLL-2003 data. The fine-tuning approach takes the representation of the first cased WordPiece sub-token of each word as the input and fine tune the pretrained BERT as a classification task. The feature-based approach froze the parameters from one or more layers of the pretrained BERT and used them as input to 768-dimensional BiLSTM for the tagging task.}}
    \label{tab:BERT_NER_F1score}
\end{table}

\hlboth{
The results with max pooling dimension 512 are presented in Table \ref{tab:NER_performance}.
The random embedding and GloVe both achieved less than 94\% in weighted F1 score on the validation data and less than 89\% in weighted F1 score on the test data.
SA-Tweedie achieved around 95\% on weighted F1 score on the validation data and 91\% weighted F1 score on the test data.}

\hlboth{
To get a sense of how good SA-Tweedie performs, we can compare it with the result from BERT models. In the BERT models on the NER task, \cite{devlin-etal-2018-bert} used \textbf{case-preserving WordPiece models along with the maximal document context provided by the data}
to learn the context specific representation of words. The F1 score of various BERT models on the validation and test set is given in Table \ref{tab:BERT_NER_F1score}. Fine-tuning pre-trained BERT large and base models achieved F1 score on test set as 92.8 and 92.4 respectively. On the validation set, fine-tuning BERT large and base showed F1 score 96.6 and 96.4 respectively. Note that fine tuning re-estimates all parameters in the pretrained BERT model in addition to the BiLSTM model parameters.
In their feature-based approach, which freezes the parameters from pre-trained BERT embeddings and estimate only BiLSTM model parameters, they reported F1 score of 91 in validation data for BERT and did not report performance on test data. Under this setting, our SA-Tweedie gives F1 score of around 95 for validation data and about 91 for test data. This comparison clearly shows that SA-Tweedie embeddings are much more effective than BERT embeddings. }

\hlboth{
In addition to the superior performance demonstrated by SA-Tweedie embeddings, it is important to highlight that these results are achieved with a significantly smaller model and less training data.} %compared to BERT.
\begin{enumerate}
\item 
Examine the feature-based approach that estimates only the BiLSTM model parameters while keeping the BERT parameters fixed. The BERT large model, for example, consists of 24 hidden layers and uses a 1024-dimensional embedding, while the BERT base model has 12 hidden layers with a 768-dimensional embedding. The total number of parameters is 345 million for the BERT large model and 110 million for the BERT base model. In contrast, our SA-Tweedie only has 300-dimensional embedding and there are no layers.  Even when using the 768-dimensional BiLSTM with BERT, the number of parameters remains much larger than that of our simpler model. Specifically, we use a one-layer, 512-dimensional BiLSTM model with SA-Tweedie’s 300-dimensional embedding and a max pooling layer, which significantly reduces the number of parameters. 
  \item The embeddings from GLoVe and BERT were trained with much larger sized corpora than our SA-Tweedie model. The corpus size of our SA-Tweedie is 4 Billon tokens from Wikipedia. The pretrained GLoVe embeddings were based on training corpus size of 42 billion tokens. The BERT embeddings were from pretraining on English Wikepedia plus additional BooksCorpus (800M words) data. We do not have access to BooksCorpus data.  Therefore SA-Tweedie's embedding has seen less training data.
\end{enumerate}
  
  With these considerations in mind, the BiLSTM model with pretrained SA-Tweedie embeddings offers a more efficient path to improved performance, leveraging a simpler architecture.

% The figure below shows BERT's NER task F1 score on validation set of the CoNLL-2003 data.
% \begin{figure}[H]
%     \centering
%     \includegraphics[width=0.7\textwidth]{figures/chapter_7/BERT_NERtask_CoNLL2003_F1score.png}
%     \label{fig:BERT_NER_F1score}
% \end{figure}

\hlboth{
What we presented in this section is for one setting of the analysis. We also experimented on some other model settings. In particular, we considered varying the model architecture a little bit by changing the max pooling output dimension. When max pooling is used, we considered max pooling output dimension 256 and 512. When no max pooling is used, we denote the max pooling dimension as 1024, which is equal to the output dimension from the previous RNN layer.
Further settings include differentiating capital versus lower case letters by adding an additional dimension in the vector representation, for which the value was taken to be the indicator of whether capital letters are present in the word. The result varies for all embedding methods.}

\hlboth{
For SA-Tweedie, we also considered using the representation of only first piece of the WordPiece tokens from any word as an alternative. For some words, the WordPiece tokenizer splits one word into multiple pieces of tokens. The SA-Tweedie gives vector representation for all pieces separately. In the end, the representation of the word needs to either combine the representation from different pieces or take just the representation of the first piece alone. The authors of \cite{devlin-etal-2018-bert} presented results of NER analysis with BERT using first piece token only. To combine the representation vectors from pieces to form a single representation of the word,
%many methods have been explored in the literature (cf. \citealt{Wu-et-al-Wordpiece:2016} and the references therein). In general,
some form of positional encoding is needed to preserve the relative location of the token in the word.
The positional encoding is included along with the token representations throughout the model parameter estimation process.
%What form of positional encoding to use is also an active research area in the NLP community. See \citealt{positional-encoding-overview-2021} for a survey of different positional encoding methods.
We used the simple form of positional encoding provided by \cite{Transformer:2017}. That is, the positional encoding is a matrix that has number of columns equal to the dimension of the word representation. The number of rows is equal to the number of tokens from the word. Suppose there are $k$ WordPiece tokens and let $pos$ enumerate the positions of the tokens in the word. The positional encoding for the $i^{th}$ column in the $pos$ row is given by}
%\colorlet{shadecolor}{yellow!35}   %
%\begin{shaded*}
%\vspace{-0.2in}
\bqa
\text{PE}_{(\text{pos}, 2 i)}  =\sin \left(\frac{\text{pos}}{ 10000^{2 i / d}}\right), \qquad
\text{PE}_{(\text{pos},2i+1) }  =\cos \left(\frac{\text { pos }} { 10000^{2 i / d}}\right),
\eqa
%\end{shaded*}
\hlboth{where
pos = $0, \ldots, k-1$, $i =0, \ldots, d/2.$ Generally, we see that the SA-Tweedie with only first piece token performs better than using combination of the token pieces.}

\hlboth{
Figure \ref{fig:NER-F1-test-compare-all} shows the comparison of performance from all three embedding methods. For the NER task, the SA-Tweedie embedding gives the best results in that the overall F1-weighted scores are consistently higher than those from GloVe and random embedding. This pattern can be seen by directly comparing performance under the same seed and the same lower case condition shown in Figure \ref{fig:NER-F1-test-compare-all}.
}

\begin{figure}[H]
    \centering
    \includegraphics[width = 1\textwidth]{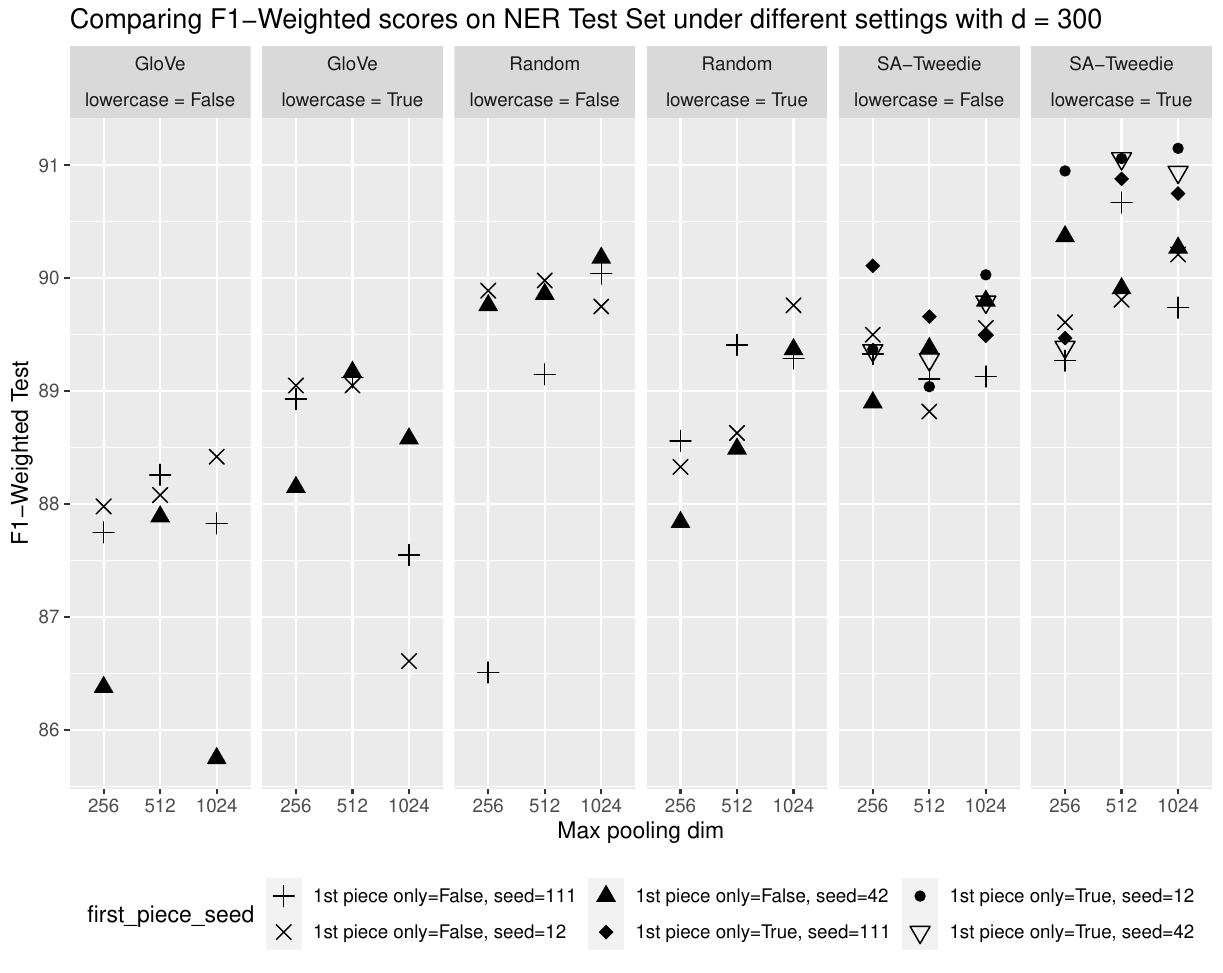}
    %\vspace{-0.6in}
    \caption{\hlboth{ Weighted F1 score on NER test set for different settings. All embeddings used 300-dimensional representations.}}
    \label{fig:NER-F1-test-compare-all}
\end{figure}
% This SA-Tweedie embedding is june17-avg version.
\hlboth{
The random embedding even outperformed the GloVe embedding in this task. Further investigation show that the model using GloVe embedding tend to quickly overfit (see Figure \ref{fig:NER_loss_f1score_plot}) in just a couple of epochs.
}

 \begin{figure}[H]
    \centering
    \includegraphics[width=0.7\textwidth]{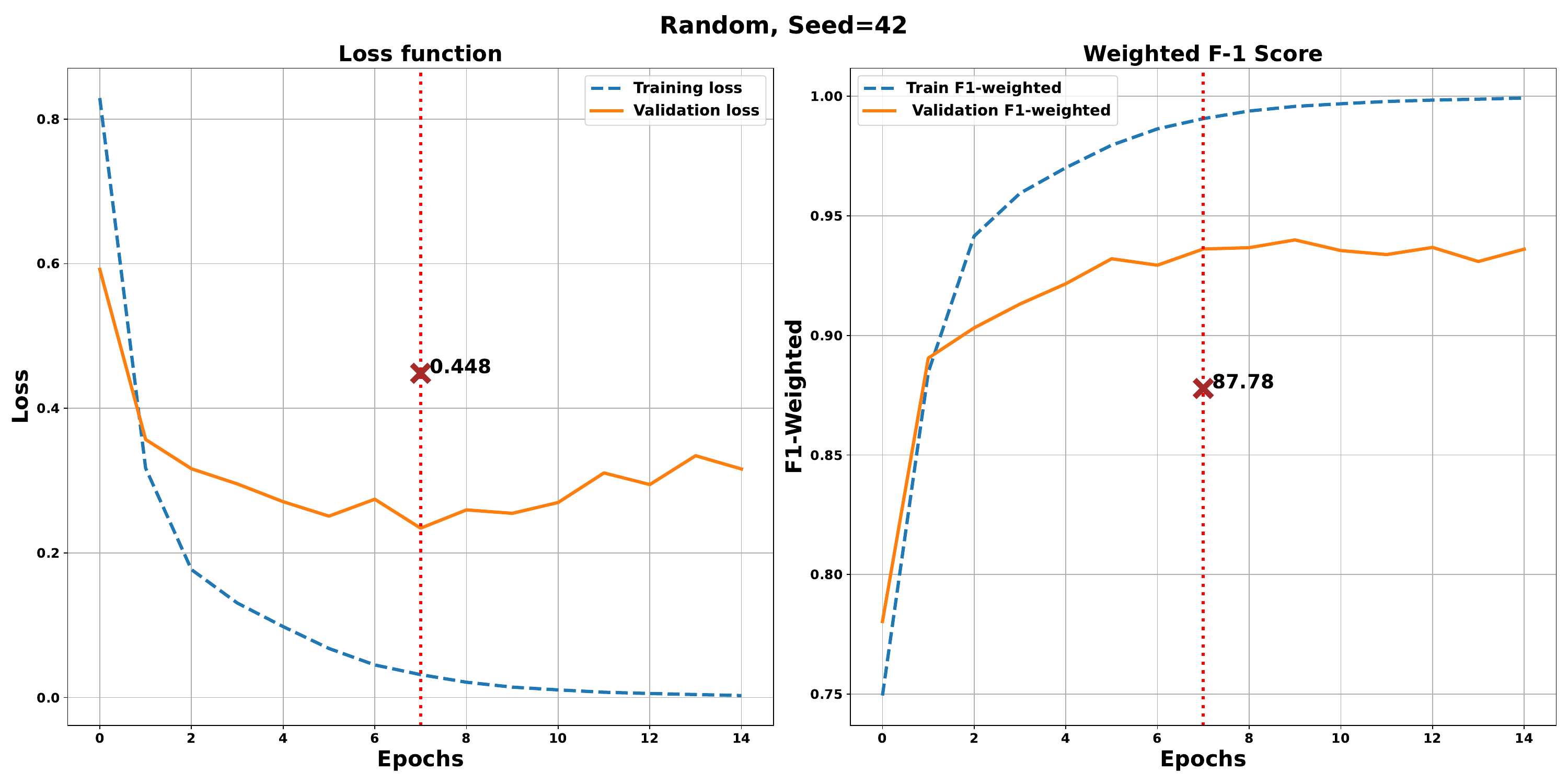}
    \includegraphics[width=0.7\textwidth]{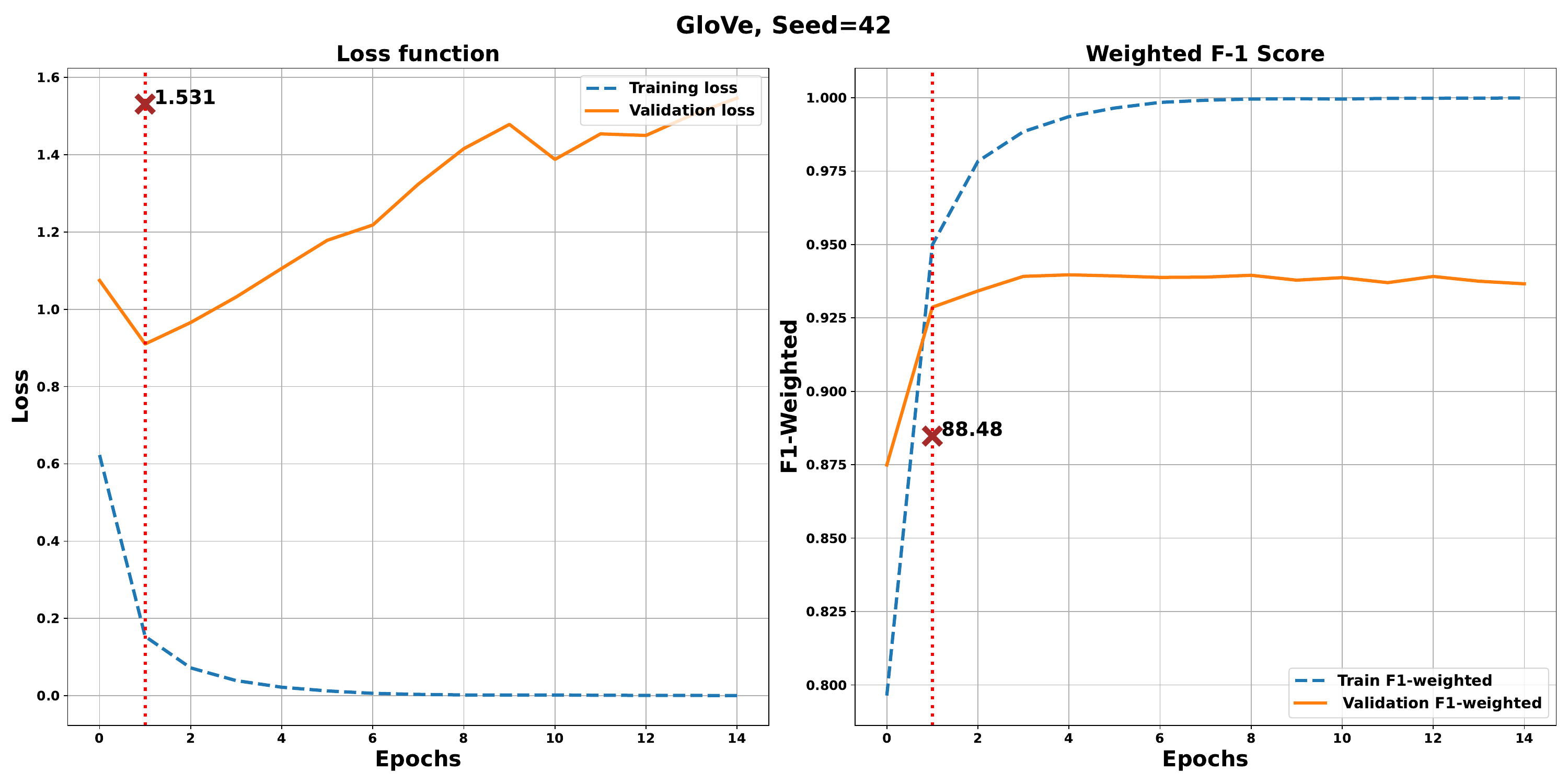}
    \includegraphics[width=0.7\textwidth]{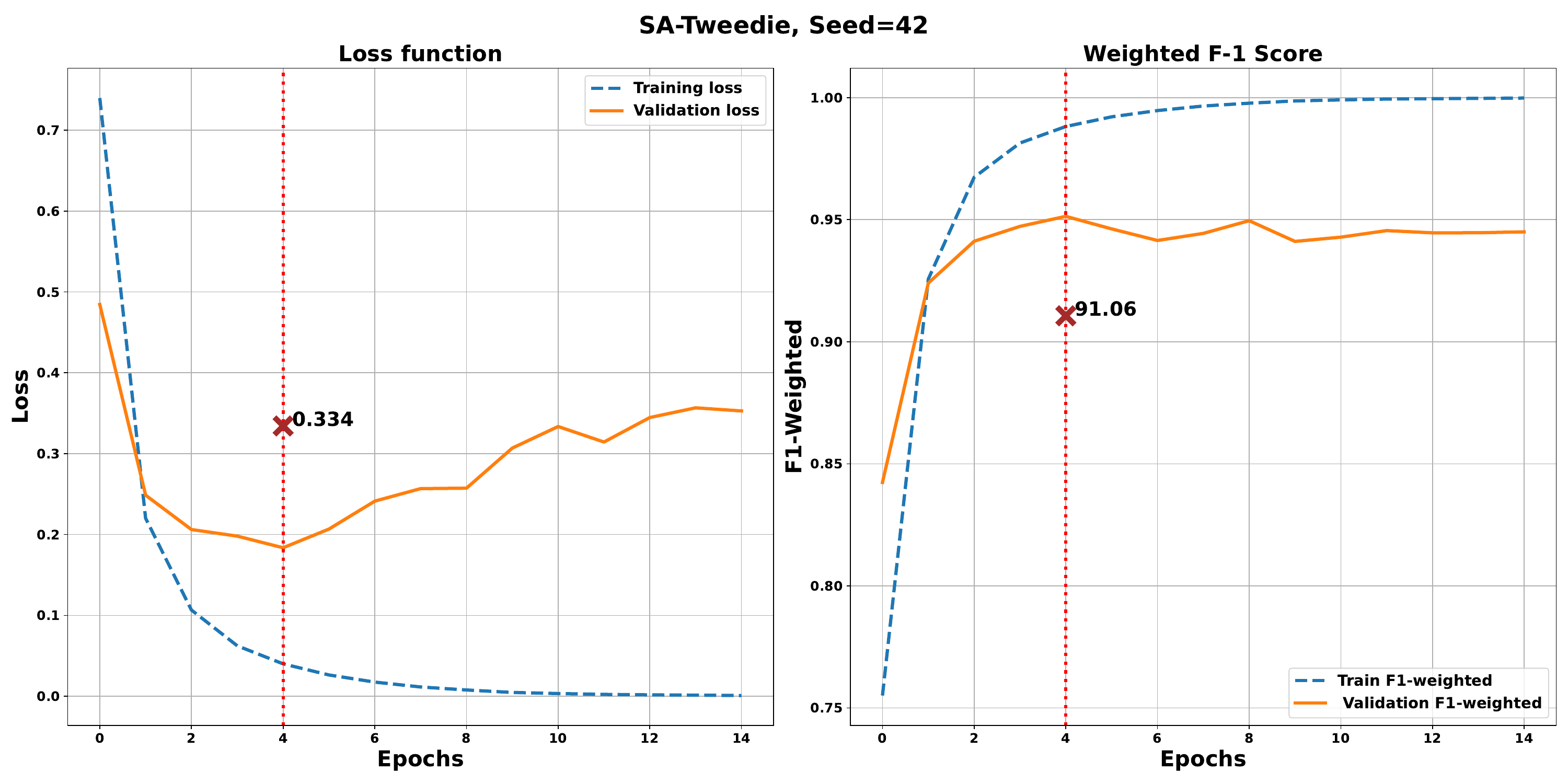}
    %\vspace{-0.3in}
    \caption{\hlboth{ Training and validation loss along with training and validation weighted F1 score for 15 epochs. Top row: random embedding. Middle row: GloVe embedding. Bottom row: SA-Tweedie embedding. All parameter initialization used identical global seed 42. The loss and weighted F1 score for test data are marked with cross marks with value given beside them.}}
    \label{fig:NER_loss_f1score_plot}
\end{figure}

\hlboth{
In fact, majority of the GloVe cases show drastic overfit starting from the second epoch. The random embedding starts with no information and then progresses little by little to reduce the training loss until a certain point. The SA-Tweedie is in between the the two. A few epochs of training and parameter updates still benefits the model with SA-Tweedie embedding.
}

\hlboth{
A slight surprise happens when we compare the models that convert all words to lower-case letters with those that are cased. It is often thought that named entities tend to use capital letters and therefore keeping capital letters may improve the performance. This is the case with the random embedding. The middle two panels in Figure \ref{fig:NER-F1-test-compare-all}
show that the extra information of capital versus lower case letters provides an additional boost to the model using random embedding. On the other hand, the trend is opposite for GloVe or SA-Tweedie embedding. We see a slight decay here in the performance of using capital letters compared to the lower cases for both embedding methods. This tells that the embeddings themselves already carried more information than what was provided in the extra column of indicator variable of capital versus lower case letters.  Instead of being beneficial, the single extra dimension of indicator variable we added in the model served as a nuisance.  To make capital letters helpful in prediction,
%the embedding of the words with capital letters should be learned during the whole training process instead of compensating at the end. We could not take this direction for the dissertation due to time limitation. A
a more suitable approach is to learn word embeddings for both capital and lower case words from the start but this significantly increases the computational demand which requires access to many TPUs or pods of GPUs for extended time access. In the current settings, SA-Tweedie using only first piece token and lower case letters outperforms all other methods and settings.
}

}

\section{Summary, discussion, and future research}

In this article, we presented SA-Tweedie model to estimate the global dense vector representa- tions for english words via language modeling task using unsupervised text data. We presented an algorithm with multiple updating schemes and compared their performance in both simulation and real applications. Our algorithm with the popular Adam update performed well in the sense that it reduces the loss both within the innermost loop over epochs and throughout the iterations in the outer loop. However, when compared to our algorithm with the Fisher scoring update in the inner loop with or without learning rate adjustment, the Adam is inefficient since its overall loss or loss within each epoch reduces slower than the other two updating schemes. The Fisher scoring update without learning rate adjustment optimizes the algorithm in the right direction but sometimes fails to detect the right direction as we have seen in multiple simulated datasets. The Fisher scoring update with learning rate adjustment performed the best because it consistently find the right direction to reduce the overall loss and reach the smallest loss value among these three update methods. In the end of all iterations, the difference of overall loss between with and without learning rate adjustment is negligible compared to its magnitude.

\hlynob{
We also derived, implemented, and experimentally evaluated pseudo-likelihood models for the co-occurrence count data. Due to space limitations, we did not include this part in the paper. Interested readers can find the details in Section 4.3 of \cite{Kim2023}.}

\hlynob{
The pseudo-likelihood model consistently updated the parameters in incorrect directions, resulting in NaN or infinite values appearing frequently in the intermediate estimates of model components at the early stages of iterations. Clipping the gradients or estimated model components only provided temporary relief for a few steps but failed to guide the algorithm towards the correct estimating direction.}

\hlynob{
Our simulations show that the pseudo-likelihood approach is unable to estimate the parameters when no covariates are observed. The favorable behavior reported in the literature for pseudo-likelihood Tweedie regression critically depends on the availability of covariate values, which is common in most statistical inference scenarios. For pseudo-likelihood to work effectively, it is crucial that the conditional distributions of each variable within the model, given the other variables, are reasonably accurate to approximate the full likelihood well.}

\hlynob{
In our setting, no covariate values are observed. At any intermediate stage of the alternating update, the conditional distribution could be far from the truth, particularly because a surrogate distribution is used in pseudo-likelihood instead of the actual distribution. Additionally, the parameters in the dot product model are not uniquely defined. Orthogonal transformations of the embedding vectors yield the same dot product. Thus, the parameters are identifiable when covariates are observed but not when they are absent. The combined issues of surrogate distribution and non-identifiability cause the pseudo-likelihood to fail in finding the correct optimizing direction.
}

\hlynob{
The unobserved row (or column) vector representations can be treated as missing data. This perspective establishes a connection between the SA-Tweedie model and the generalized EM algorithm (GEM), providing insights into why the SA-Tweedie model is effective. For estimating $\vbeta$, consider the complete data $\vx = (y, \wtvbeta)$. The entire $\wtvbeta$ matrix is missing.
We can treat the $\wtvbeta$  as randomly drawn
from a distribution that has relatively few parameters. Assume the distribution of $\vx$ is in
the exponential family. The convergence behavior of likelihood-based estimation is governed by the relationship between the complete data likelihood and the conditional likelihood. When the mechanism of missing is completely at random and the natural parameter $\theta$ in the exponential family format in equation (\ref{canonical}) is a priori independent of the parameters of the missing data process, the sufficient statistics for the natural parameter in the complete data problem are linear in the data. In this case, the estimation in GEM algorithm is equivalent to the procedure which first estimates the missing data points and then computes the sufficient statistics using fill-in values \citep{Dempster77}.
The sufficient statistics gives the maximum likelihood estimate of the conditional mean upon convergence.}

\hlynob{
The convergence for the alternating procedure was established based on two conditions: (1). The log likelihood function is bounded on the sequence of estimated parameter values;  (2). the sufficient statistic for natural parameter must be linear in the observed data. In SA-Tweedie, these conditions are met.
This is the theoretical foundation for both our SA-Tweedie model and the matrix factorization using Alternating Least Squares (ALS).
 By contrast, the SA-Tweedie model employs a two-stage Bernoulli-Tweedie model rather than a normal distribution as a hidden normality assumption behind the ALS based matrix factorization.
In the case of normal distribution, the constant variance allows the use of ALS to estimate both the row and column representation vectors.
However, the SA-Tweedie model cannot employ ALS because the variance of the Tweedie distribution is not constant. Instead, our alternating scheme leverages the Fisher scoring algorithm, with or without learning rate adjustment, to achieve convergence. This approach benefits from a quadratic rate of convergence when the true parameters and their estimates lie within the interior of the parameter space, conditional on the values of row vector representations during the estimation of column vector representations (and vice versa).}

\hlynob{
Learning rate adjustment is crucial in practical applications because the conditional distribution depends on the current estimates of the vectors being conditioned upon. When these estimates are far from accurate, large updates can lead to instability. By moderating the update step size, the learning rate adjustment ensures more stable and reliable convergence.
}

\hlboth{The global representation provides a starting point for smaller models with improved performance. One of the future research is to develope adaptive fine-tuning techniques for word embeddings that allow for fast, resource-efficient model adaptation to new domains or tasks. This could involve meta-learning or other approaches that enable models to quickly adjust their representations based on limited labeled data for new tasks. Another future direction is to investigate the integration of global word embeddings with multimodal models (combining text with images, audio, and video). This could involve developing novel architectures that learn to combine embeddings from different modalities to improve applications such as multimodal sentiment analysis and cross-modal retrieval, where understanding the relationship between text and other forms of data is critical.
   %It could also be interesting to investigate methods for transferring global word embeddings across diverse domains without losing task-specific performance. This could involve developing techniques to align embeddings from various domains and using domain-adaptive training to improve transferability.
%   Cross-domain transfer learning could lead to more generalizable models, enabling NLP applications in fields like law, healthcare, and education, even with limited labeled data from those domains.
   These research directions, if pursued, could significantly advance the state of global word representations by addressing critical issues such as computational efficiency, resource demands, adaptability across domains, and interpretability. By combining these innovations, we could pave the way for more sustainable and scalable NLP models that can handle diverse and complex real-world tasks.
   } 